\pdfoutput=1

\documentclass[final,12pt]{colt2022}
\usepackage[utf8]{inputenc}
\usepackage[T1]{fontenc}
\usepackage{graphicx}
\usepackage{hyperref}
\usepackage{subfiles}
\usepackage{setspace}
\usepackage{amsmath}
\usepackage{amssymb}
\usepackage{thmtools,thm-restate}
\usepackage{mathtools}
\usepackage[boxruled,lined,linesnumbered]{algorithm2e}
\usepackage{times}
\usepackage{enumitem}
\SetKw{KwRequire}{Require}
\SetKw{KwInput}{Input}
\SetKw{KwInitialize}{Initialize}

\renewcommand*\hat\widehat

\newcommand{\Vtilde}{\widetilde{V}}

\newcommand{\maxOp}{\vee}
\newcommand{\minOp}{\wedge}
\newcommand{\Max}[1]{\max\Set{#1}}
\newcommand{\Min}[1]{\min\Set{#1}}

\newcommand{\tautilde}{\widetilde{\tau}}

\newcommand{\sumtT}{\sum_{t=1}^T}

\usepackage{tikz}
\newcommand\encircle[1]{%
  \tikz[baseline=(X.base)]
  \node (X) [draw, shape=circle, inner sep=0] {\strut #1};}

\newcommand{\R}{\mathbb{R}}

\newcommand{\Set}[1]{\left\{#1\right\}}
\newcommand{\sbrac}[1]{\left[#1\right]}
\newcommand{\brac}[1]{\left(#1\right)}
\newcommand{\cA}{\mathcal{A}}
\newcommand{\cI}{\mathcal{I}}

\newcommand{\cM}{\mathcal{M}}

\newcommand{\cX}{\mathcal{X}}

\newcommand{\cP}{\mathcal{P}}

\newcommand{\cS}{\mathcal{S}}

\newcommand{\cL}{\mathcal{L}}
\newcommand{\cT}{\mathcal{T}}

\newcommand{\cV}{\mathcal{V}}

\newcommand{\argmin}{\operatorname{arg\,min}}

\newcommand{\norm}[1]{\left\|#1\right\|}

\newcommand{\grad}{\nabla}

\newcommand{\inner}[1]{\left\langle #1 \right\rangle}

\newcommand{\Log}[1]{\log\left(#1\right)}
\newcommand{\Exp}[1]{\exp\left(#1\right)}

\newcommand{\E}{\mathbb E}

\newcommand{\defeq}{\overset{\text{def}}{=}}

\newcommand{\eps}{\varepsilon}

\newcommand{\abs}[1]{\left|#1\right|}
\newcommand{\proj}{\Pi}

\newcommand{\zeros}{\mathbf{0}}

\newcommand{\half}{\frac{1}{2}}

\newcommand{\tpp}{{t+1}}
\newcommand{\tmm}{{t-1}}

\newcommand{\Otilde}{\widetilde{O}}

\newcommand{\gtilde}{\widetilde{g}}

\newcommand{\gbar}{\overline{g}}

\newcommand{\Ohat}{\widehat{O}}

\newcommand*\textcite\citet
\newcommand*\parencite\citep

\newcommand{\regret}{\text{Regret}}
\newcommand{\sign}{\text{sign}}

\newcommand{\w}{w}
\newcommand{\wtilde}{\widetilde{\w}}

\newcommand{\x}{x}
\newcommand{\xt}{\x_t}
\newcommand{\xtpp}{\x_{t+1}}

\newcommand{\wt}{\w_t}
\newcommand{\wtpp}{\w_{t+1}}
\newcommand{\wtmm}{\w_{t-1}}

\newcommand{\cmp}{u}
\newcommand{\psiq}{F}

\newcommand{\Vhat}{\widehat{V}}
\newcommand{\Dhat}{\widehat{D}}

\usepackage{natbib}

\title[Parameter-free Mirror Descent]{Parameter-free Mirror Descent}
\usepackage{times}
\coltauthor{%
  \Name{Andrew Jacobsen}
  \Email{ajjacobs@ualberta.ca}\\
  \addr University of Alberta
  \AND
  \Name{Ashok Cutkosky}
  \Email{ashok@cutkosky.com}\\
  \addr Boston University
}

\usepackage{cleveref}

\newtheorem{myproposition}{Proposition}

\newtheorem{manualtheoreminner}{Theorem}
\newenvironment{manualtheorem}[1]{%
  \renewcommand\themanualtheoreminner{#1}%
  \manualtheoreminner
}{\endmanualtheoreminner}
\newtheorem{manualpropositioninner}{Proposition}
\newenvironment{manualproposition}[1]{%
  \renewcommand\themanualpropositioninner{#1}%
  \manualpropositioninner
}{\endmanualpropositioninner}

\begin{document}

\maketitle

\begin{abstract}%
We develop a modified online mirror descent framework that is suitable for
building adaptive and parameter-free algorithms in unbounded domains. We
leverage this technique to develop the first unconstrained online linear
optimization algorithm achieving an optimal dynamic regret bound, and we further
demonstrate that natural strategies based on Follow-the-Regularized-Leader are
unable to achieve similar results. We also apply our mirror descent framework to
build new parameter-free implicit updates, as well as a simplified and improved unconstrained scale-free algorithm.
\end{abstract}


\newcommand{\secPF}{Parameter-free Learning}
\newcommand{\secScaleFree}{Lipschitz Adaptivity and Scale-free Learning}
\newcommand{\secDynamicAlg}{Dynamic Regret Algorithm}
\newcommand{\secImplicit}{Adapting to Gradient Variability}
\newcommand{\secOptTradeoff}{Trade-offs in the Horizon Dependence}

\section{Online Learning}

This paper introduces new techniques for online convex optimization (OCO), a standard framework used to model learning from a stream of data \citep{cesa2006prediction,shalev2011online, hazan2016introduction, orabona2019modern}. Formally, consider $T$ rounds of interaction between an algorithm and an environment. In each round, the algorithm chooses a $w_t$ in some convex subset $W$ of a Hilbert space, after which the environment chooses a convex loss function $\ell_t:W\to \R$. Performance is measured by the \emph{regret} $R_T(u)$, the total loss relative to a benchmark $u\in W$:
\begin{align*}
    R_T(u) = \sum_{t=1}^T \ell_t(w_t) -\ell_t(u)
\end{align*}

Almost all of our development is focused on \emph{online linear optimization}
(OLO), in which the $\ell_t$ are assumed to be linear functions,
$\ell_t(w) = \langle g_t, w\rangle$. This focus is justified by a well-known
reduction (\textit{e.g.} \cite{zinkevich2003online}) which employs the identity
$\ell_t(w_t)-\ell_t(u) \le \langle g_t, w_t-u\rangle$ for any
$g_t\in\partial \ell_t(w_t)$ to show that OLO algorithms can be used to solve OCO problems. The classical algorithm for this setting is \emph{online gradient descent}, which achieves the minimax optimal regret $R_T(u)\le \|u\|\sqrt{\sum_{t=1}^T \|g_t\|^2}$ when the learning rate $\eta$ is set as $\eta = \frac{\|u\|}{\sqrt{\sum_{t=1}^T \|g_t\|^2}}$.

This optimal $\eta$ is of course unknown \textit{a priori}, and so there has been a concerted push to develop algorithms that achieve similar bounds \emph{without} requiring such oracle tuning \citep{duchi10adagrad, mcmahan2010adaptive, foster2015adaptive, mhammedi2020lipschitz, van2019user, cutkosky2018black}. A standard result in this setting is
\begin{align}
    R_T(u)&\le O\left(\epsilon+\|u\|\sqrt{\sum_{t=1}^T \|g_t\|^2 \log( T \|u\|/\epsilon)}\right)\label{eqn:paramfreebound},
\end{align}
which holds for any user-specified $\epsilon$ for all $u$. This is known to be optimal up to constants \citep{orabona2013dimension}. Bounds of this form have many names in the literature, such as ``comparator adaptive'' or ``parameter-free''. We will use ``parameter-free'' in the following.

We will develop a new framework for parameter-free regret bounds that is based on \emph{online mirror descent} (OMD). While OMD is already a standard technique in online learning, it has proven difficult to apply it to the unconstrained setting and achieve parameter-free regret. As a consequence of our development, we are able to produce several new kinds of algorithms. First, we consider the \emph{dynamic regret}. In this setting, the benchmark point $u$ is not a fixed value. Instead, we define the regret with respect to an arbitrary \emph{sequence} of benchmarks $\vec{u} = (u_1,\dots,u_T)$:
\begin{align*}
    R_T(\vec{u}) = \sum_{t=1}^T \ell_t(w_t)-\ell_t(u_t)\le \sum_{t=1}^T \langle g_t, w_t -u_t\rangle
\end{align*}
Dynamic regret is clearly more demanding than the previous definition (called \emph{static} regret). It is also more appropriate for true streaming settings in which data might drift over time, so that a fixed benchmark $u$ is too weak. Using our approach, for any $\epsilon$ we obtain a dynamic regret bound of:
\begin{align*}
    R_T(\vec{u}) &\le \tilde O\left(\epsilon + \sqrt{\sum_{t=1}^T \|g_t\|^2\|u_t\| \sum_{i=1}^{T-1} \|u_{i+1}-u_i\|\log(T\max_t \|u_t\|/\epsilon)}\right)
\end{align*}
This bound holds in both unconstrained and constrained settings. In the
unconstrained setting, this is to our knowledge the \emph{first} non-trivial
dynamic regret bound of any form. In the constrained setting, this bound still
improves upon prior work (\textit{e.g.}
\citet{zhang2018adaptive,jadbabaie2015online,zhao2020dynamic}) by virtue of increased adaptivity to the individual $\|u_t\|$ values. Moreover, we show that our OMD-based analysis appears to be crucial to this result: essentially no ``reasonable'' variant of prior methods for unconstrained online learning is capable of achieving a similar result.

We are also able to apply our OMD analysis in two other ways: first, we can
essentially immediately produce a general framework for parameter-free regret
using \emph{implicit} updates and so move beyond pure OLO. Finally, we show how
to use OMD to improve upon the ``scale-free'' bounds presented by
\cite{mhammedi2020lipschitz} by removing impractical doubling-like restart
strategies as well as reducing logarithmic factors in the regret.

\textbf{Notations.} For brevity, we occasionally abuse notation
by letting $\grad f(x)$ denote an element of $\partial f(x)$. The Bregman
divergence \textit{w.r.t.} a differentiable function $\psi$
is
$D_{\psi}(x|y)=\psi(x)-\psi(y)-\inner{\grad\psi(y),x-y}$.
We use the compressed sum notation $g_{i:j}=\sum_{t=i}^{j}g_{t}$ and $\norm{g}_{a:b}^{2}=\sum_{t=a}^{b}\norm{g_{t}}^{2}$.
We denote $a\maxOp b = \Max{a,b}$ and $a\minOp b=\Min{a,b}$. The indicator
function $\cI_{W}(\cdot)$ is the function such that $\cI_{W}(\w)=0$ if $\w\in W$
and $\cI_{W}=\infty$ otherwise. The notation $O(\cdot)$ hides constants,
$\Ohat(\cdot)$ hides constants and $\log(\log)$ terms, and $\Otilde(\cdot)$
hides up to and including $\log$ factors.

\section{Centered Mirror Descent}\label{sec:centered-md}

\begin{figure*}
\begin{algorithm}[H]
  \SetAlgoLined
  \KwInput{Initial regularizer $\psi_{1}:\R^{d}\to\R_{\ge0}$}\\
  \KwInitialize{ $w_{1}=\argmin_{\w}\psi_{1}(\w)$}\\
  \For{$t=1:T$}{
    Play $\w_t$, observe loss function $\ell_{t}(\cdot)$\\
    Choose regularizer $\psi_{\tpp}$, composite penalty $\varphi_{t}$\\
    Define
    \(
    \Delta_{t}(\w)=D_{\psi_{\tpp}}(\w|\w_{1})-D_{\psi_{t}}(\w|\w_{1})
    \)
    and \(\phi_{t}(\w)=\Delta_{t}(\w)+\varphi_{t}(\w)\)\\
    Update
    \(
      \wtpp = \argmin_{\w}\ell_{t}(\w)+D_{\psi_{t}}(\w|\wt) + \phi_{t}(\w)
    \)
}
\caption{Centered Mirror Descent}
\label{alg:centered-md}
\end{algorithm}
\end{figure*}
In this section we introduce our framework and key technical tools.
Our algorithms are constructed from an instance of
composite mirror descent \citep{duchi2010composite}
depicted in \Cref{alg:centered-md}.
Composite mirror descent can be interpreted as
a mirror descent update which adds an
auxiliary penalty $\phi_{t}(\w)$ to the loss
function $\ell_{t}(\w)$. Typically, $\phi_{t}(\w)$
is a composite loss function which enforces some additional
desirable properties of the solution, such as sparsity.
In contrast, we will use these terms $\phi_{t}(\w)$ as a
crucial part of our algorithms. This composite term is composed
of two parts, $\Delta_{t}(\w)$ and $\varphi_{t}(\w)$, with the distinguishing
feature of our approach being the
$\Delta_{t}(\w)=D_{\psi_{\tpp}}(\w|\w_{1})-D_{\psi_{t}}(\w|\w_{1})$.

To see what this term $\Delta_{t}(\w)$ contributes, assume $\ell_{t}(\w)=\inner{g_{t},\w}$ for
some $g_{t}\in\R^{d}$ and
suppose we set $\psi_{t}$ and $\w_{1}$ such that
$\min_{\w}\psi_{t}(\w)=\psi_{t}(\w_{1})=0$ for all $t$ and $\varphi_{t}(\w)\equiv 0$.
From the first-order optimality condition
$\wtpp=\argmin_{\w\in\R^{d}}\inner{g_{t},\wt}+D_{\psi_{t}}(\w|\wt)+\phi_{t}(\w)$,
we find that $\grad\psi_{\tpp}(\wtpp)=\grad\psi_{t}(\wt)-g_{t}$,
so unrolling the recursion and solving for $\wtpp$ yields
$\wtpp=\grad\psi_{\tpp}^{*}(-g_{1:t})$, where $\psi_{\tpp}^{*}$ is the Fenchel
conjugate of $\psi_{\tpp}$.
This latter expression is equivalent to the \emph{Follow-the-Regularized-Leader} (FTRL) update
$\wtpp=\argmin_{\w\in\R^{d}}\inner{g_{1:t},\w}+\psi_{\tpp}(\w)$ \citep{mcmahan2017survey}. Moreover, in the constrained setting, letting $\psi_{\tpp,W}(\w)$ denote the
restriction of $\psi_{\tpp}$ to constraint set $W$,
\Cref{alg:centered-md} captures both the ``greedy projection'' update
$\wtpp=\grad\psi_{\tpp,W}^{*}(\grad\psi_{t}(\wt)-g_{t})$
and the
``lazy projection'' update $\wtpp=\grad \psi_{\tpp,W}^{*}(-g_{1:t})$ by adding
the indicator function
$\cI_{W}(\w)$ to the $\varphi_{t}$ terms or to the $\psi_{t}$ terms respectively.
Hence, including $\Delta_t(w)$ in \Cref{alg:centered-md} incorporates some properties of FTRL into a mirror
descent framework.

In the unconstrained setting, the function $\Delta_{t}(\w)$ is a
critical feature of the update.
Indeed, \citet{orabona2018scale} showed that adaptive mirror descent
algorithms can incur \emph{linear} regret in settings
where the divergence $D_{\psi_{t}}(\cdot|\cdot)$ may be unbounded. The
issue is that vanilla mirror descent doesn't properly account for changes in the regularizer $\psi_t$, allowing
the iterates $\wt$ to travel away from their initial position $\w_{1}$ too quickly; \Cref{alg:centered-md}
fixes this by adding a corrective penalty $\Delta_{t}(\w)$ related to how much  $\psi_{t}$ has
changed between rounds. Since this penalty acts to bias the iterates back towards some central reference point $\w_{1}$, we refer to \Cref{alg:centered-md} as \textit{Centered
  Mirror Descent}.

Our approach is similar to
\textit{dual-stabilized mirror descent} (DS-MD),
recently proposed by
\citet{fang2020online}, which employs the update
$\wtpp=\argmin_{\w\in\R^{d}}\gamma_{t}\big(\inner{\eta_{t}g_{t},\w}+D_{\psi}(\w|\wt)) +(1-\gamma_{t})D_{\psi}(\w|\w_{1})$
for scalars $\gamma_{t}\in(0,1)$.
This prevents the
iterates $\wt$ from moving too far from $\w_{1}$
by decaying the dual representation of $\wt$ towards that of $\w_{1}$.
The DS-MD approach
considers only $\psi_t$ of the form $\psi_t = \frac{\psi}{\eta_t}$ for a fixed $\psi$, whereas
Centered Mirror Descent applies more generally to
$\psi_{t}$. This property is crucial for our purposes, as
the $\psi_t$s we employ cannot
be captured by a linear scaling of a fixed underlying $\psi$. One could view our approach as a generalization of \citet{fang2020online} that easily captures a variety of
applications, such as dynamic regret, composite losses, and implicit updates.
The following Lemma provides the generic regret decomposition that we'll
use throughout the rest of this work.
\\
\begin{restatable}{mylemma}{CenteredMDLemma}\label{lemma:centered-md}
  \textbf{(Centered Mirror Descent Lemma)}
  Let $\psi_{t}(\cdot)$ be an arbitrary sequence of differentiable non-negative
  convex functions, and assume that $\w_{1}\in\argmin_{\w\in\R^{d}}\psi_{t}(\w)$
  for all $t$. Let $\varphi_{t}(\cdot)$ be an arbitrary sequence of
  sub-differentiable non-negative convex functions.
  Then for any $\cmp_{1},\ldots,\cmp_{T}$,
  \Cref{alg:centered-md} guarantees
  \begin{align}
    R_{T}(\vec{\cmp})
    &\le
    \psi_{T+1}(\cmp_{T})+\sumtT\varphi_{t}(\cmp_{t})+\sum_{t=1}^{T-1}\underbrace{\inner{\grad \psi_{\tpp}(\wtpp),\cmp_{\tpp}-\cmp_{t}}}_{=:\rho_{t}}\nonumber\\
  &\qquad
    +\sumtT
    \underbrace{\inner{g_{t},\wt-\wtpp}-D_{\psi_{t}}(\wtpp|\wt)-\phi_{t}(\wtpp)}_{=:\delta_{t}},\label{eq:centered-md-regret}
  \end{align}
  where $g_{t}\in\partial\ell_{t}(\wt)$ and $\phi_{t}(\w)=\Delta_{t}(\w)+\varphi_{t}(\w)$.
\end{restatable}
Proof of this Lemma can be found in \Cref{app:centered-md}.
The proof follows as a special case of a regret \textit{equality}
which we derive in \Cref{app:strong-centered-md}.
To build intuition for how to use the Lemma, consider the static regret of
\Cref{alg:centered-md} with
$\varphi_{t}(\w)\equiv0$.
In this case,
\Cref{eq:centered-md-regret} becomes
$R_{T}(\cmp)\le \psi_{T+1}(\cmp)+\sumtT\delta_{t}$.
Now, to guarantee a parameter-free bound of the form
$R_{T}(\cmp)\le \tilde O\big(\norm{\cmp}\sqrt{T}\big)$
for all $\cmp$, a natural approach
is to set $\psi_{T+1}(\cmp)=\widetilde O\big(\norm{\cmp}\sqrt{T}\big)$, and
then focus our efforts on controlling the stability terms
$\sumtT\delta_{t}$. To this end, the following Lemma (proven in \Cref{app:md-stability})
provides a set of simple conditions for bounding an expression closely related to $\delta_{t}$: 
\begin{restatable}{mylemma}{MDStability}\label{lemma:md-stability}
  \textbf{(Stability Lemma)}
  Let $\Psi_{t}:\R_{\ge 0}\to\R_{\ge0}$ be a twice
  differentiable, three-times subdifferentiable function such that
  $\Psi_t'(x)\ge 0$, $\Psi_t''(x)\ge 0$, and $\Psi_{t}'''(x)\le 0$
  for all $x> 0$.
  Let $G_{t}\ge\norm{g_{t}}$ and $\eta_{t}:\R_{\ge 0}\to\R_{\ge0}$ be a $1/G_{t}$ Lipschitz convex
  function, and assume there is an $x_{0}\ge 0$ such that
  $\abs{\Psi_{t}'''(x)}\le \frac{\eta_{t}'(x)}{2}\Psi_{t}''(x)^{2}$ for all
  $x> x_{0}$. Then with $\psi_t(w)=\Psi_t(\|w\|)$, for all $w_t$, $w_{t+1}$:
  \begin{align*}
    \hat\delta_{t}\defeq\inner{g_{t},\wt-\wtpp}-D_{\psi_{t}}(\wtpp|\wt)-\eta_{t}(\norm{\wtpp})\norm{g_{t}}^{2}\le \frac{2\norm{g_{t}}^{2}}{\Psi_{t}''(x_{0})}
  \end{align*}
\end{restatable}
To see the utility of  \Cref{lemma:md-stability}, observe that the only difference
between the
$\delta_{t}$ of \Cref{lemma:centered-md} and the $\hat\delta_{t}$
of \Cref{lemma:md-stability} is that in the former
has a $-\phi_{t}(\wtpp)$ where the latter has a
$-\eta_{t}(\norm{\wtpp})\norm{g_{t}}^{2}$.
Our approach throughout this work will be to
design $\phi_{t}(\w)$  to satisfy
$\phi_{t}(\w)\ge \eta_{t}(\norm{\w})\norm{g_{t}}^{2}$
for all $\w\in\R^{d}$ so that $\delta_{t}\le\hat\delta_{t}$, and then apply the Stability Lemma
to get
$\sumtT\delta_{t}\le \sumtT\hat\delta_{t}\le\sumtT\frac{2\norm{g_{t}}^{2}}{\Psi_{t}''(x_{0})}$.
Then, we design $\Psi_{t}(\cdot)$ to
ensure $\sumtT\frac{2\norm{g_{t}}^{2}}{\Psi_{t}''(x_{0})}\le O(1)$, leading to small regret.

  In the sections to follow we will see several examples
  of $\psi_{t}$ which meet the conditions of the Stability Lemma, but for
  concreteness let us consider as a simple demonstration the fixed function
  $\psi_{t}(\w)=\Psi(\norm{\w})=2\int_{0}^{\norm{\w}}\frac{\Log{x/\eta+1}}{\eta}dx$
  where $\eta\le\frac{1}{G}$.
  Careful calculation shows that $\Psi(\cdot)$ satisfies the conditions of
  \Cref{lemma:md-stability} with $\eta_t(\x)=\eta x$. Hence,
  $\hat\delta_{t}\le\frac{2\norm{g_{t}}^{2}}{\Psi_{t}''(0)}=2\eta^2\norm{g_{t}}^{2}$.
  Now, we wish to achieve $\phi_t(w_{t+1})\ge \eta_t(\|w_{t+1}\|)\|g_t\|^2$. This is
  easily accomplished by setting $\varphi_t(w) = 2\eta^2 \|g_t\|^2 \|w\|$. Thus,
  setting $\eta=O(1/\sqrt{T})$ yields
  $ \sumtT\delta_t\le \sumtT\hat\delta_t = \sumtT 2\eta^2 \|g_t\|^2\le O(1)$ so that overall we would achieve a regret of $\tilde O(\|u\|\sqrt{T})$.
  
  This example demonstrates the purpose of $\varphi_{t}$ in the update. When
  $\Delta_t(w_{t+1})\ge \eta_t(\|w_{t+1}\|)\|g_t\|^2$, we already obtain
  $\delta_t \le \hat \delta_t$. However, this identity may be false (as in the
  previous example) or difficult to prove.\footnote{Proving an analogous
    identity is the principle technical challenge in  deriving FTRL-based parameter-free algorithms.} In such cases, we include a small additional $\varphi_{t}$ term to easily ensure the desired bounds. In fact, this strategy can be viewed as generalizing a certain ``correction'' term which appears in the experts literature (e.g. \cite{steinhardt2014eg, chen2021impossible}), but to our knowledge is not typically employed in the general online linear optimization setting.

\section{\secPF}\label{sec:pf}

As a warm-up, we first use our framework to construct a parameter-free
algorithm which achieves the optimal static regret (\Cref{eqn:paramfreebound}).
\begin{restatable}{mytheorem}{AdaptiveSelfStabilizing}\label{thm:adaptive-self-stabilizing}
  Let $\ell_{1},\ldots,\ell_{T}$ be $G$-Lipschitz convex functions and $g_{t}\in\partial\ell_{t}(\wt)$ for
  all $t$. Let $\epsilon >0$,
  $V_{t}= 4 G^{2} +\norm{g}_{1:\tmm}^{2}$,
  $\alpha_{t}=\frac{\epsilon G}{\sqrt{V_{t}}\log^{2}(V_{t}/G^{2})}$,
  and set
  \(
    \psi_{t}(\w)
    =
      3\int_{0}^{\norm{\w}}\min_{\eta\le\frac{1}{G}}\sbrac{\frac{\Log{x/\alpha_{t}+1}}{\eta}+\eta V_{t}}dx
  \).
  Then
  for all $\cmp\in\R^{d}$, \Cref{alg:centered-md} guarantees
  \begin{align*}
    R_{T}(\cmp)
    &\le
      \Ohat\brac{G\epsilon + \norm{\cmp}\sbrac{\sqrt{\norm{g}_{1:T}^{2}\Log{\frac{\norm{\cmp}\sqrt{\norm{g}_{1:T}^{2}}}{\epsilon G}+1}}\maxOp G\Log{\frac{\norm{\cmp}\sqrt{\norm{g}_{1:T}^{2}}}{\epsilon G}+1}}}
  \end{align*}
  where $\Ohat(\cdot)$ hides constant and $\log(\log)$ factors (but not $\log$ factors).
\end{restatable}
The full proof can be found in \Cref{app:pf}, along with an efficient closed-form update formula. It follows the intuition
developed in the previous section: \Cref{lemma:centered-md} implies
$R_{T}(\cmp)\le\psi_{T+1}(\cmp)+\sumtT\delta_{t}$.
Then, we show that
$\psi_{t}$ satisfies the conditions of \Cref{lemma:md-stability}
while
the growth rate $\Delta_{t}(\w)$
ensures that $\delta_{t}\le\hat \delta_{t}$,
so that
$\delta_{t}\le\hat\delta_{t}\le O\brac{\frac{2\norm{g_{t}}^{2}}{\Psi_{t}''(\alpha_{t})}}\le O(\frac{\alpha_{t}\norm{g_{t}}^{2}}{\sqrt{V_{t}}})$.
Finally, we choose $\alpha_{t}$ small enough to ensure $\sumtT\delta_{t}\le O(1)$.

Treating $\log(\log)$ terms as effectively constant, the bound in
\Cref{thm:adaptive-self-stabilizing} achieves the ``ideal'' dependence
on $V_{T}$ in the logarithmic factors. Indeed, given oracle access to
$\norm{\cmp}$ and $V_{T}$, we could set
$\epsilon=O\Big(\frac{\norm{\cmp}\sqrt{V_{T}}}{G}\Big)$, causing all the $\log$
terms to disappear from the bound and leaving only
$R_{T}(\cmp)\le \Ohat\big(\norm{\cmp}\sqrt{V_{T+1}}\big)$, which matches the optimal
rate vanilla gradient descent would achieve with
oracle tuning up to a $\log(\log)$ factor.

\section{Dynamic Regret}\label{sec:dynamic}

Now that we've had a taste of how to use our techniques,
we turn to the challenging problem of
competing with an arbitrary \textit{sequence} of comparators $\cmp_{t}$.
To appreciate why this is difficult using existing techniques,
let us recall
the \textit{reward-regret duality}, a key player in most analyses of parameter-free algorithms.
Suppose that we wish to guarantee
static regret of $R_{T}(u)=\sumtT\inner{g_{t},\wt-\cmp} \le B_{T}(\cmp)$
for all $\cmp\in\R^{d}$ for some function $B_T$.
Because this holds for \textit{all} $\cmp\in\R^{d}$,
we must have:
\begin{align*}
  \sumtT\inner{g_{t},\wt}+\sup_{\cmp\in\R^{d}}\inner{-g_{1:T},\cmp}-B_{T}(\cmp)=\sumtT\inner{g_{t},\wt} + B_{T}^{*}(-g_{1:t})\le0,
\end{align*}
where $B_{T}^{*}(\cdot)$ denotes the Fenchel conjugate of $B_{T}(\cdot)$.
Rearranging, we find that guaranteeing
$R_{T}(\cmp)\le B_{T}(\cmp)$ for all $\cmp\in\R^{d}$
is equivalent to guaranteeing
$\sumtT\inner{-g_{t},\wt}\ge B_{T}^{*}(-g_{1:T})$. This latter expression
has no dependence on the \textit{unknown} comparator $\cmp$, making it
appealing from an algorithm design perspective. Hence,
many prior works revolve around
designing algorithms which ensure that
$\sumtT\inner{-g_{t},\wt}\ge B_{T}^{*}(-g_{1:T})$ for some $B_{T}^{*}$ of interest.

However, notice that the assumption of a \textit{fixed} comparator
$\cmp\in\R^{d}$ was vital for the argument above to work.
It is unclear
what the analogue of this argument should be for dynamic regret, where we instead have
a \textit{sequence} of comparators.
In fact, in the next section, we show that most algorithms designed using this duality cannot guarantee
sublinear dynamic regret.
Then, in \Cref{sec:dynamic-alg}, we will remedy this issue
via our mirror-descent framework.

\subsection{Lower Bounds}\label{sec:lowerbounds}
In this section, we show that common algorithm design strategies for the unconstrained setting cannot achieve optimal dynamic regret. Specifically, we consider 1-D OLO algorithms such that: 
{\small
  \vspace{-2ex}
  \begin{singlespace}
  \begin{enumerate}
      \item The algorithm sets $w_{t+1} = F_t(g_{1:t}, \|g_1\|,\dots,\|g_t\|)$ for some functions $F_1,\dots,F_t$. We will frequently elide the dependence on $\|g_t\|$ to write $F_t(g_{1:t})$.\label{propertyone}
      \item $F_t$ satisfies $\sign(F_t(g_{1:t}))=-\sign(g_{1:t})$.\label{propertytwo}
      \item $F_t$ is odd: $F_t(g_{1:t}) = -F_t(-g_{1:t})$.\label{propertythree}
      \item $F_t(-x)$ is non-decreasing for positive $x$.\label{propertyfour}
  \end{enumerate}
  \end{singlespace}
}
\noindent Notice that the vast majority of parameter-free FTRL algorithms satisfy these properties with $F_t=\nabla \psi_t^\star$. First, we consider the \emph{constrained setting}. Here, it is actually relatively easy to show that no algorithm satisfying these conditions can obtain low dynamic regret:
\begin{restatable}{mytheorem}{FTRLConstrainedLB}\label{thm:ftrl-constrained-lb}
Suppose an algorithm $\cA$ satisfies the conditions [\ref{propertyone}, \ref{propertytwo}, \ref{propertythree}, \ref{propertyfour}] for a 1-d OLO game with domain $[-1,1]$. Then for all even $T$ there exists a sequence of costs $g_1,\dots,g_T$ with $\|g_t\|=1$ for all $t$ and a comparator sequence $u_1,\dots,u_T$ such that the path length $P=\sum_{t=1}^{T-1}\|u_t-u_{t+1}\|\le 2$, but the regret is at least $T/2$.
\end{restatable}
\begin{proof}
Set $g_t=1$ for $t\le T/2$ and $g_t=-1$ otherwise. Notice that $g_{1:t}\ge 0$ for all $t$, so that $-1\le w_t\le 0$ for all $t$. Thus $\sum_{t=1}^T w_t g_t\ge \sum_{t=1}^{T/2} w_tg_t\ge -T/2$. Let $u_t=-1$ for $t\le T/2$ and $u_t=1$ otherwise. Then clearly $P=2$, and $\sum_{t=1}^T g_tu_t=-T$ for a total regret of at least $T/2$.
\end{proof}
The essential idea behind this result is that when presented with a sequence of alternating runs of $+1$ and $-1$, we have $g_{1:t}\ge 0$ for all $t$ so that $w_t\le 0$ for all $t$. Intuitively, this means that we cannot compete with the competitor when $g_t=-1$ and $u_t\ge 0$. 

To formally establish bounds in the unconstrained setting is a bit more
complicated. In this case, we must guard against the possibility that the
algorithm experiences some significant \emph{negative regret} during the periods
in which $g_t$ has the opposite sign of $w_t$. This is not an issue in the
constrained setting because the algorithm clearly cannot have less than $-1$
loss on any given round. In the unconstrained setting however, the loss is in
principle unbounded. Thus, we consider the maximum possible value of
$\regret_T(0)$ as a measure of ``complexity'' of the algorithm. The following
Lemma~\ref{lem:boundediterate} provides an constraint on the algorithm in terms
of this complexity. Then, by carefully tuning our adversarial sequence to this
complexity measure, we are able to guarantee poor dynamic regret.
Proofs of the following two results can be found in \Cref{app:dynamic}.
\begin{restatable}{mylemma}{FTRLBoundedIterate}\label{lem:boundediterate}
Suppose an algorithm $\cA$ satisfying the conditions [\ref{propertyone},
\ref{propertytwo}, \ref{propertythree}, \ref{propertyfour}] also guarantees that $\sum_{i=1}^t g_i w_i \le \epsilon$ for all $t$ for some $\epsilon$ if $\|g_i\|=1$ for all $i$. Then there is a universal constant $C$ (not depending on $\cA$) such that for any sequence $g_1,\dots,g_t$ satisfying $\|g_i\|=1$ for all $i\le t$ and $\|g_{1:t}\|\le C\sqrt{t}/2$, we have $|w_{t+1}|\le \frac{2 \epsilon }{C^2}$.
\end{restatable}
Using this Lemma, we can show our lower bound:
\begin{restatable}{mytheorem}{FTRLUnconstrainedLB}\label{thm:ftrl-unconstrained-lb}
Let $C$ be the universal constant from Lemma~\ref{lem:boundediterate}. Suppose
an algorithm $\cA$ satisfying the conditions [\ref{propertyone},
\ref{propertytwo}, \ref{propertythree}, \ref{propertyfour}] also guarantees
$\sum_{i=1}^t g_i w_i \le \epsilon$ for all $t$ for some $\epsilon$ if $\|g_i\|=1$
for all $i$.
Then there is a universal $\widetilde{T}$ such that for all $T\ge \widetilde{T}$,
there exists a sequence $\{g_t\}$ and a comparator sequence $\{u_{t}\}$ that does not depend on $\cA$ such that:
\begin{align*}
    P_T+\max_t |u_t|&\in\left[ \frac{\epsilon\sqrt{2T}}{C^3},\  \frac{8\epsilon\sqrt{T}}{C^3}\right]\\
    \sum_{t=1}^T g_t(w_t-u_t)&\ge \frac{C}{16} (P_T+\max_t |u_t|) \sqrt{T}
\end{align*}
\end{restatable}
Theorem \ref{thm:ftrl-unconstrained-lb} essentially shows that all FTRL-based
algorithms we are aware of cannot achieve a dynamic regret better than
$O(P_T \sqrt{T})$. It should be noted that there are a few algorithms (e.g. the ONS-based betting algorithm of \cite{cutkosky2018black}) that are not captured by the conditions imposed on the
Algorithm in Theorem~\ref{thm:ftrl-unconstrained-lb}. However, we believe all such previously known exceptions satisfy the constraint that $w_t\le 0$ for all $t$ when presented with alternating signs of gradients as in our lower bound constructions. We therefore hypothesize that they also fail to achieve optimal dynamic regret, but we leave open establishing formal bounds.

Finally, we stress that the adversarial sequence of \Cref{thm:ftrl-unconstrained-lb} is \emph{algorithm independent}. Thus, one likely cannot hope to ``combine'' several such suboptimal algorithms into an optimal algorithm as is done in the constrained setting: all the suboptimal algorithms could be \emph{simultaneously bad}.

\subsection{\secDynamicAlg}\label{sec:dynamic-alg}

The results in the previous section suggest that
any algorithm which updates using an FTRL-like update
of the form $\wtpp=F_{t}(g_{1:t})$ will be unable
to guarantee dynamic regret with a sublinear dependence on the
path-length $P_{T}$ due to an excessive resistance to
changing direction. In what follows, we'll show that
our additional $\varphi_{t}$ term can mitigate this issue by biasing the iterates ever-so-slightly back towards
the origin. This facilitates a more rapid change of sign when the
losses change direction, and enables us to avoid the pathogenic behavior
observed in the previous section.

Our approach is as follows. Using the tools developed in \Cref{sec:centered-md},
we'll first derive an algorithm which, for any $\eta\le \frac{1}{G}$, guarantees
$R_{T}(\cmp)\le \Otilde\Big(\frac{P_{T}+\max_t\|\cmp_t\|}{\eta}+\eta\sumtT\norm{g_{t}}^{2}\norm{\cmp_{t}}\Big)$. Note that such a bound is out-of-reach for algorithms covered by \Cref{thm:ftrl-unconstrained-lb}.
Further, for all $\eta$ we have $R_{T}(\zeros)=O(1)$.
Hence, following \cite{cutkosky2019combining},
we run an instance of this algorithm $\cA_{\eta}$ for each $\eta$ in some set $\cS=\Set{\eta\in\R: 0< \eta\le\frac{1}{G}}$, and on each round
we play $\wt=\sum_{\eta\in\cS}\wt^{\eta}$
where $\wt^{\eta}$ is the output of $\cA_{\eta}$.
Then for any arbitrary
$\widetilde{\eta}\in\cS$, we can write
$\inner{g_{t},\wt-\cmp_{t}}=\inner{g_{t},\wt^{\widetilde{\eta}}-\cmp_{t}}+\sum_{\eta\ne\widetilde{\eta}}\inner{g_{t},\wt^{\eta}}$,
so the regret is bounded as
\begin{align*}
  R_{T}(\vec{\cmp})
  \le\sumtT\inner{g_{t},\wt^{\widetilde{\eta}}-\cmp_{t}}+\sum_{\eta\ne\widetilde{\eta}}\Big[\sumtT\inner{g_{t},\wt^{\eta}}\Big]
    &=
      R_{T}^{\widetilde{\eta}}(\vec{\cmp})+\sum_{\eta\ne\widetilde{\eta}}R_{T}^{\eta}(\zeros)
      \le O(R_{T}^{\widetilde{\eta}}(\vec{\cmp})+\abs{\cS}).
\end{align*}
Since this holds for an \textit{arbitrary} $\widetilde{\eta}\in\cS$, it must
hold for the $\eta\in\cS$ for which $R_{T}^{\eta}(\vec{\cmp})$ is smallest,
so we need only ensure that there is \textit{some} near-optimal $\eta\in\cS$, and that $|\cS|$ is not too large, which is easily accomplished by setting $|\cS|$ to be a logarithmically-spaced grid.
These algorithms $\cA_{\eta}$ and their corresponding
regret guarantee are given in
the following proposition.
\begin{restatable}{myproposition}{DynamicFixedEta}\label{prop:dynamic-fixed-eta}
  Let $\ell_{1},\ldots,\ell_{T}$ be $G$-Lipschitz convex functions and
  $g_{t}\in\partial\ell_{t}(\wt)$ for all $t$.
  Let $\epsilon > 0$, $V_{t}=4G^{2}+\norm{g}_{1:\tmm}^{2}$,
  $\alpha_{t}=\frac{\epsilon G^{2}}{V_{t}\log^{2}(V_{t}/G^{2})}$,
  and set
  \(
  \psi_{t}(\w)=2\int_{0}^{\norm{\w}}\frac{\Log{x/\alpha_{t}+1}}{\eta}dx
  \) and
  \(
  \varphi_{t}(\w)=2\eta\norm{g_{t}}^{2}\norm{\w}.
  \)
  Then
  for any $\cmp_{1},\ldots,\cmp_{T}$ in $\R^{d}$, \Cref{alg:centered-md} guarantees
  \begin{align*}
    R_{T}(\vec{\cmp})
    &\le
      \Ohat\Bigg(
    \epsilon+ \frac{\brac{M+P_{T}}\Big[\Log{\frac{MT^{2}\norm{g}_{1:T}^{2}}{\epsilon G^{2}}+1}\maxOp 1\Big]}{\eta}
      +\eta\sumtT\norm{g_{t}}^{2}\norm{\cmp_{t}}
      \Bigg),
  \end{align*}
  where $M=\max_t \|\cmp_t\|$ and $\Ohat(\cdot)$ hides constant and $\log(\log)$ factors.
\end{restatable}
The proof can be found in \Cref{app:dynamic-fixed-eta}, and
again follows the intuition in \Cref{sec:centered-md}:
we first apply
\Cref{lemma:centered-md} to get
$R_{T}(\vec{\cmp})\le \psi_{T+1}(\cmp_{T})+\sum_{t=1}^{T-1}\rho_{t}+\sumtT\varphi_{t}(\cmp_{t})+\sumtT\delta_{t}$.
Unlike in \Cref{sec:pf}, $\Delta_{t}(\w)$ is generally not
large enough to ensure that $\delta_{t}\le\hat\delta_{t}$. Instead, we include
an additional composite regularizer $\varphi_{t}$ in the update
and show that this now ensures $\delta_{t}\le \hat\delta_{t}$, so that by \Cref{lemma:md-stability} we have
$\delta_{t}\le\hat\delta_{t}\le\frac{2\norm{g_{t}}^{2}}{\Psi_{t}''(0)}\le 2\eta\alpha_{t}\norm{g_{t}}^{2}$.
Then we choose $\alpha_{t}$ to be small enough to ensure that
$\sumtT\delta_{t}\le O(1)$. 
We also now need to control the additional terms associated with the time-varying
comparator, $\rho_{t}=\inner{\grad\psi_{\tpp}(\wtpp),\cmp_{t}-\cmp_{\tpp}}$.
To handle these, we again exploit $\varphi_t$: by increasing it slightly more, we
can decrease $\delta_t$ enough to cancel out the additional $\rho_t$.

\begin{figure*}
  \vspace{-6ex}
\begin{algorithm}[H]
  \SetAlgoLined
  \KwInput{Lipschitz bound $G$, value $\eps>0$,
    step-sizes
    $\cS=\Set{\frac{2^{k}}{G\sqrt{T}}\minOp\frac{1}{G}:1\le k\le \lceil\log_{2}{\sqrt{T}}\rceil}$}\\
  \KwInitialize{
    $\epsilon=\frac{\eps}{\abs{\cS}}=\frac{\eps}{\lceil\log_{2}(\sqrt{T})\rceil}$,
    $V_1=4G^2$, $\w_{1}^{\eta}=\zeros$ and $\theta_{t}^{\eta}=\zeros$
    for each $\eta\in\cS$}\\
  \For{$t=1:T$}{
    Play $\wt=\sum_{\eta\in\cS}\wt^{\eta}$, receive subgradient $g_{t}$\\
    Update $V_{t+1} = V_t + \|g_t\|^2$ and $\alpha_{t+1} = \frac{\epsilon G^2}{V_{t+1}\log^2(V_{t+1}/G^2)}$\\
    \For{$\eta\in\cS$}{
    Set $\theta_t^\eta = \frac{2w_t^\eta\log(\|w_t^\eta\|/\alpha_t+1)}{\eta\|w_t^{\eta}\|}-g_t$ \qquad(with $\theta_t^\eta = -g_t$ if $w_t^\eta=\zeros$)\\
    Update $\wtpp^\eta =\frac{\alpha_{t+1}\theta_t^\eta}{\|\theta_t^\eta\|}\left[\exp\left[\frac{\eta}{2}\max(\|\theta_t^\eta\|-2\eta\|g_t\|^2,0)\right]-1\right]$\\

    }
  }
  \caption{Dynamic Regret Algorithm}
  \label{alg:dynamic-meta}
\end{algorithm}
\end{figure*}
With this result in hand, we proceed to ``tune'' the optimal step-size
by simply adding the iterates of a collection of these simple
learners $\cA_{\eta}$, as discussed above. The full algorithm is given in \Cref{alg:dynamic-meta},
and the overall regret guarantee is given in \Cref{thm:dynamic-combined-main}
(with proof in \Cref{app:dynamic-combined}).
\newpage
\begin{restatable}{mytheorem}{DynamicCombinedMain}\label{thm:dynamic-combined-main}
  For any $\eps>0$ and $\cmp_{1},\ldots,\cmp_{T}$ in $\R^{d}$, \Cref{alg:dynamic-meta} guarantees
  \begin{align*}
    R_{T}(\vec{\cmp})
    &\le
      \Ohat\Bigg(\eps+\sqrt{(M+P_{T})\sumtT\norm{g_{t}}^{2}\norm{\cmp_{t}}\Log{\frac{MT^{2}\norm{g}_{1:T}^{2}}{\eps G^{2}}+1}}
      +P_{T}\Log{\frac{MT^{2}\norm{g}_{1:T}^{2}}{\eps G^{2}}+1}
      \Bigg)
  \end{align*}
  where $M=\max_t\|\cmp_t\|$ and $\Ohat(\cdot)$ hides constant and $\log(\log)$ factors.
\end{restatable}
The bound achieved by \Cref{alg:dynamic-meta} is the first
non-vacuous dynamic regret guarantee of any kind that we are aware of in unbounded domains.
Further, \Cref{thm:dynamic-combined-main}
exhibits a stronger \emph{per-comparator adaptivity} than previously obtained by depending on the individual comparators $\norm{\cmp_{t}}$,
in contrast to the $R_{T}(\vec{\cmp})\le \Otilde\Big(\sqrt{(M^{2}+MP_{T})\sumtT\norm{g_{t}}^{2}}\Big)$
rate attained by prior works in bounded domains \citep{zhang2018adaptive,jadbabaie2015online}.

To see why this per-comparator adaptivity is interesting, let us
consider a learning scenario in which there is a nominal ``default'' decision $\overline{\cmp}$
which we expect to perform well \textit{most} of the time, but may
perform poorly during certain rare/unpredictable events.
One example of such a situation is when one has access to an batch of data collected \textit{offline}, which
we can leverage to fit a parameterized model $\cM(\overline{\cmp})$ to the data to use as a baseline predictor.
Deploying such a model online can be dangerous in practice because there may be
certain events that are poorly covered by our dataset,
leading to unpredictable behavior from the model.
In this context, we can think of $\overline{\cmp}$ as the learned model
parameters, and without loss of generality we can assume $\overline{\cmp}=\zeros$ (since otherwise
we could just translate the decision space to be centered at $\overline{\cmp}$).
In this context, \Cref{thm:dynamic-combined-main} tells us that \Cref{alg:dynamic-meta}
will accumulate \textit{no regret} over any intervals where we would want to
compare performance against the baseline model,
and over any intervals $[a,b]$ where the model is a poor comparison
we are still guaranteed to accumulate no more than a
$\Otilde\Big(\sqrt{(M^{2}+M P_{[a,b]})\norm{g}_{a:b}^{2}}\Big)$ penalty,
where
$P_{[a,b]}=\sum_{t=a+1}^{b}\norm{\cmp_{t}-\cmp_{\tmm}}$ is the path-length of
any other arbitrary sequence of comparators over the interval $[a,b]$.

The property in the preceeding discussion is similar to
the notion of \textit{strong adaptivity} in the \emph{constrained} setting, in which an algorithm
guarantees the optimal static regret over all sub-intervals of $[1,T]$
\textit{simultaneously} \citep{jun2017improved,daniely2015strongly}. One might wonder if instead we should hope for the natural analog in the unconstrained setting: $R_{[a,b]}(\cmp) = \sum_{t=a}^b \langle g_t, \wt - \cmp\rangle \le \tilde O(\|\cmp\|\sqrt{b-a})$ for all $[a,b]$.
Unfortunately, this natural analog is likely unattainable.
To see why, notice that for all intervals $[a,b]$ of some fixed length $\tau=b-a$, we would require
$R_{[a,b]}(\zeros)=\sum_{t=a}^{b}\inner{g_{t},\wt}\le O(1)$, suggesting that
no $\wt$ can be larger than some fixed constant (dependent on $\tau$).
Yet clearly for large enough $T$
this can't be guaranteed while simultaneously guaranteeing
$R_{T}(\cmp)\le O\big(\norm{\cmp}G\sqrt{T\Log{\norm{\cmp}GT}}\big)$ for all $\cmp\in\R^{d}$,
since via reward-regret duality this entails
competing against a fixed comparator $\cmp\in\R^{d}$ with
$\norm{\cmp}=O\big(\Exp{T}/T\big)$ in the worst-case, which can get arbitrarily
large as $T$ increases.
For this reason, we consider
\Cref{thm:dynamic-combined-main} to be a suitable relaxation of the strongly adaptive guarantee for unbounded domains.

Interestingly, if one is willing to forego adaptivity
to the individual $\norm{\cmp_{t}}$, we show in \Cref{app:dynamic-redux}
that achieving the weaker
$R_{T}(\vec{\cmp})\le \Otilde\Big(\sqrt{(M^{2}+MP_{T})\norm{g}_{1:T}^{2}}\Big)$ can
be attained in unbounded domains using the one-dimensional
reduction of \citet[Algorithm 2]{cutkosky2018black}.

\section{\secImplicit}
\label{sec:implicit}

A useful consequence of our mirror descent formulation is that
we can easily incorporate the entire loss function
$\ell_{t}(\cdot)$ rather than
the linear proxy $w\mapsto\inner{\grad\ell_{t}(\wt),\w}$
used in the usual mirror descent update. Mirror descent updates
incorporating $\ell_{t}(\cdot)$ in their update are called \textit{implicit},
because setting $\wtpp=\argmin_{\w\in\R^{d}}\ell_{t}(\w)+D_{\psi}(\w|\wt)$
leads to an equation of the form
$\wtpp=\grad\psi^{*}(\grad\psi(\wt)-\grad\ell_{t}(\wtpp))$, which must be
solved for $\wtpp$ to obtain the update.

Implicit updates are appealing in practice because they enable one
to more directly incorporate known properties of the loss functions or
additional modeling assumptions
to improve convergence rates \citep{asi2019stochastic}.
Moreover, in bounded domains there may be advantages even
without any additional assumptions on the loss functions. Indeed,
\citet{campolongo2020temporal} recently developed an
implicit mirror descent which guarantees
$R_{T}(\cmp)\le O\Big(\min\Big\{\sqrt{\|g\|_{1:T}^{2}},\cV_{T}\Big\}\Big)$
where $\cV_{T}=\sum_{t=2}^{T}\sup_{x\in\cX}\ell_{t}(\x)-\ell_{\tmm}(\x)$ is the
\textit{temporal variability} of the loss sequence.
This bound has the appealing property that
$R_{T}(\cmp)\le O(1)$ when the loss functions are fixed $\ell_{t}(\cdot)=\ell(\cdot)$.

\begin{figure*}
  \vspace{-4ex}
\begin{algorithm}[H]
  \SetAlgoLined
  \KwInput{Initial regularizer $\psi_{1}:\R^{d}\to\R_{\ge 0}$, initial $\widehat\ell_{1}(\cdot)$}\\
  \KwInitialize{$\x_{1}=\argmin_{x}\psi_{1}(x)$, $\w_{1}=\argmin_{w}\hat\ell_{1}(w)+D_{\psi_{1}}(w|x_{1})$}\\
  \For{$t=1:T$}{
    Play $\wt$, observe loss function $\ell_{t}(\cdot)$\\
    Set $g_{t}\in\partial\ell_{t}(\wt)$\\
    Choose functions $\psi_{\tpp}$, $\hat\ell_{\tpp}$, 
    and define
    $\Delta_{t}(\w)=D_{\psi_{\tpp}}(\w|\x_{1})-D_{\psi_{t}}(\w|\x_{1})$\\
    Update
    \(\x_{\tpp}=\argmin_{\x}\inner{g_{t},x}+D_{\psi_{t}}(\x|\x_{t})+\Delta_{t}(\x)
    \)\\
    \(\hphantom{Update}\wtpp = \argmin_{\w}\hat\ell_{\tpp}(\w)+D_{\psi_{\tpp}}(\w|\xtpp)\)
  }
  \caption{Implicit-Optimistic Centered Mirror Descent}
  \label{alg:implicit-optimism}
\end{algorithm}
\vspace{-2ex}
\end{figure*}

In this section we leverage our
mirror descent formulation to
incorporate an additional implicit
update on each step to guarantee
$R_{T}(\cmp)\le\Otilde\big(\norm{\cmp}\sqrt{\sumtT\|\grad\ell_{t}(\wt)-\grad\ell_{\tmm}(\wt)\|^{2}}\big)$,
which can be significantly smaller than the usual
$R_{T}(\cmp)\le \Otilde\big(\norm{\cmp}\sqrt{\sumtT\|\grad\ell_{t}(\wt)\|^{2}}\big)$ bound when
the loss functions are ``slowly moving''.
Similar to \citet{campolongo2020temporal}, this bound
guarantees that $R_{T}(\cmp)\le O(1)$ when the loss functions are fixed,
yet our result holds even in unconstrained domains. In fact,
in the setting of Lipschitz losses in unconstrained domains,
the quantity $\sqrt{\sumtT\|\grad\ell_{t}(\wt)-\grad\ell_{\tmm}(\wt)\|^{2}}$
is perhaps a more suitable way to achieve this property, since in unbounded domains
$\cV_{T}$ is typically infinite unless $\ell_t-\ell_{t-1}$ is constant.

The only prior method we are aware of to incorporate implicit updates into parameter-free learning was recently developed by \cite{chen2022implicit}. They propose an interesting new regret decomposition and apply it to develop closed-form implicit updates for truncated linear losses. We adopt different goals: without attempting to build efficient closed-form updates, we consider general loss functions and show that implicit updates fall easily out of our mirror-descent formulation.

Our algorithm is derived as a special case of
the algorithm shown in \Cref{alg:implicit-optimism},
which can be understood as an instance of centered mirror descent
with an additional \textit{optimistic} step on each round.
The optimistic step leverages an arbitrary guess $\hat \ell_{\tpp}(\cdot)$
about what the next loss function will be.
Intuitively, if the learner could deduce the trajectory
of the loss functions, they'd be able to ``think ahead''
and play a point $\wtpp$ for which the next loss $\ell_{\tpp}(\cdot)$
is minimized. The following theorem
provides an algorithm which guarantees
$R_{T}(\cmp)\le \Otilde\Big(\norm{\cmp}\sqrt{\sumtT\|\grad\ell_{t}(\wt)-\grad\hat\ell_{t}(\wt)\|^{2}}\Big)$
using an arbitrary sequence of optimistic guesses $\hat\ell_{t}(\cdot)$.
\begin{restatable}{mytheorem}{OptimisticImplicit}\label{thm:optimistic-implicit}
  Let $\ell_{1},\ldots,\ell_{T}$ and $\hat\ell_{1},\ldots,\hat\ell_{T}$ be $G$-Lipschitz convex functions.
  For all $t$,
  let
  \(
    \psi_{t}(\w)=3\int_{0}^{\norm{\w}}\min_{\eta\le \frac{1}{2G}}\Big[\frac{\Log{x/\widehat{\alpha}_{t}+1}}{\eta}+\eta\Vhat_{t}\Big]dx
    \), where
  $\Vhat_{t}=16G^{2}+\sum_{s=1}^{t-1}\|\grad\ell_{s}(\w_{s})-\grad\hat \ell_{s}(\w_{s})\|^{2}$,
  $\widehat{\alpha}_{t}=\frac{\epsilon G}{\sqrt{\Vhat_{t}}\log^{2}(\Vhat_{t}/G^{2})}$, and  $\epsilon>0$.
  Then for all $\cmp\in\R^{d}$, \Cref{alg:implicit-optimism} guarantees
  \begin{align*}
    R_{T}(\cmp)
    &\le
      \Ohat\brac{
    \epsilon G + \norm{\cmp}\sbrac{\sqrt{\Vhat_{T+1}\Log{\frac{\norm{\cmp}\sqrt{\Vhat_{T+1}}}{G\epsilon}+1}}\maxOp G\Log{\frac{\norm{\cmp}\sqrt{\Vhat_{T+1}}}{G\epsilon}+1}}
      },
  \end{align*}
  where $\Ohat(\cdot)$ hides constant and $\log(\log)$ factors.
\end{restatable}
The proof is similar to the proof of \Cref{thm:adaptive-self-stabilizing}, with some tweaks to account for the optimistic step, and is deferred to
\Cref{app:implicit}.
As an immediate corollary, we have that by setting $\hat\ell_{\tpp}(\w)=\ell_{t}(\w)$, the
regret is bounded as
\(
  R_{T}(\cmp)
  \le
    \Otilde\Big(\norm{\cmp}\sqrt{\sumtT\norm{\grad\ell_{t}(\wt)-\grad\ell_{\tmm}(\wt)}^{2}}\Big)
\).
To our knowledge, bounds of this form have previously only been obtained
in bounded domains \citep{zhao2020dynamic}.

Note that our algorithm only makes use of an implicit update during
the optimistic step.
One could also implement an implicit update in the primary update,
but it is unclear what concrete improvements this would yield in the regret bound in the unbounded setting. We leave this as an exciting direction for future work.

\section{\secScaleFree}\label{sec:scale-free}

The algorithms in the previous sections require \textit{a priori} knowledge of
the Lipschitz constant $G$ to run.
This is unfortunate, as such knowledge may not be known in practice.
An ideal algorithm would instead \textit{adapt} to an unknown Lipschitz constant $G$ on-the-fly, while still
maintaining $R_{T}(\cmp)\le \widetilde O\left(\norm{\cmp}G\sqrt{T}\right)$ static regret.
Unfortunately,
various lower bounds (e.g. \citet{cutkosky2017online, mhammedi2020lipschitz})
show that this goal is not achievable in general, but nevertheless one can make
significant partial progress.

One simple approach to this problem, suggested by \citet{cutkosky2019artificial}, is the
following reduction based on a gradient-clipping approach.
First, we
design an algorithm $\cA$ which achieves suitable regret when given
prescient ``hints'' $h_{t}$ satisfying $h_{t}\ge\norm{g_{t}}$ at the start of
round $t$. In practice, we obviously can not provide such hints because we
have not yet observed $g_{t}$, so instead we pass
our best estimate, $h_{t}=\max_{s<t}\norm{g_{s}}$. Then,
we simply pass $\cA$ \textit{clipped} subgradients
$\gbar_{t}=g_{t}\Min{1,\frac{h_{t}}{\norm{g_{t}}}}$,
which ensures that $h_{t}\ge \norm{\gbar_{t}}$, so that the hint given to $\cA$ is never incorrect.
Finally, the outputs $\wt$ of $\cA$ are constrained to lie in the domains
$W_{t}=\big\{\w\in\R^{d}: \norm{\w}\le\sqrt{\sum_{s=1}^{\tmm}\norm{g_{s}}/G_{s}}\big\}$
where $G_{t}=\max_{\tau\le t}\norm{g_{\tau}}$.
\cite{cutkosky2019artificial} showed that this approach ensures
$R_{T}(\cmp)\le R_{T}^{\cA}(\cmp)+G_{T}\norm{\cmp} +G_{T}\sqrt{\sumtT\norm{g_{t}}/G_{t}} + G_{T}\norm{\cmp}^{3}$, where
$R_{T}^{\cA}(\cmp)$ is the regret of $\cA$ on the losses $\gbar_{t}$, and
\citet{mhammedi2020lipschitz} showed that these additive penalties
are unimprovable, so this bound captures the best-possible compromise.

While this hint-based strategy can be used to mitigate the problem of unknown Lipschitz constant
$G$, a truly ideal algorithm would be \emph{scale-free}. That is, the algorithm's outputs $\wt$ are invariant to any
constant rescaling of the gradients $g_{t}\mapsto c g_{t}$ for all $t$. Scale-free regret bounds scale
with the maximal subgradient \textit{encountered}
$G_{T}=\max_{t\le T}\norm{g_{t}}$,  while non-scale free bounds typically depend on
some user-specified estimate of $G_T$ and may perform much worse if this estimate
is very poor\footnote{Typical parameter-free algorithms rely on an \emph{upper bound} for $G_T$, while \cite{cutkosky2019artificial} implicitly relies on a \emph{lower bound}.}.
\citet{mhammedi2020lipschitz} used the approach
proposed by \citet{cutkosky2019artificial} to develop FreeGrad, the first parameter-free and scale-free algorithm.

Unfortunately, FreeGrad suffers from an analytical difficulty
called the \textit{range-ratio} problem.
Briefly, the range-ratio problem occurs when $h_T/h_1$ (called the range-ratio) is very large: in principle if we set $h_1=\|g_1\|$, then this quantity could grow arbitrarily large, and so even logarithmic dependencies can make the regret  bound vacuous.
In order to circumvent this difficulty,
\cite{mhammedi2020lipschitz} utilize a doubling-based scheme,
restarting FreeGrad whenever a particular technical
condition was met. While such restart strategies only lose a constant factor in the regret in theory, they are unsatisfying: scale-free updates are motivated by potential practical
performance benefits, yet any algorithm which is
forced to restart from scratch several times during deployment is
unlikely to achieve high performance in practice.
The following theorem, proven in \Cref{app:scale-free}, characterizes a new base algorithm that, when combined with the reduction of \citet{cutkosky2019artificial}, generates a
scale-free algorithm which avoids the range-ratio problem
without resorting to restarts. Our approach
employs a simple analysis which follows easily using the tools
developed in \Cref{sec:centered-md}, enabling us to
achieve tighter logarithmic factors than FreeGrad.
\begin{restatable}{mytheorem}{ScaleInvariant}\label{thm:scale-invariant}
  Let $\ell_{1},\ldots,\ell_{T}$ be convex functions and
  $g_{t}\in\partial\ell_{t}(\wt)$
  for all $t$.
  Let $h_{1}\le\ldots\le h_{T}$ be a sequence of hints such that
  $h_{t}\ge \norm{g_{t}}$, and assume that $h_{t}$ is provided at the start of
  each round $t$.
  Set
  \(
    \psi_{t}(\w)=
    3\int_{0}^{\norm{\w}}\min_{\eta\le\frac{1}{h_{t}}}\sbrac{\frac{\Log{x/\alpha_{t}+1}}{\eta}+\eta V_{t}}dx
  \)
  where
  $V_{t}=4h_{t}^{2}+\norm{g}_{1:\tmm}^{2}$, $\alpha_{t}=\frac{\epsilon }{\sqrt{B_{t}}\log^{2}(B_{t})}$,
  $B_{t}=4\sum_{s=1}^{t}\brac{4+\sum_{s'=1}^{s-1}\frac{\norm{g_{s'}}^{2}}{h_{s'}^{2}}}$,
  and $\epsilon>0$.
  Then
  for all $\cmp\in\R^{d}$, \Cref{alg:centered-md} guarantees
  \begin{align*}
    R_{T}(\cmp)
    &\le
      \Ohat\Bigg(\epsilon h_{T} +
      \norm{\cmp}\sbrac{\sqrt{\norm{g}_{1:T}^{2}\Log{\frac{\norm{\cmp}\sqrt{B_{T+1}}}{\epsilon}+1}}\maxOp h_{T}\Log{\frac{\norm{\cmp}\sqrt{B_{T+1}}}{\epsilon}+1}}\Bigg)
  \end{align*}
  where $\Ohat(\cdot)$ hides constant and $\log(\log)$ factors
\end{restatable}
The proof of this Theorem follows the strategy of previous sections:
from \Cref{lemma:centered-md}
we have $R_{T}(\cmp)\le \psi_{T+1}(\cmp)+\sumtT\delta_{t}$.
To bound $\sumtT\delta_{t}$, we apply \Cref{lemma:md-stability} and show that the growth rate $\Delta_{t}(\w)$
is sufficiently large to ensure
$\sumtT\delta_{t}\le\sumtT\frac{2\norm{g_{t}}^{2}}{\Psi_{t}''(x_{0})}$ for some small
$x_{0}$. The main subtlety compared to \Cref{thm:adaptive-self-stabilizing} is
the influence of the terms $B_{t}$.

The terms $B_{t}$ are carefully chosen to address the range-ratio problem
in an \textit{online} fashion:
we show that $\sqrt{B_t}$ upper bounds the quantity
$h_{t}/h_{\tau_{t}}$, where starting from $\tau_{1}=1$, the variable
$\tau_{t}$ roughly tracks the most-recent round $t$
where the ratio $h_{t}/h_{\tau_{t-1}}$ exceeds a threshold analogous to the one used by FreeGrad to trigger restarts. That is, $B_t$ enacts a kind of ``soft restarting'' by shrinking $w_t$ according to the restarting threshold, just as setting a learning rate of $1/\sqrt{t}$ in online gradient descent can be viewed as a ``soft restart'' in contrast to the standard doubling trick.
It is quite possible that FreeGrad could be similarly modified to
avoid restarts by incorporating $B_{t}$ directly, but this is difficult to verify due to
highly non-trivial polynomial expressions appearing
in the analysis.

We include the pseudocode for the complete scale-free algorithm --- consisting of the gradient clipping
and artificial constraints reductions of \citet{cutkosky2019artificial} applied
with the algorithm characterized in \Cref{thm:scale-invariant} --- and corresponding regret guarantee in
\Cref{app:scale-invariant-alg}.

\section{\secOptTradeoff}
\label{sec:opt-tradeoff}

In the preceeding sections, we focused primarily on standard standard parameter-free
guarantees of the form
\begin{align}
  R_{T}(\cmp)\le O\brac{G\epsilon + \norm{\cmp}\sqrt{\norm{g}_{1:T}^{2}\Log{\frac{\norm{\cmp}\sqrt{\norm{g}_{1:T}^{2}}}{G\epsilon}+1}}}.\label{eq:paramfree}
\end{align}
However, recently there has been
interest in a variant of \Cref{eq:paramfree} that scales instead as
\begin{align}
R_{T}(\cmp)\le O\brac{\epsilon\sqrt{\norm{g}_{1:T}^{2}}+\norm{\cmp}\sbrac{\sqrt{\norm{g}_{1:T}^{2}\Log{\frac{\norm{\cmp}}{\epsilon}+1}}\maxOp G\Log{\frac{\norm{\cmp}}{\epsilon}+1}}}\label{eq:opt-tradeoff}
\end{align}
which captures the optimal \emph{asymptotic} dependence on the variance terms
$\norm{g}_{1:T}^{2}$ by moving them out of the logarithm
\parencite{zhang2022pdebased,zhang2022adversarial,zhang2023improving,zhang2023unconstrained}.
It is easy to see that non-adaptive forms of \Cref{eq:opt-tradeoff} can be
achieved using the usual parameter-free guarantee, \Cref{eq:paramfree},
by setting $\epsilon = \sqrt{T}$. The result can likewise also be achieved
in a horizon-independent manner by applying the doubling trick.
The first work to achieve guarantees of the form
$R_{T}(\cmp)\le O\big(G\epsilon\sqrt{T}+G\norm{\cmp}\sqrt{T\log(\norm{\cmp}/\epsilon+1)}\big)$
\emph{without} resorting to the doubling trick was \cite{zhang2022pdebased}, using 
a novel approach based on discretizing the dynamics of a continuous-time potential function.
The fully-adaptive
bound shown in \Cref{eq:opt-tradeoff} was then later achieved by
\cite{zhang2023improving} by using an improved
discretization strategy.

\newcommand{\LessText}[1]{}
Inspired by these works, in this section we show that bounds in the
form of \Cref{eq:opt-tradeoff}
can also be attained in a straight-forward manner using our mirror descent framework. Interestingly,
each of our static regret algorithms attain a bound analogous
to \Cref{eq:opt-tradeoff} by simply setting $\alpha_{t}=\epsilon$ for all $t$.
The following proposition shows the core argument in the context of our
scale-free algorithm in \Cref{sec:scale-free}.
\LessText{This is somewhat unsurprising: the role of $\alpha_{t}$ in our framework is to ensure
that the stability terms $\sumtT\delta_{t}$ sum to a constant.
In \Cref{thm:adaptive-self-stabilizing},
for instance,
the analysis of the stability terms leads to
$\sumtT\delta_{t}\le O\big(\sumtT \frac{\alpha_{t}\norm{g_{t}}^{2}}{\sqrt{\norm{g}_{1:t}^{2}}}\big)$,
which can be shown to sum to a constant when
$\alpha_{t}\approx \frac{1}{\sqrt{t}\log^{2}(t)}$.
This setting then leads to the horizon-dependence in the logarithmic
terms $\Log{\norm{\cmp}/\alpha_{t}+1}\approx \Log{\norm{\cmp}\sqrt{T}\Log{T}+1}$.
If we instead set $\alpha_{t}=O(1)$, then we avoid the horizon dependence in the 
logarithmic term, but the stability terms will no longer sum to a constant.
Concretely,
the following proposition shows that bounds of the form \Cref{eq:opt-tradeoff}
can be achieved for our scale-free base algorithm in \Cref{sec:scale-free}
by setting $\alpha_{t}=\epsilon$.}
Analogous results hold
for the parameter-free algorithm in \Cref{sec:pf} and the optimistic
algorithm in \Cref{sec:implicit} using an identical argument.
\begin{restatable}{myproposition}{SimpleOptTradeOff}\label{prop:simple-opt-tradeoff}
  Under the same assumptions as \Cref{thm:scale-invariant}, suppose we instead
  set $\alpha_{t}=\epsilon$ for all $t$. Then
  \begin{align*}
    R_{T}(\cmp)
    &\le
      O\brac{\epsilon\sqrt{\norm{g}_{1:T}^{2}}+\norm{\cmp}\sbrac{\sqrt{\norm{g}_{1:T}^{2}\Log{\frac{\norm{\cmp}}{\epsilon}+1}}\maxOp h_{T}\Log{\frac{\norm{\cmp}}{\epsilon}+1}}}.
  \end{align*}

\end{restatable}
\begin{proof}
  Following the same arguments as
  \Cref{thm:scale-invariant} and recalling that
  $V_{t}=4h_{t}^{2}+\norm{g}_{1:\tmm}^{2}\ge \norm{g}_{1:t}^{2}$, we can bound
  \begin{align*}
    R_{T}(\cmp)
    &\le
      \psi_{T+1}(\cmp)+\sumtT\frac{2\alpha_{t}\norm{g_{t}}^{2}}{\sqrt{V_{t}}}
    \le
      \psi_{T+1}(\cmp)+2\epsilon\sumtT\frac{\norm{g_{t}}^{2}}{\sqrt{\norm{g}_{1:t}^{2}}}
    \le
      \psi_{T+1}(\cmp)+4\epsilon\sqrt{\norm{g}_{1:T}^{2}},
  \end{align*}
  where the last line invokes \Cref{lemma:sqrt-integral-bound}. The result then
  follows by using the same argument as \Cref{thm:scale-invariant} to bound
  $\psi_{T+1}(\cmp)\le O\Big(\norm{\cmp}\Big[\sqrt{\norm{g}_{1:T}^{2}\Log{\norm{\cmp}/\epsilon+1}}\maxOp h_{T}\Log{\norm{\cmp}/\epsilon+1}\Big]\Big)$.
\end{proof}

Interestingly, as observed by \cite{zhang2023improving},
the scale-free guarantees in the form of \Cref{eq:opt-tradeoff}
naturally avoid the range-ratio problem. Indeed,
\Cref{prop:simple-opt-tradeoff}
requires neither the restarting strategy of FreeGrad nor the
soft-restarting scheme of our scale-free algorithm in \Cref{thm:scale-invariant}.
This is because in order to achieve a scale-free version of the
standard parameter-free guarantee (\Cref{eq:paramfree}), we must balance out the
gradient ``units'' in the logarithm term of the regularizer, and this
unit-balancing is what gives rise to the range-ratio problem.
By instead setting $\alpha_{t}=\epsilon$, no such unit correction is needed and the
range-ratio problem is naturally avoided.

More generally,
it is possible to  achieve any of the intermediate results
between the two types of parameter-free guarantee
using a similar argument to \Cref{prop:simple-opt-tradeoff}.
The following theorem
provides a result
analogous to \Cref{thm:adaptive-self-stabilizing}, and
is proven in \Cref{app:pf-ii}. An equivalent result also holds
for our optimistic algorithm in \Cref{sec:implicit}, which we defer to \Cref{app:optimistic-tradeoff}.
\begin{restatable}{mytheorem}{PFII}\label{thm:pf-ii}
  Under the same assumptions as \Cref{thm:adaptive-self-stabilizing},
  let $\rho\in[0,\half)$ and suppose we set $\alpha_{t}=\epsilon G^{2\rho}/V_{t}^{\rho}$ for all $t$.
  Then for all $\cmp\in\R^{d}$, \Cref{alg:centered-md} guarantees
  \begin{align*}
    R_{T}(\cmp)
    &\le
      O\Bigg(\frac{\epsilon G^{2\rho}}{1-2\rho}  V_{T+1}^{\half-\rho}+
      \norm{\cmp}\sbrac{\sqrt{V_{T+1}\Log{\frac{\norm{\cmp}V_{T+1}^{\rho}}{\epsilon G^{2\rho}}+1}}\maxOp G\Log{\frac{\norm{\cmp}V_{T+1}^{\rho}}{\epsilon G^{2\rho}}+1}}\Bigg),
  \end{align*}
  where $V_{T+1}\le O(\norm{g}_{1:T}^{2})$.
\end{restatable}
Hence, at $\rho=0$ we have the bound matching the bound in
\cite{zhang2023improving} up to constant terms, and as
$\rho\to\half$ we move toward the usual parameter-free bound, \Cref{eq:paramfree}.
It should be noted that this result is complimentary to
\Cref{thm:adaptive-self-stabilizing}, rather than a generalization: the leading term blows up as we approach
$\rho=\half$. This is unsurprising, as the $\log\log(T)$ penalty incurred by
setting $\alpha_{t}=\frac{\epsilon}{\sqrt{V_{t}}\log^{2}(V_{t}/G^{2})}$ in
\Cref{thm:adaptive-self-stabilizing} is necessary --- without this $\log\log(T)$
dependence, it would be
possible to use the regret guarantee to contradict the Law of Iterated
Logarithm.
Indeed, there are well-known connections between
regret guarantees and concentration inequalities
\citep{rakhlin2017equivalence}, and the regret guarantees of
parameter-free algorithms in particular can be used to 
derive tight concentration inequalities matching the Law of Iterated Logarithm
(see, \textit{e.g.}, \cite{orabona2023concentration}).

A similar result can also be shown for the scale-free base algorithm of
\Cref{sec:scale-free} (Proof in \Cref{app:scale-free-ii}).
\begin{restatable}{mytheorem}{ScaleFreeII}\label{thm:scale-free-ii}
  Under the same assumptions as \Cref{thm:scale-invariant},
  let $\rho\in[0,\half)$ and suppose we set
  $B_{t}^{\rho}=\brac{4\sum_{s=1}^{t}\sbrac{2^{\frac{1}{\rho}}+\sum_{s'=1}^{s-1}\frac{\norm{g_{s'}}^{2}}{h_{s'}^{2}}}}^{\rho}$
  and
  $\alpha_{t}=\epsilon/B_{t}^{\rho}$ for all $t$.\footnote{
    Note that $\lim_{\rho\to 0}B_{t}^{\rho}=2$, so
    for $\rho=0$ we allow an abuse of notation
    by letting $B_{t}^{\rho}:=2$ to avoid specifying
    separate cases.
  }
  Then
  for all $\cmp\in\R^{d}$, \Cref{alg:centered-md} guarantees
  \begin{align*}
    R_{T}(\cmp)
    &\le
      O\brac{
    \frac{\epsilon h_{T}^{2\rho}}{1-2\rho}V_{T+1}^{\half-\rho}
    +
    \norm{\cmp}\sbrac{\sqrt{V_{T+1}\Log{\frac{\norm{\cmp}B_{T+1}^{\rho}}{\epsilon}+1}}\maxOp h_{T}\Log{\frac{\norm{\cmp}B_{T+1}^{\rho}}{\epsilon}+1}}}
  \end{align*}
  where
  and $V_{T+1}\le O(\norm{g}_{1:T}^{2})$.
\end{restatable}

Overall, our approach can be thought of as a
complimentary tool to the approach proposed by \cite{zhang2023improving},
each having different strengths. By starting with a
continuous-time
potential, \cite{zhang2023improving} are able to more carefully control the
stability of the algorithm, enabling them to
achieve tighter constants than our approach.\footnote{In particular
  the continuous-time approach can guarantee regret scaling as
  $k\norm{\cmp}\sqrt{\norm{g}_{1:T}^{2}\Log{\norm{\cmp}/\epsilon +1}}$ for
  the tightest possible constant $k=\sqrt{2}$,
  whereas our result scales with $k=6$.}
On the other hand,
the modular nature of mirror descent algorithms
gives our approach a unifying perspective,
enabling us to design algorithms that make both
types of parameter-free guarantee as special cases.

\section{Conclusion}\label{sec:conclusion}

In this work, we developed a specialization of the standard mirror descent
framework that is particularly suitable for building parameter-free algorithms.
Although we focus our discussion on the unconstrained setting, we emphasize that
our techniques apply equally well in the constrained setting, either by adding
an indicator function to our regularizers $\psi$ or $\varphi$, or via the
unconstrained-to-constrained conversion proposed by \cite{cutkosky2018black}. The
mirror descent formulation allows us to obtain optimal dynamic regret bounds,
implicit updates, and streamlined scale-free algorithms. We nevertheless leave important open questions.
For example, observe
that our dynamic regret algorithm requires $O(d\log(T))$ time and space: can this
be improved?
As some partial progress, in \Cref{app:amortized} we show that one can maintain the
$O\Big(\sqrt{(M^{2}+MP_{T})\norm{g}_{1:T}^{2}}\Big)$ bound up to poly-logarithmic factors
using only $O(d)$ \textit{amortized} computation.
Further, we focused on the use of $\varphi_{t}$ to achieve novel bounds, and did not explore  its more traditional use in incorporating a user-provided composite objective.
We look forward to exciting developments in this area.

\acks{This work was completed in part while AJ was visiting
  Boston University. AJ gratefully acknowledges support from the
  NSERC Alexander Graham Bell Scholarship and Alberta Innovates
  Data-enabled Innovation Scholarship.}

\clearpage
\bibliography{paper.bib}

\clearpage

\appendix

\newtheorem{applemma}{Lemma}

\section{A Strong Mirror Descent Lemma}\label{app:strong-centered-md}

In this section we derive a regret template for Centered Mirror Descent which
holds for arbitrary sequences of loss functions and choices of
$\psi_{t}$ and $\varphi_{t}$.
The result is analogous to the Strong FTRL Lemma of
\textcite{mcmahan2017survey}, but applies to
a sequence of comparators and is tailored to mirror descent-style analysis.

In this section, the following short-hand
notation will be convenient:
\begin{align*}
  \Dhat_{f}(x,y, g_{y})\defeq f(x)-f(y)-\inner{g_{y}, x-y}.
\end{align*}
where $f$ is a subdifferentiable function and $g_{y}$
is an arbitrary element of $\partial f(y)$. Note that when $f$ is
differentiable, then $\partial f(y)=\Set{\grad f(y)}$, so
the short-hand reduces to the standard bregman divergence.
Moreover, observe that $\Dhat$ still satisfies the usual subgradient inequalities. For instance,
if $f$ is convex, then for any $g_{y}\in\partial f(y)$ we have
$\Dhat_{f}(x,y,g_{y})\ge 0$.

\begin{restatable}{mylemma}{StrongMDLemma}\label{lemma:strong-centered-md}
  \textbf{(Strong Centered Mirror Descent Lemma)}
  For all $t$, let $\ell_{t}(\cdot)$ be a subdifferentiable function,
  $\varphi_{t}(\cdot)$ be a subdifferentiable non-negative function,
  and $\psi_{t}(\cdot)$ be a differentiable non-negative function.
  Define $\Delta_{t}(\w)=D_{\psi_{\tpp}}(\w|\w_{1})-D_{\psi_{t}}(\w|\w_{1})$,
  $\phi_{t}(\w)=\Delta_{t}(\w)+\varphi_{t}(\w)$, and
  set $\wtpp=\argmin_{\w\in\R^{d}}\ell_{t}(\w)+D_{\psi_{t}}(\w|\wt)+\phi_{t}(\w)$.

  Then, for all $t$ there is some $\nabla \ell_t(w_{t+1})\in\partial \ell_t(w_{t+1})$ and $\nabla \varphi_t(w_{t+1})\in \partial\varphi_t(w_{t+1})$ such that $\nabla\ell_{t}(\wtpp)+ \grad\psi_{\tpp}(\wtpp)-\grad\psi_{t}(\wt) + \nabla \varphi_t(w_{t+1})=\zeros$,
  and for any $\cmp_{1},\ldots,\cmp_{T}$ in $\R^{d}$,
  \begin{align*}
    \sumtT\ell_{t}(\wt)-\ell_{t}(\cmp_{t})
    &=
      D_{\psi_{T+1}}(\cmp_{T}|\w_{1})-D_{\psi_{T+1}}(\cmp_{T}|\w_{T+1})+\sumtT\varphi_{t}(\cmp_{t})\\
    &\qquad
      +\sum_{t=2}^{T}\underbrace{\inner{\grad\psi_{t}(\wt)-\grad\psi_{t}(\w_{1}),\cmp_{\tmm}-\cmp_{t}}}_{=:\mathcal{P}_{t}}\\
    &\qquad
      +\sumtT\underbrace{\inner{g_{t},\wt-\wtpp}-D_{\psi_{t}}(\wtpp|\wt)-\phi_{t}(\wtpp)}_{=:\delta_{t}}\\
    &\qquad
      +\sumtT \underbrace{-\Dhat_{\ell_{t}+\varphi_{t}}\Big(\cmp_{t},\wtpp, \grad\ell_{t}(\wtpp)+\grad\varphi_{t}(\wtpp)\Big)-\Dhat_{\ell_{t}}\Big(\wtpp, \wt, g_{t}\Big)}_{=:\cL_{t}}.
  \end{align*}
  where $g_{t}$ is an arbitrary element of $\partial\ell_{t}(\wt)$.
\end{restatable}

The bound includes three terms not included in
Lemma~\ref{lemma:centered-md}:
$-D_{\psi_{T}}(\cmp_{T}|w_{T+1})$,
$-\hat D_{\ell_{t}}(\wtpp,\wt,g_{t})$, and
$-\Dhat_{\ell_{t}+\varphi_{t}}(\cmp_{t},\wtpp,\grad\ell_{t}(\wtpp)+\grad\varphi_{t}(\wtpp))$.
Observe that the lemma holds even for non-convex losses; in
this case we'll need to account for the fact that the terms
$-\Dhat_{\ell_{t}}(\cmp_{t},\wtpp,\grad\ell_{t}(\wtpp))$ and $-\Dhat_{\ell_{t}}(\wtpp,\wt,g_{t})$ may be positive and
may require additional effort to control.
When the losses
are convex the terms
$-\Dhat_{\ell_{t}}(\cmp_{t},\wtpp,\grad\ell_{t}(\wtpp))$ and $-\Dhat_{\ell_{t}}(\wtpp,\wt,g_{t})$ can often be leveraged
in useful ways, particularly when the $\ell_{t}$ have nice properties
such as strong convexity. In this work we only assume
convexity of $\ell_{t}$ and drop these terms.
Similarly, for simplicity we assume that $\varphi_{t}$ is convex
so that we can bound $-\Dhat_{\varphi_{t}}(\cmp_{t},\wtpp,\grad\varphi_{t}(\wtpp))\le 0$. It's possible that
this term could also be leveraged in some useful way, but we
do not investigate this in the current work.

\newcommand{\elltilde}{\widetilde{\ell}}

\begin{proof} \textbf{of \Cref{lemma:strong-centered-md}}

First, observe that the existence of the specified $\nabla \ell_t(w_{t+1})\in\partial \ell_t(w_{t+1})$ and $\nabla \varphi_t(w_{t+1})\in \partial\varphi_t(w_{t+1})$ follows directly from the first order optimality conditions applied to the update $w_{t+1}= \argmin_{w} \ell_t(w) + D_{\psi_{t}}(w|w_t) + \Delta_t(w) + \varphi_t(w)$.

Further, again by first order optimality conditions, we have:
\begin{align*}
    \nabla \ell_t(w_{t+1})+\nabla \varphi_t(w_{t+1}) + \nabla \Delta_t(w_{t+1}) + \nabla \psi_t(w_{t+1}) - \nabla \psi_t(w_t)=\zeros
\end{align*}
Now, we define $\elltilde_{t}(\w)=\ell_{t}(\w)+\phi_{t}(\w)$, and begin by writing
  \begin{align}
    \sumtT\ell_{t}(\wt)-\ell_{t}(\cmp_{t})
    &=
      \sumtT\ell_{t}(\wtpp)-\ell_{t}(\cmp_{t})+\sumtT\ell_{t}(\wt)-\ell_{t}(\wtpp)\nonumber\\
    &=
      \sumtT \elltilde_{t}(\wtpp)-\elltilde_{t}(\cmp)+\phi_{t}(\cmp_{t})+\sumtT\ell_{t}(\wt)-\ell_{t}(\wtpp)-\phi_{t}(\wtpp)\label{eq:strong-md-1},
  \end{align}
  where the last line adds and subtracts $\phi_{t}(\wtpp)$ and
  $\phi_{t}(\cmp_{t})$. For the rest of this proof, we define $\tilde g_t = \nabla \ell_t(w_{t+1})+\nabla \varphi_t(w_{t+1}) + \nabla \Delta_t(w_{t+1})$.
  Thus,
  $\gtilde_{t} + \grad\psi_{t}(\wtpp)-\grad\psi_{t}(\wt)=\zeros$. Further,
  with $\elltilde_{t}(\w)=\ell_{t}(\w)+\phi_{t}(\w)$, we have
  $\tilde g_t\in\partial \elltilde_t(\wtpp)$.
  Now observe that we can write
  \begin{align*}
    \sumtT\elltilde_{t}(\wtpp)-\elltilde_{t}(\cmp_{t})
    &=
    \sumtT\inner{\gtilde_{t},\wtpp-\cmp_{t}}+\inner{\gtilde_{t},\cmp_{t}-\wtpp}+\elltilde_{t}(\wtpp)-\elltilde_{t}(\cmp_{t})\\
    &=
      \sumtT\inner{\gtilde_{t},\wtpp-\cmp_{t}}-\Dhat_{\elltilde_{t}}(\cmp_{t},\wtpp,\gtilde_{t})\\
    &=
      \sumtT\inner{\grad\psi_{t}(\wt)-\grad\psi_{t}(\wtpp),\wtpp-\cmp_{t}}
      -\Dhat_{\elltilde_{t}}(\cmp_{t},\wtpp,\gtilde_{t})\\
    &=
      \sumtT D_{\psi_{t}}(\cmp_{t}|\wt)-D_{\psi_{t}}(\cmp_{t}|\wtpp)-D_{\psi_{t}}(\wtpp|\wt)-\Dhat_{\elltilde_{t}}(\cmp_{t},\wtpp,\gtilde_{t})
  \end{align*}
  where the last line uses the well-known three-point relation
  $\inner{\grad f(a)-\grad f(b),b-c}=D_{f}(c|a)-D_{f}(c|b)-D_{f}(b|a)$.
  Hence, recalling that $\phi_{t}(\wt)=\Delta_{t}(\wt)+\varphi_{t}(\w)$ and observing that
  $D_{\Delta_{t}}(\cmp_{t}|\wtpp)=D_{\psi_{\tpp}-\psi_{t}}(\cmp_{t}|\wtpp)=D_{\psi_{\tpp}}(\cmp_{t}|\wtpp)-D_{\psi_{t}}(\cmp_{t}|\wtpp)$
  leaves us with
  \begin{align*}
    \sumtT\elltilde_{t}(\wtpp)-\elltilde_{t}(\cmp_{t})
    &=
      \sumtT D_{\psi_{t}}(\cmp_{t}|\wt)-D_{\psi_{t}}(\cmp_{t}|\wtpp)-D_{\psi_{t}}(\wtpp|\wt)\\
    &\qquad
      -\Dhat_{\ell_{t}+\varphi_{t}+\Delta_{t}}(\cmp_{t},\wtpp,\grad\ell_{t}(\wtpp)+\grad\varphi_{t}(\wtpp)+\grad\Delta_{t}(\wtpp))\nonumber\\
    &\overset{(*)}{=}
      \sumtT D_{\psi_{t}}(\cmp_{t}|\wt)-D_{\psi_{t}}(\cmp_{t}|\wtpp)-D_{\psi_{t}}(\wtpp|\wt)\\
    &\qquad
      -D_{\Delta_{t}}(\cmp_{t}|\wtpp)-\Dhat_{\ell_{t}+\varphi_{t}}(\cmp_{t},\wtpp,\grad\ell_{t}(\wtpp)+\grad\varphi_{t}(\wtpp))\nonumber\\
    &=
      \sumtT D_{\psi_{t}}(\cmp_{t}|\wt)-D_{\psi_{\tpp}}(\cmp_{t}|\wtpp)-D_{\psi_{t}}(\wtpp|\wt)\\
    &\qquad
      -\Dhat_{\ell_{t}+\varphi_{t}}(\cmp_{t},\wtpp,\grad\ell_{t}(\wtpp)+\grad\varphi_{t}(\wtpp))\nonumber\\
    &=
      \underbrace{D_{\psi_{1}}(\cmp_{1}|\w_{1})-D_{\psi_{T+1}}(\cmp_{T}|\w_{T+1})+\sum_{t=2}^{T}D_{\psi_{t}}(\cmp_{t}|\wt)-D_{\psi_{t}}(\cmp_{\tmm}|\w_{t})}_{\encircle{A}}\nonumber\\
    &\qquad
      +\sumtT -D_{\psi_{t}}(\wtpp|\wt)-\Dhat_{\ell_{t}+\varphi_{t}}(\wt,\wtpp,\grad\ell_{t}(\wtpp)+\grad\varphi_{t}(\wtpp)),
  \end{align*}
  where $(*)$ uses the fact that $\Dhat_{\Delta_{t}}(x,y,g_{y})=D_{\Delta_{t}}(x|y)$
  since $\Delta_{t}(x)$ is a differentiable  function.
  Returning to the full regret bound we have
  \begin{align*}
    \sumtT\ell_{t}(\wt)-\ell_{t}(\cmp_{t})
    &=
      \encircle{A}+\sumtT\phi_{t}(\cmp_{t})+\sumtT\ell_{t}(\wt)-\ell_{t}(\wtpp)-D_{\psi_{t}}(\wtpp|\wt)-\phi_{t}(\wtpp)\\
    &\qquad
      +\sumtT-\Dhat_{\ell_{t}+\varphi_{t}}(\cmp_{t},\wtpp,\grad\ell_{t}(\wtpp)+\grad\varphi_{t}(\wtpp))\\
    &=
      \encircle{A}+\sumtT\sbrac{\Delta_{t}(\cmp_{t})+\varphi_{t}(\cmp_{t})}+\sumtT\underbrace{\inner{g_{t},\wt-\wtpp}-D_{\psi_{t}}(\wtpp|\wt)-\phi_{t}(\wtpp)}_{=:\delta_{t}}\\
    &\qquad
      +\sumtT\underbrace{-\Dhat_{\ell_{t}+\varphi_{t}}(\cmp_{t},\wtpp,\grad\ell_{t}(\wtpp)+\grad\varphi_{t}(\wtpp))-\Dhat_{\ell_{t}}(\wtpp,\wt,g_{t})}_{=:\cL_{t}},
  \end{align*}
  where the last line lets $g_{t}\in\partial\ell_{t}(\wt)$ and observes
  $\ell_{t}(\wt)-\ell_{t}(\wtpp)=\inner{g_{t},\wt-\wtpp}-\inner{g_{t},\wt-\wtpp}+\ell_{t}(\wt)-\ell_{t}(\wtpp)=\inner{g_{t},\wt-\wtpp}-\Dhat_{\ell_{t}}(\wtpp,\wt, g_{t})$.
  Finally, observe that
  $\sumtT\Delta_{t}(\cmp_{t})=D_{\psi_{T+1}}(\cmp_{T}|\w_{1})-D_{\psi_{1}}(\w_{1}|\cmp_{1})+\sum_{t=2}^{T}D_{\psi_{t}}(\cmp_{\tmm}|\w_{1})-D_{\psi_{t}}(\cmp_{t}|\w_{1})$,
  so combining with \encircle{A} yields
  \begin{align*}
    \encircle{A}+\sumtT\Delta_{t}(\cmp_{t})
    &=
      D_{\psi_{1}}(\cmp_{1}|\w_{1})-D_{\psi_{T+1}}(\cmp_{T}|\w_{T+1})+\sum_{t=2}^{T}D_{\psi_{t}}(\cmp_{t}|\wt)-D_{\psi_{t}}(\cmp_{\tmm}|\wt)\\
    &\qquad
      +D_{\psi_{T+1}}(\cmp_{T}|\w_{1})-D_{\psi_{1}}(\w_{1}|\cmp_{1})+\sum_{t=2}^{T}D_{\psi_{t}}(\cmp_{\tmm}|\w_{1})-D_{\psi_{t}}(\cmp_{t}|\w_{1})\\
    &=
      D_{\psi_{T+1}}(\cmp_{T}|\w_{1})-D_{\psi_{T+1}}(\cmp_{T}|\w_{T+1})
      +\sum_{t=2}^{T}D_{\psi_{t}}(\cmp_{t}|\wt)-D_{\psi_{t}}(\cmp_{\tmm}|\w_{t})\\
    &\qquad+
      \sum_{t=2}^{T}D_{\psi_{t}}(\cmp_{\tmm}|\w_{1})-D_{\psi_{t}}(\cmp_{t}|\w_{1})\\
    &=
      D_{\psi_{T+1}}(\cmp_{T}|\w_{1})-D_{\psi_{T+1}}(\cmp_{T}|\w_{T+1})\\
    &\qquad
      +\sum_{t=2}^{T}\sbrac{\psi_{t}(\cmp_{t})-\psi_{t}(\cmp_{\tmm})-\inner{\grad\psi_{t}(\wt),\cmp_{t}-\cmp_{\tmm}}}\\
    &\qquad
      +\sum_{t=2}^{T}\sbrac{\psi_{t}(\cmp_{\tmm})-\psi_{t}(\cmp_{t})-\inner{\grad\psi_{t}(\w_{1}),\cmp_{\tmm}-\cmp_{t}}}\\
    &=
      D_{\psi_{T+1}}(\cmp_{T}|\w_{1})-D_{\psi_{T+1}}(\cmp_{T}|\w_{T+1})
      +\sum_{t=2}^{T}\underbrace{\inner{\grad\psi_{t}(\wt)-\grad\psi_{t}(\w_{1}),\cmp_{\tmm}-\cmp_{t}}}_{=:\cP_{t}}
  \end{align*}
  and hence
  \begin{align*}
    \sumtT\ell_{t}(\wt)-\ell_{t}(\cmp_{t})
    &=
      \encircle{A}+\sumtT\Delta_{t}(\cmp_{t})+\sumtT\varphi_{t}(\cmp_{t})+\sumtT\delta_{t}+\sumtT\cL_{t}\\
    &=
      D_{\psi_{T+1}}(\cmp_{T}|\w_{1})-D_{\psi_{T+1}}(\cmp_{T}|\w_{T+1})
      +\sum_{t=2}^{T}\cP_{t}+\sumtT\varphi_{t}(\cmp_{t})+\sumtT\delta_{t}+\sumtT\cL_{t}.
  \end{align*}

\end{proof}

\newpage
\section{Proofs for Section~\ref{sec:centered-md} (Centered Mirror Descent)}

\subsection{Proof of Lemma~\ref{lemma:centered-md}}\label{app:centered-md}

\CenteredMDLemma*
\begin{proof}
  From \Cref{lemma:strong-centered-md} we have that
  \begin{align*}
    \sumtT\ell_{t}(\wt)-\ell_{t}(\cmp_{t})
    &=
      D_{\psi_{T+1}}(\cmp_{T}|\w_{1})-D_{\psi_{T+1}}(\cmp_{T}|\w_{T+1})+\sumtT\varphi_{t}(\cmp_{t})+\sum_{t=2}^{T}\cP_{t}+\sumtT\delta_{t}+\sumtT\cL_{t},
  \end{align*}
  where
  \begin{align*}
    \cP_{t}&=\inner{\grad\psi_{t}(\wt)-\grad\psi_{t}(\w_{1}),\cmp_{\tmm}-\cmp_{t}}\\
    \delta_{t}&=\inner{g_{t},\wt-\wtpp}-D_{\psi_{t}}(\wtpp|\wt)-\phi_{t}(\wtpp)\\
    \cL_{t}&=-\Dhat_{\ell_{t}}(\cmp_{t},\wtpp,\grad\ell_{t}(\wtpp))-\Dhat_{\varphi_{t}}(\cmp_{t},\wtpp,\grad\varphi_{t}(\wtpp))-\Dhat_{\ell_{t}}(\wtpp,\wt,g_{t}),
  \end{align*}
  where $g_{t}\in\ell_{t}(\wt)$ and $\Dhat_{f}(x,y,g_{y})=f(x)-f(y)-\inner{g_{y},x-y}$ for subdifferentiable
  function $f$ and $g_{y}\in\partial f(y)$.
  Since $\ell_{t}(\cdot)$ and $\varphi_{t}(\cdot)$ are convex, for any
  $x,y\in\R^{d}$ we have
  $\Dhat_{\ell_{t}}(x,y,\grad\ell_{t}(y))\ge 0$ for any $\grad\ell_{t}(y)\in\partial\ell_{t}(y)$ and
  $\Dhat_{\varphi_{t}}(x,y\grad\varphi_{t}(y))\ge 0$ for any
  $\grad\varphi_{t}(y)\in\partial\varphi_{t}(y)$,
  so $\sumtT\cL_{t}\le 0$.
  Further,
  using the assumption that $\w_{1}\in\argmin_{\w\in\R^{d}}\psi_{t}(\w)$ and
  $\psi_{t}(\w)\ge 0$ for all
  $t$, we have that $\grad\psi_{t}(\w_{1})=\zeros$ and $D_{\psi_{t}}(\w|\w_{1})\le\psi_{t}(\w)$ for any $\w\in\R^{d}$.
  Using this along with the fact that Bregman divergences \textit{w.r.t} convex
  functions are non-negative
  yields
  \begin{align*}
    \sumtT\ell_{t}(\wt)-\ell_{t}(\cmp_{t})
    &\le
      \psi_{T+1}(\cmp_{T})+\sum_{t=2}^{T}\inner{\grad\psi_{t}(\wt),\cmp_{\tmm}-\cmp_{t}}+\sumtT\varphi_{t}(\cmp_{t})+\sumtT\delta_{t}
  \end{align*}
  The stated bound then follows by re-indexing
  $\sum_{t=2}^{T}\inner{\grad\psi_{t}(\wt),\cmp_{\tmm}-\cmp_{t}}$.
\end{proof}

\newpage
\subsection{Proof of Lemma \ref{lemma:md-stability}}\label{app:md-stability}

\MDStability*
\begin{proof}
First, consider the case that the origin is contained in the line segment connecting $w_t$ and $w_{t+1}$. Then, there exists sequences $\hat w^1_t,\hat w^2_t\dots$ and $\hat w^1_{t+1},\hat w^2_{t+1}\dots$ such that $\lim_{n\to\infty} \hat w^n_t=w_t$, $\lim_{n\to\infty} \hat w^n_{t+1}=w_{t+1}$ and $0$ is not contained in the line segment connecting $\hat w^n_t$ and $\hat w^n_{t+1}$ for all $n$. Since $\psi$ is twice differentiable everywhere except the origin, if we define $\hat \delta_t^n = \langle g_t, \hat w_t^n - \hat w_{t+1}^n\rangle - D_{\psi_t}(\hat w^n_{t+1}|\hat w^n_t) - \eta_t(\|\hat w^n_{t+1}\|) \|g_t\|^2$, then $\lim_{n\to\infty}\hat \delta_t^n=\hat \delta_t$. Thus, it suffices to prove the result for the case that the origin is \emph{not} contained in the line segment connecting $w_t$ and $w_{t+1}$. The rest of the proof considers exclusively this case.

For brevity denote $\hat\delta_{t}\defeq\inner{g_{t},\wt-\wtpp}-D_{\psi_{t}}(\wtpp|\wt)-\eta_{t}(\norm{\wtpp})\norm{g_{t}}^{2}$.
Since the origin is not in the line segment connecting $w_t$ and $w_{t+1}$, $\psi_t$ is twice differentiable on this line segment. Thus, By Taylor's theorem, there is a $\wtilde$ on the line connecting
$\wt$ and $\wtpp$ such that
\begin{align*}
  D_{\psi_{t}}(\wtpp|\wt)
  &=
    \half\norm{\wt-\wtpp}^{2}_{\grad^{2}\psi_{t}(\wtilde)}\\
  &\ge
    \half\norm{\wt-\wtpp}^{2}\Psi_{t}''(\norm{\wtilde})
\end{align*}
where the last line observes $\psi_{t}(\w)=\Psi_{t}(\norm{\w})$ and uses the
assumptions that $\Psi_{t}'(x)\ge 0$ and $\Psi_{t}'''(x)\le 0$ (hence
$\Psi_{t}'$ is concave) to apply \Cref{lemma:simple-radially-symmetric-bound}.
Thus,
\begin{align*}
  \hat\delta_{t}
  &=
    \inner{g_{t},\wt-\wtpp}-D_{\psi_{t}}(\wtpp|\wt)-\eta_{t}(\norm{\wtpp})\norm{g_{t}}^{2}\\
  &\le
    \inner{g_{t},\wt-\wtpp}-\half\norm{\wt-\wtpp}^{2}\Psi_{t}''(\norm{\wtilde})-\eta_{t}(\norm{\wtpp})\norm{g_{t}}^{2}\\
  &\overset{(a)}{\le}
    \inner{g_{t},\wt-\wtpp}-\half\norm{\wt-\wtpp}^{2}\Psi_{t}''(\norm{\wtilde})-\eta_{t}(\norm{\wtilde})\norm{g_{t}}^{2}+\eta_{t}'(\norm{\wtilde})\norm{g_{t}}^{2}\brac{\norm{\wtpp}-\norm{\wtilde}}\\
  &\overset{(b)}{\le}
    \inner{g_{t},\wt-\wtpp}-\half\norm{\wt-\wtpp}^{2}\Psi_{t}''(\norm{\wtilde})-\eta_{t}(\norm{\wtilde})\norm{g_{t}}^{2}+\norm{g_{t}}\norm{\wtilde-\wtpp}\\
  &\overset{(c)}{\le}
    2\norm{g_{t}}\norm{\wt-\wtpp}-\half\norm{\wt-\wtpp}^{2}\Psi_{t}''(\norm{\wtilde})-\eta_{t}(\norm{\wtilde})\norm{g_{t}}^{2}\\
  &\le
    \frac{2\norm{g_{t}}^{2}}{\Psi_{t}''(\norm{\wtilde})} - \eta_{t}(\norm{\wtilde})\norm{g_{t}}^{2},
\end{align*}
where $(a)$ uses convexity of $\eta_{t}(x)$, $(b)$ uses the Lipschitz assumption
$\eta_{t}'(\norm{\wtilde})\le 1/G_{t}\le\frac{1}{\norm{g_{t}}}$ and triangle inequality,
and $(c)$ uses Cauchy-Schwarz inequality and the fact that $\norm{\wtilde-\wtpp}\le\norm{\wt-\wtpp}$ for any
$\wtilde$ on the line connecting $\wtpp$ and $\wt$.
Next, by assumption we know that there is an $x_{0}\ge 0$ such that
$\abs{\Psi_{t}'''(x)}\le\frac{\eta_{t}'(x)}{2}\Psi_{t}''(x)^{2}$. If it happens that
$\norm{\wtilde}\le x_{0}$, then
\begin{align*}
  \hat\delta_{t}
  &\le
    \frac{2\norm{g_{t}}^{2}}{\Psi_{t}''(\norm{\wtilde})} - \eta_{t}(\norm{\wtilde})\norm{g_{t}}^{2}
    \le \frac{2\norm{g_{t}}^{2}}{\Psi_{t}''(x_{0})},
\end{align*}
which follows from the fact that $\Psi_{t}'''(x)\le 0$ implies that $\Psi_{t}''(x)$ is
non-increasing in $x$, and hence $\Psi_{t}''(\norm{\wtilde})\ge\Psi_{t}''(x_{0})$
whenever $\norm{\wtilde}\le x_{0}$. Otherwise, when $\norm{\wtilde}> x_{0}$
we have by assumption that
$\frac{\abs{\Psi_{t}'''(x)}}{\Psi_{t}''(x)^{2}}=\frac{-\Psi_{t}'''(x)}{\Psi_{t}''(x)^{2}}=\frac{d}{dx}\frac{1}{\Psi_{t}''(x)}\le\frac{\eta_{t}'(x)}{2}$
for any $x> x_{0}$, so
integrating from $x_{0}$ to $\norm{\wtilde}$ we have
\begin{align*}
  \frac{1}{\Psi_{t}''(\norm{\wtilde})}-\frac{1}{\Psi_{t}''(x_{0})}&\le \half\int_{x_{0}}^{\norm{\wtilde}}\eta_{t}'(x)dx\\
  \implies
  \frac{1}{\Psi_{t}''(\norm{\wtilde})}&\le \frac{1}{\Psi_{t}''(x_{0})}+\half\int_{x_{0}}^{\norm{\wtilde}}\eta_{t}'(x)dx
                                        \le \frac{1}{\Psi_{t}''(x_{0})}+\half\int_{0}^{\norm{\wtilde}}\eta_{t}'(x)dx=\frac{1}{\Psi_{t}''(x_{0})}+\frac{\eta_{t}(\norm{\wtilde})}{2},
\end{align*}
so
\begin{align*}
  \hat\delta_{t}
  \le
  \frac{2\norm{g_{t}}^{2}}{\Psi_{t}''(\norm{\wtilde})}-\eta_{t}(\norm{\wtilde})\norm{g_{t}}^{2}
  \le
  2\norm{g_{t}}^{2}\brac{\frac{1}{\Psi_{t}''(x_{0})} + \frac{\eta_{t}(\norm{\wtilde})}{2}}-\eta_{t}(\norm{\wtilde})\norm{g_{t}}^{2}
  =\frac{2\norm{g_{t}}^{2}}{\Psi_{t}''(x_{0})}.
\end{align*}
Thus, in either case we have
$\hat \delta_{t}\le \frac{2\norm{g_{t}}^{2}}{\Psi_{t}''(x_{0})}$.

\end{proof}

\section{Proofs for Section~\ref{sec:pf} (\secPF)}\label{app:pf}
\subsection{Proof of Theorem~\ref{thm:adaptive-self-stabilizing}}

The theorem is restated below. Pseudocode for the algorithm
characterized by this theorem is given in \Cref{alg:optimal-pf} for convenience.

\begin{algorithm}
  \SetAlgoLined
  \KwInput{Lipschitz bound G, Value $\epsilon>0$}\\
  \KwInitialize{$V_{1}=4G^{2}$, $\w_{1}=\zeros$, $\theta_{1}=\zeros$}\\
  \For{$t=1:T$}{
    Play $\wt$, receive subgradient $g_{t}$\\
    Set $\theta_{\tpp}=\theta_{t}-g_{t}$, $V_{\tpp}=V_{t}+\norm{g_{t}}^{2}$,
    $\alpha_{\tpp}=\frac{\epsilon G}{\sqrt{V_{\tpp}}\log^{2}(V_{\tpp}/G^{2})}$, and
    define
    \(f_{\tpp}(\theta)=\begin{cases} \frac{\norm{\theta}^{2}}{36 V_{\tpp}}&\text{if
    }\norm{\theta}\le\frac{6V_{\tpp}}{G}\\
    \frac{\norm{\theta}}{3 G}-\frac{V_{\tpp}}{G^{2}}&\text{otherwise}
  \end{cases}\)\\
  \BlankLine
    Update
    $\wtpp = \frac{\alpha_{t+1}\theta_\tpp}{\|\theta_\tpp\|}\sbrac{\Exp{f_{\tpp}(\theta_{\tpp})}-1}$\\
  }
  \caption{Parameter-free Learning via Centered Mirror Descent}
  \label{alg:optimal-pf}
\end{algorithm}
\newpage
\begin{manualtheorem}{\ref{thm:adaptive-self-stabilizing}}
  Let $\ell_{1},\ldots,\ell_{T}$ be $G$-Lipschitz convex functions and
  $g_{t}\in\partial\ell_{t}(\wt)$ for all $t$. Let $\epsilon >0$,
  $k\ge 3$,
  $V_{t}= 4 G^{2} +\norm{g}_{1:\tmm}^{2}$, and
  $\alpha_{t}=\frac{\epsilon G}{\sqrt{V_{t}}\log^{2}(V_{t}/G^{2})}$. For all $t$, set
  \begin{align*}
    \psi_{t}(\w)
    &=
      k\int_{0}^{\norm{\w}}\min_{\eta\le 1/G}\sbrac{\frac{\Log{x/\alpha_{t}+1}}{\eta}+\eta V_{t}}dx.
  \end{align*}
  Then for all $\cmp\in\R^{d}$, \Cref{alg:optimal-pf} guarantees
  \begin{align*}
    R_{T}(\cmp)
    &\le
    4G\epsilon+2k\norm{\cmp}\Max{\sqrt{V_{T+1}\Log{\norm{\cmp}/\alpha_{T+1}+1}},G\Log{\norm{\cmp}/\alpha_{T+1}+1}}\\
  \end{align*}
\end{manualtheorem}

\begin{proof}
First, let us derive the update formula, which can be seen in
\Cref{alg:optimal-pf}. By first-order optimality conditions
for $\wtpp=\argmin_{\w\in\R^{d}}\inner{g_{t},\w}+D_{\psi_{t}}(\w|\wt)+\Delta_{t}(\w)$
we have:
\begin{align*}
g_t + \nabla \psi_t(w_{t+1}) -\nabla \psi_t(w_t) + \nabla \Delta_t(w_{t+1})=\zeros
\end{align*}
Expanding the definition of $\Delta_t(\w)=\psi_{\tpp}(\w)-\psi_{t}(\w)$, we obtain:
\begin{align*}
    g_t + \nabla \psi_{t+1}(w_{t+1}) - \nabla \psi_t(w_t) =\zeros,
\end{align*}
and unrolling the recursion we have
\begin{align*}
  \grad\psi_{\tpp}(\wtpp)=\grad\psi_{t}(\wt)-g_{t}=\grad\psi_{\tmm}(\wtmm)-g_{\tmm}-g_{t}=\ldots=-g_{1:t}.
\end{align*}
Inspecting the equation for $\psi_{\tpp}$ then yields:
\begin{align*}
   \frac{\wtpp}{\|\wtpp\|}\Psi'_{t+1}(\|\wtpp\|)= - g_{1:t}
\end{align*}
where we define the function
\begin{align*}
  \Psi_{\tpp}'(x)&=k\min_{\eta\le 1/G}\sbrac{\frac{\Log{x/\alpha_{\tpp}+1}}{\eta}+\eta V_{\tpp}}\\
  &=\begin{cases}
    2k\sqrt{V_{\tpp}\Log{x/\alpha_{\tpp}+1}}&\text{if } G\sqrt{\Log{x/\alpha_{\tpp}+1}}\le\sqrt{V_{\tpp}}\\
    kG\Log{x/\alpha_{\tpp}+1}+\frac{kV_{\tpp}}{G}&\text{otherwise}.
    \end{cases}
\end{align*}
From this, we immediately see that $\wtpp = x\frac{-g_{1:t}}{\|g_{1:t}\|}$ for some constant $x$ that satisfies:
\begin{align*}
    \Psi'_{t+1}(x)=\|g_{1:t}\|
\end{align*}
Now we see that
one of two cases occurs: either
\begin{align*}
  \Psi_{\tpp}'(x)=2k\sqrt{V_{\tpp}\Log{x/\alpha_{\tpp}+1}},
\end{align*}
which holds when $\frac{1}{G}\ge\sqrt{\Log{x/\alpha_{\tpp}+1}/V_{\tpp}}$,
or alternatively we have
\begin{align*}
  \Psi_{\tpp}'(x)=kG\Log{x/\alpha_{\tpp}+1}+\frac{kV_{\tpp}}{G}
\end{align*}
which holds when $\frac{1}{G}\le\sqrt{\Log{x/\alpha_{\tpp}+1}/V_{\tpp}}$.
Observe that at the boundary value where
$\frac{1}{G}=\sqrt{\Log{x/\alpha_{\tpp}+1}/V_{\tpp}}$
we have
\begin{align*}
  \Psi_{\tpp}'(x)=2k\sqrt{V_{\tpp}\Log{x/\alpha_{\tpp}+1}}=\frac{2kV_{\tpp}}{G}.
\end{align*}
Using this, we consider two cases. First, if
$\norm{g_{1:t}}\le\frac{2k V_{\tpp}}{G}$,
then we have
\begin{align*}
  2k\sqrt{V_{\tpp}\Log{\norm{\wtpp}/\alpha_{\tpp}+1}}&=\norm{g_{1:t}}\\
  \norm{\wtpp}&=\alpha_{\tpp}\sbrac{\Exp{\frac{\norm{g_{1:t}}^{2}}{4k^{2}V_{\tpp}}}-1}.
\end{align*}
On the other hand, if $\norm{g_{1:t}}\ge\frac{2kV_{\tpp}}{G}$ then
\begin{align*}
  kG\Log{\norm{\wtpp}/\alpha_{\tpp}+1}+\frac{kV_{\tpp}}{G}&=\norm{g_{1:t}}\\
  \norm{\wtpp}&=\alpha_{\tpp}\sbrac{\Exp{\frac{\norm{g_{1:t}}}{kG}-\frac{V_{\tpp}}{G^{2}}}-1}.
\end{align*}
Putting these cases together yields the update described in \Cref{alg:optimal-pf} (with $k=3$, which is important later in the regret analysis).

Now, we concentrate on proving the regret bound.

For brevity we define the function $F_{t}(x)=\Log{x/\alpha_{t}+1}$. Recall that we have set $\Psi_{t}'(x)=k\min_{\eta\le 1/G}\sbrac{\frac{F_{t}(x)}{\eta}+\eta V_{t}}$ so that
$\Psi_{t}(x)=k\int_{0}^{x}\min_{\eta\le 1/G}\sbrac{\frac{F_{t}(z)}{\eta}+\eta V_{t}}dz$
and $\psi_{t}(w)=\Psi_{t}(\norm{\w})$, and $\phi_{t}(\w)=\Delta_{t}(\w)=\Psi_{\tpp}(\norm{\w})-\Psi_{t}(\norm{\w})$.
We have by
\Cref{lemma:centered-md} that
\begin{align*}
  R_{T}(\cmp)
  &\le
    \psi_{T+1}(\cmp)+\sumtT\delta_{t}\\
  &\overset{(a)}{\le}
    \norm{\cmp}\Psi_{T+1}'(\norm{\cmp}) + \sumtT\delta_{t}\\
  &\overset{(b)}{\le}
    2k\norm{\cmp}\Max{\sqrt{V_{T+1}\Log{\norm{\cmp}/\alpha_{T+1}+1}},G\Log{\norm{\cmp}/\alpha_{T+1}+1}}+ \sumtT\delta_{t}
\end{align*}
where $(a)$ observes that $\Psi_{T+1}'(x)$ is non-decreasing in $x$, so
\begin{align*}
\psi_{T+1}(\cmp)=\int_{0}^{\norm{\cmp}}\Psi_{T+1}'(x)dx\le\int_{0}^{\norm{\cmp}}dx\Psi_{T+1}'(\norm{\cmp})=\norm{\cmp}\Psi_{t}'(\norm{\cmp}),
\end{align*}
and $(b)$
observes that
$V_{t}/G\le GF_{t}(x)$ whenever $\Psi_{t}'(x)=kGF_{t}(x)+\frac{kV_{t}}{G}$ and hence
\begin{align*}
  \Psi_{T+1}'(\norm{\cmp})
  &=
    \begin{cases}
      2k\sqrt{V_{T+1}F_{T+1}(\norm{\cmp})}&\text{if }G\sqrt{F_{T+1}(\norm{\cmp})}\le\sqrt{V_{T+1}}\\
      kGF_{T+1}(\norm{\cmp})+\frac{kG}{V_{T+1}}&\text{otherwise}
    \end{cases}\\
  &\le
    \begin{cases}
      2k\sqrt{V_{T+1}F_{T+1}(\norm{\cmp})}&\text{if }G\sqrt{F_{T+1}(\norm{\cmp})}\le\sqrt{V_{T+1}}\\
      2kGF_{T+1}(\norm{\cmp})&\text{otherwise}
    \end{cases}\\
  &=2k\Max{\sqrt{V_{T+1}F_{T+1}(\norm{\cmp})},GF_{T+1}(\norm{\cmp})}.
\end{align*}
Thus, we need only bound the stability terms
$\sumtT\delta_{t}$, which we will handle using the Stability Lemma
(\Cref{lemma:md-stability}).

For any $x> 0$, we have
\begin{align*}
  \Psi_{t}'(x)
  &=
    \begin{cases}
      2k\sqrt{V_{t}\psiq_{t}(x)}&\text{if }G\sqrt{\psiq_{t}(x)}\le\sqrt{V_{t}}\\
      kG\psiq_{t}(x)+\frac{kV_{t}}{G}&\text{otherwise}
    \end{cases}\\
  \Psi_{t}''(x)
  &=
    \begin{cases}
      \frac{k\sqrt{V_{t}}}{(x+\alpha_{t})\sqrt{\psiq_{t}(x)}}&\text{if }G\sqrt{\psiq_{t}(x)}\le\sqrt{V_{t}}\\
      \frac{kG}{x+\alpha_{t}}&\text{otherwise}
    \end{cases}\\
  \Psi_{t}'''(x)
  &=
    \begin{cases}
      \frac{-k\sqrt{V_{t}}\brac{1+2\psiq_{t}(x)}}{2(x+\alpha_{t})^{2}\psiq_{t}(x)^{3/2}}&\text{if }G\sqrt{\psiq_{t}(x)}\le\sqrt{V_{t}}\\
      \frac{-kG}{(x+\alpha_{t})^{2}}&\text{otherwise}
    \end{cases}.
\end{align*}
Clearly $\Psi_{t}(x)\ge 0$, $\Psi_{t}'(x)\ge 0$, $\Psi_{t}''(x)\ge 0$,
$\Psi_{t}'''(x)\le 0$ for all $x> 0$.
Moreover,
observe that for any $x> \alpha_{t}(e-1)=:x_{0}$, we have
\begin{align*}
   \frac{\abs{\Psi_{t}'''(x)}}{\Psi_{t}''(x)^{2}}
  &=
    \begin{cases}
      \frac{k\sqrt{V_{t}}(1+2\psiq_{t}(x))}{2(x+\alpha_{t})^{2}\psiq_{t}(x)^{3/2}}\frac{(x+\alpha_{t})^{2}\psiq_{t}(x)}{k^{2}V_{t}}&\text{if
      } G\sqrt{\psiq_{t}(x)}\le\sqrt{V_{t}}\\
      \frac{kG}{(x+\alpha_{t})^{2}}\frac{(x+\alpha_{t})^{2}}{k^{2}G^{2}}&\text{otherwise}
    \end{cases}\\
  &=
    \begin{cases}
      \frac{1}{2k\sqrt{V_{t}}}\brac{\frac{1}{\sqrt{\psiq_{t}(x)}}+2\sqrt{\psiq_{t}(x)}}&\text{if
      } G\sqrt{\psiq_{t}(x)}\le\sqrt{V_{t}}\\
      \frac{1}{kG}&\text{otherwise}
    \end{cases}
    \intertext{Now, since $x>\alpha_t(e-1)$, we have $\psiq_t(x)> 1$ so that $\frac{1}{\sqrt{\psiq_{t}(x)}}\le \sqrt{\psiq_t(x)}$. Thus:}
  &\le
    \begin{cases}
      \frac{3}{2k}\sqrt{\frac{\psiq_{t}(x)}{V_{t}}}&\text{if
      } G\sqrt{\psiq_{t}(x)}\le\sqrt{V_{t}}\\
      \frac{1}{kG}&\text{otherwise}
    \end{cases}\\
  &\le
    \frac{1}{2}\Min{\sqrt{\frac{\psiq_{t}(x)}{V_{t}}},\frac{1}{G}}
  = \half \eta_{t}'(x),
\end{align*}
where the last line defines
$\eta_{t}(x)=\int_{0}^{x}\Min{\sqrt{\frac{\psiq_{t}(v)}{V_{t}}},\frac{1}{G}}dv$ and uses
$k\ge 3$.
We also have
$\eta_{t}'(x)=\Min{\sqrt{\frac{\psiq_{t}(x)}{V_{t}}},\frac{1}{G}}\le\frac{1}{G}$,
and $\eta_{t}'(x)$ is monotonic, so $\eta_{t}(x)$ is convex and $1/G$ Lipschitz.
Hence, by \Cref{lemma:md-stability} we have
\begin{align}
  \hat\delta_{t}=\inner{g_{t},\wt-\wtpp}-D_{\psi_{t}}(\wtpp|\wt)-\eta_{t}(\norm{\wtpp})\norm{g_{t}}^{2}\le\frac{2\norm{g_{t}}^{2}}{\Psi_{t}''(x_{0})}\label{eq:pf-eq-1}
\end{align}
with $x_{0}=\alpha_{t}(e-1)$.

Next, we want to show that
$\phi_{t}(\w)=\Delta_{t}(\w)\ge \eta_{t}(\norm{\w})\norm{g_{t}}^{2}$,
so that $\delta_{t}\le\hat\delta_{t}$.
To this end, let $x> 0$ and observe
that for $\alpha_{\tpp}\le \alpha_{t}$, we have
$F_{\tpp}(x)=\Log{x/\alpha_{\tpp}+1}\ge\Log{x/\alpha_{t}+1}=F_{t}(x)$, so
\begin{align*}
  \Psi_{\tpp}'(x)-\Psi_{t}'(x)
  &=
    k\min_{\eta\le\frac{1}{G}}\sbrac{\frac{F_{\tpp}(x)}{\eta}+\eta V_{\tpp}}-k\min_{\eta\le\frac{1}{G}}\sbrac{\frac{F_{t}(x)}{\eta}+\eta V_{t}}\\
  &\ge
    k\min_{\eta\le\frac{1}{G}}\sbrac{\frac{F_{t}(x)}{\eta}+\eta V_{\tpp}}-k\min_{\eta\le\frac{1}{G}}\sbrac{\frac{F_{t}(x)}{\eta}+\eta V_{t}},
    \intertext{and using the fact that for any $\eta\le 1/G$ we can bound
    $\frac{F_{t}(x)}{\eta}+\eta V_{\tpp}=\frac{F_{t}(x)}{\eta}+\eta V_{t}+\eta\norm{g_{t}}^{2}\ge\min_{\eta^{*}\le 1/G}\sbrac{\frac{F_{t}(x)}{\eta^{*}}+\eta^{*} V_{t}}+\eta\norm{g_t}^{2}$,
    we have
    }
    &\ge
      k\norm{g_{t}}^{2}\Min{\sqrt{\frac{F_{t}(x)}{V_{\tpp}}},\frac{1}{G}}+k\min_{\eta\le\frac{1}{G}}\sbrac{\frac{F_{t}(x)}{\eta}+\eta V_{t}}-k\min_{\eta\le\frac{1}{G}}\sbrac{\frac{F_{t}(x)}{\eta}+\eta V_{t}}\\
  &=
    k\norm{g_{t}}^{2}\Min{\sqrt{\frac{F_{t}(x)}{V_{\tpp}}},\frac{1}{G}}
    \ge
    \frac{k}{\sqrt{2}}\norm{g_{t}}^{2}\Min{\sqrt{\frac{F_{t}(x)}{V_{t}}},\frac{1}{G}}\ge \norm{g_{t}}^{2}\eta_{t}'(x),
\end{align*}
where the last line uses $k\ge3$ and
$\frac{1}{V_{t}}=\frac{1}{V_{\tpp}}\frac{V_{\tpp}}{V_{t}}=\frac{1}{V_{\tpp}}\brac{1+\norm{g_{t}}^{2}/V_{t}}\le\frac{2}{V_{\tpp}}$
for $V_{t}\ge \norm{g_{t}}^{2}$. From this, we immediately have
\begin{align*}
  \Delta_{t}(\w)=\int_{0}^{\norm{\w}}\Psi_{\tpp}'(x)-\Psi_{t}'(x)dx \ge\norm{g_{t}}^{2}\int_{0}^{\norm{\w}}\eta_{t}'(x)dx=\eta_{t}(\norm{\w})\norm{g_{t}}^{2},
\end{align*}
and hence
\begin{align*}
  \delta_{t}
  &=
    \inner{g_{t},\wt-\wtpp}-D_{\psi_{t}}(\wtpp|\wt)-\Delta_{t}(\wtpp)\\
  &\le
    \inner{g_{t},\wt-\wtpp}-D_{\psi_{t}}(\wtpp|\wt)-\eta_{t}(\norm{\wtpp})\norm{g_{t}}^{2} = \widehat{\delta}_{t}\le\frac{2\norm{g_{t}}^{2}}{\Psi_{t}''(x_{0})}
\end{align*}
for $x_{0}=\alpha_{t}(e-1)$ via \Cref{eq:pf-eq-1}.
Summing over $t$ then yields
\begin{align*}
  \sumtT\delta_{t}
  &\le
    \sumtT\frac{2\norm{g_{t}}^{2}}{\Psi_{t}''(\alpha_{t}(e-1))}
  \le
    \sumtT\frac{2e\alpha_{t}}{k}\norm{g_{t}}^{2}\sqrt{\frac{\psiq_{t}(\alpha_{t}(e-1))}{V_{t}}}\\
  &\le
    \sumtT\frac{2e\alpha_{t}}{k}\norm{g_{t}}^{2}\frac{1}{\sqrt{V_{t}}}
    \le
    \sumtT\frac{6}{k}\frac{\alpha_{t}\norm{g_{t}}^{2}}{\sqrt{V_{t}}}\\
  &\overset{(a)}{\le}
    2G\epsilon\sumtT\frac{\norm{g_{t}}^{2}}{V_{t}\log^{2}(V_{t}/G^{2})}\\
  &\overset{(b)}{\le}
    4G\epsilon
\end{align*}
where $(a)$ chooses
$\alpha_{t}=\frac{\epsilon G}{\sqrt{V_{t}}\log^{2}(V_{t}/G^{2})}$ and recalls $k\ge 3$, and $(b)$
recalls $V_{t}=4 G^{2}+\norm{g}_{1:t-1}^{2}$ and
uses \Cref{lemma:v-log-sqr-integral-bound} to bound
$\sumtT\frac{\norm{g_{t}}^{2}}{V_{t}\log^{2}(V_{t}/G^{2})}\le 2$.
Returning to our regret bound we have
\begin{align*}
  R_{T}(\cmp)
  &\le
    2k\norm{\cmp}\Max{\sqrt{V_{T+1}\Log{\norm{\cmp}/\alpha_{T+1}+1}},G\Log{\norm{\cmp}/\alpha_{T+1}+1}}
  +\sumtT\delta_{t}\\
  &\le
    4G\epsilon+2k\norm{\cmp}\Max{\sqrt{V_{T+1}\Log{\norm{\cmp}/\alpha_{T+1}+1}},G\Log{\norm{\cmp}/\alpha_{T+1}+1}}\\
  &\le
    \Ohat\brac{G\epsilon+\norm{\cmp}\sbrac{\sqrt{\norm{g}_{1:T}^{2}\Log{\frac{\norm{\cmp}\sqrt{\norm{g}_{1:T}^{2}}}{\epsilon G}+1}}\maxOp G\Log{\frac{\norm{\cmp}\sqrt{\norm{g}_{1:T}^{2}}}{\epsilon G}+1}}}
\end{align*}
\end{proof}

\section{Proofs for Section~\ref{sec:dynamic} (Dynamic Regret)}\label{app:dynamic}

\subsection{Proof of Lemma~\ref{lem:boundediterate}}\label{app:boundediterate}

\FTRLBoundedIterate*
\begin{proof}
Notice that if $\|g_t\|=1$ for all $t$, since $F_t$ is odd, we can write
\begin{align*}
    F_t(g_{1:t-1},\|g_1\|,\dots,\|g_t\|) = -\sign(g_{1:t-1}) F_t(-|g_{1:t-1}|,1,1,\dots,1)
\end{align*} which justifies our simpler notation of dropping the dependence on $\|g_t\|$ from $F_t$ in this setting.  Further, since $F_{t}(-x)$ is non-decreasing for positive $x$, it suffices to show that $F_t(-C\sqrt{t}/2)\le \frac{2\epsilon}{C^2}$.

Let $g_1,\dots,g_{t-1}$ all be independent random signs, and let
$g_t=\sign(F_t(g_{1:t-1}))$. Notice that since $F_t$ is odd, we can write
$F_t(g_{1:t-1}) = -\sign(g_{1:t-1}) F_t(-|g_{1:t-1}|)$. Further, notice that
$|g_{1:t-1}|$ satisfies $\E[|g_{1:t-1}|]\ge C\sqrt{t}$ for some absolute constant
$C$, and $\E[|g_{1:t-1}|^2]\le t$. Thus, by Paley-Zygmund inequality, $P[|g_{1:t-1}|>\theta C\sqrt{t}]\ge (1-\theta)^2C^2$ for all $\theta$. In particular, setting $\theta=1/2$ yields $P[|g_{1:t-1}|>C/2\sqrt{t}]\ge C^2/4$

Then we have:
\begin{align*}
    \epsilon \ge &\E\left[\sum_{i=1}^t w_i g_i\right] \\
    \intertext{using $\E[g_i]=0$ for $i<t$:}
    &=\E[w_t g_t]
    \intertext{Using $g_tw_t = |F_t(g_{1:t-1})|=F(-|g_{1:t-1}|)$:}
    &= \E\left[F_t(-|g_{1:t-1}|)\right]\\
    &\ge \sum_{S \in \mathbb{N}} F_t(-S)P(|g_{1:t-1}|=S)
    \intertext{using $F_t(-|x|)\ge 0$ for all $x$:}
    &\ge 2\sum_{S\ge C/2\sqrt{t}}F_t(-S)P(|g_{1:t-1}| = S)\\
    &\ge 2F_t(-C\sqrt{t}/2)P(|g_{1:t-1}| \ge C\sqrt{t}/2)\\
    &\ge \frac{C^2}{2} F_t(-C\sqrt{t}/2)
    \intertext{rearranging:}
    \frac{2 \epsilon }{C^2}&\ge F_t(-C\sqrt{t}/2)
\end{align*}
\end{proof}

\subsection{Proof of Theorem~\ref{thm:ftrl-unconstrained-lb}}\label{app:ftrl-unconstrained-lb}

\begin{manualtheorem}{\ref{thm:ftrl-unconstrained-lb}}
  Let $C$ be the universal constant from Lemma~\ref{lem:boundediterate}. Suppose
  an algorithm $\cA$ satisfying the conditions [\ref{propertyone},
  \ref{propertytwo}, \ref{propertythree}, \ref{propertyfour}] also guarantees
  $\sum_{i=1}^t g_t w_t \le \epsilon$ for all $t$ for some $\epsilon$ if $\|g_t\|=1$
  for all $t$. Then,
  for all $T$ large enough that:
  \begin{enumerate*}
      \item $C\sqrt{T/2}\ge 2$
      \item $T-\lfloor C\sqrt{T/2}/2\rfloor\ge T/2$
      \item $2+\frac{2}{C}\sqrt{2T}\le \frac{4}{C}\sqrt{T}$
  \end{enumerate*}
  there exists a sequence $\{g_t\}$ and a comparator sequence $\{u_{t}\}$ that does not depend on $\cA$ such that such that:
  \begin{align*}
      \frac{\epsilon\sqrt{2T}}{C^3}\le P_T+\max_t |u_t|&\le \frac{8\epsilon\sqrt{T}}{C^3}\\
      \sum_{t=1}^T g_t(w_t-u_t)&\ge \frac{T\epsilon}{2C^2}\\
      &\ge \frac{C}{16} (P_T+\max_t |u_t|) \sqrt{T}
  \end{align*}
\end{manualtheorem}
\begin{proof}
Consider the sequence $g_t= (-1)^{t+1}$ for $t\in[1,\lceil T/2\rceil]$, and afterwards for all natural numbers $k$ and $j$ with $j<\lfloor C\sqrt{T/2}/2\rfloor$: $g_{\lceil T/2\rceil+k\lfloor C\sqrt{T/2}/2\rfloor + j}=(-1)^k$. Notice that $|g_t|=1$ for all $t$, and also, for $t\le  \lceil T/2\rceil-1$, we have $g_{1:t-1}=0$ for all odd $t\le  T/2-1$, so that $w_t=0$ for all odd $t$. Next, for even $t\le  \lceil T/2\rceil-1$, $g_{1:t-1}=1$ so that $w_t\le 0$ and so $w_tg_t=-w_t\ge 0$. Therefore $\sum_{t=1}^{\lceil T/2\rceil-1}w_t g_t\ge 0$.

Now, let's consider the indices $t\ge \lceil T/2\rceil$. Observe that our sequence satisfies $g_{1:t}\ge 0$ for all $t$ so that $w_t\le 0$ for all $t$. Further, $|g_{1:t}|\le \lfloor C\sqrt{T/2}/2\rfloor\le C\sqrt{t}/2$ for all $t\ge \lceil T/2\rceil$. Thus, by Lemma~\ref{lem:boundediterate}, we have that $-\frac{2\epsilon}{C^2}\le w_t$ for all $t$. Now, let $S^+$ be the indices $t$ such that $t=\lceil T/2\rceil+k\lfloor C\sqrt{T/2}/2\rfloor + j$ for even $k$ and $j\in [0,\lfloor C\sqrt{T/2}/2\rfloor]$, and let $S^-$ be the indices with odd $k$. Then for $t\in S^+$, $g_t=1$ and for $t\in S^-$, $g_t=-1$. Further, $|S^+|\le |S^-|+\lfloor C\sqrt{T/2}/2\rfloor$. Thus, since $|S^+|+|S^-| = T-\lceil T/2\rceil + 1\ge T/2$, we have $|S^+|\ge \frac{T-\lfloor C\sqrt{T/2}/2\rfloor}{2}$.

Putting all these observations together:
\begin{align*}
    \sum_{t=\lceil T/2\rceil}^T g_t w_t&\ge \sum_{t\in S^+}-\frac{2\epsilon}{C^2}+\sum_{t\in S^-} -w_t
    \intertext{Using the fact that $w_t\in[-2\epsilon/C^2, 0]$:}
    &\ge -\frac{2\epsilon}{C^2}|S^+|\\
    &\ge -\frac{2\epsilon}{C^2}\cdot \frac{T-\lfloor C\sqrt{T/2}/2\rfloor}{2}\\
    &\ge -\frac{T \epsilon}{2C^2}
\end{align*}

Next, consider the comparator sequence $u_t=0$ for $t\le \lceil T/2\rceil-1$, and $u_t=-2\epsilon/C^2$ for $t\in S^+$ and $u_t=2\epsilon/C^2$ for $t\in S^-$. Notice that the path length is
\begin{align*}
    P_T+\max_t |u_t|&=\frac{2\epsilon}{C^2}+\sum_{t=1}^{T-1}|u_t-u_{t+1}|\\
    &\le \frac{2\epsilon}{C^2}\left(1+ \left\lceil \frac{\lceil T/2\rceil+1}{2\lfloor C\sqrt{T/2}/2\rfloor}\right\rceil\right)\\
    &\le \frac{2\epsilon}{C^2}\left(2+ \frac{T/2}{\lfloor C\sqrt{T/2}/2\rfloor}\right)\\
    &\le \frac{2\epsilon}{C^2}\left(2+\frac{T}{ C\sqrt{T/2}-1}\right)\\
    &\le \frac{2\epsilon}{C^2}\left(2+ \frac{2T}{ C\sqrt{T/2}}\right)\\
    &\le \frac{2\epsilon}{C^2}\left(2+\frac{2}{C}\sqrt{2T}\right)\\
    &\le \frac{8\epsilon\sqrt{T}}{C^3}
\end{align*}
Similarly, we have
\begin{align*}
P_T + \max_t |u_t|&\ge \frac{2\epsilon}{C^2} \left( 1+ \left\lfloor \frac{T/2}{2\lfloor C\sqrt{T/2}/2\rfloor}\right\rfloor\right)\\
&\ge  \frac{2\epsilon}{C^2} \left(  \frac{T}{2 C\sqrt{T/2}}\right)\\
&\ge \frac{\epsilon\sqrt{2T}}{C^3}
\end{align*}

Further,
\begin{align*}
    \sum_{t=1}^T g_tu_t=-\frac{2(T-\lceil T/2\rceil +1)\epsilon}{C^2}\le -\frac{T\epsilon}{C^2}
\end{align*}

Thus, overall we obtain dynamic regret:
\begin{align*}
    \sum_{t=1}^T g_t(w_t-u_t)&\ge \frac{T\epsilon}{2C^2}
\end{align*}
Substituting the bound on $P_T+\max_t|u_t|$ completes the argument.
\end{proof}

\subsection{Proof of Proposition \ref{prop:dynamic-fixed-eta}}\label{app:dynamic-fixed-eta}

We break the proof of \Cref{prop:dynamic-fixed-eta} into parts; we first
derive a partial result in \Cref{prop:partial-dynamic}, and then make particular choices for
the unspecified parameters $\alpha_{t}$ and $b_{t}$.

\begin{restatable}{myproposition}{PartialDynamic}\label{prop:partial-dynamic}
  $(\alpha_{t})_{t=1}^{T}$ be a non-increasing sequence
  and consider \Cref{alg:centered-md} with
  \begin{align*}
    \psi_{t}(w)&=2\int_{0}^{\norm{\w}}\frac{\Log{x/\alpha_{t}+1}}{\eta}dx\\
    \varphi_{t}(w)&=\brac{\eta\norm{g_{t}}^{2}+b_{t}}\norm{\w},
  \end{align*}
  where $b_{t}\ge 0$ and $\eta\le\frac{1}{G}$.
  Then for all $\cmp_{1},\ldots,\cmp_{T}$ in $\R^{d}$, \Cref{alg:centered-md} guarantees
  \begin{align*}
    R_{T}(\vec{\cmp})
    &\le
    \frac{2M\Log{M/\alpha_{T+1}+1}}{\eta}+\sum_{t=1}^{T-1}\sbrac{\frac{2\norm{\cmp_{\tpp}-\cmp_{t}}\Log{\norm{\wtpp}/\alpha_{\tpp}+1}}{\eta}-b_{t}\norm{\wtpp}}\\
  &\qquad
    +\sumtT\brac{\eta\norm{g_{t}}^{2}+b_{t}}\norm{\cmp_{t}}
    +\eta\sumtT\alpha_{t}\norm{g_{t}}^{2},
  \end{align*}
  where $M=\max_{t}\norm{\cmp_{t}}$.
\end{restatable}
\begin{proof}
  Using \Cref{lemma:centered-md}
  we have
\begin{align*}
  R_{T}(\vec{\cmp})
  &\le
    \psi_{T+1}(\cmp_{T})+\sum_{t=1}^{T-1}\rho_{t}+\sumtT\varphi_{t}(\cmp_{t})+\sumtT\delta_{t}\\
  &\le
    \frac{2\norm{\cmp_{T}}\Log{\norm{\cmp_{T}}/\alpha_{T+1}+1}}{\eta}+\sum_{t=1}^{T-1}\rho_{t}+\sumtT\varphi_{t}(\cmp_{t})+\sumtT\delta_{t}\\
  &\le
    \frac{2M\Log{M/\alpha_{T+1}+1}}{\eta}+\sum_{t=1}^{T-1}\rho_{t}+\sumtT\varphi_{t}(\cmp_{t})+\sumtT\delta_{t}\\
\end{align*}
where $M=\max_{t\le T}\norm{\cmp_{t}}$ and
\begin{align*}
  \sum_{t=1}^{T-1}\rho_{t}
  &=
    \sum_{t=1}^{T-1}\inner{\grad\psi_{\tpp}(\wtpp),\cmp_{t}-\cmp_{\tpp}}
    \le
    \sum_{t=1}^{T-1}\norm{\grad\psi_{\tpp}(\wtpp)}\norm{\cmp_{t}-\cmp_{\tpp}}\\
  &=
    2\sum_{t=1}^{T-1}\frac{\Log{\norm{\wtpp}/\alpha_{\tpp}+1}}{\eta}\norm{\cmp_{t}-\cmp_{\tpp}}\\
  \sumtT\delta_{t}
  &=
    \sumtT\inner{g_{t},\wt-\wtpp}-D_{\psi_{t}}(\wtpp|\wt)-\phi_{t}(\wtpp)\\
  &=
    \sumtT\inner{g_{t},\wt-\wtpp}-D_{\psi_{t}}(\wtpp|\wt)-\Delta_{t}(\wtpp)-\varphi_{t}(\wtpp)
\end{align*}
First consider the terms $\sumtT\delta_{t}$.
Since $(\alpha_{t})_{t=1}^{T}$ is a non-increasing sequence, we have
$\Delta_{t}(\wtpp)=\psi_{\tpp}(\wtpp)-\psi_{t}(\wtpp)\ge 0$ and
\begin{align*}
  \delta_{t}
  &=
    \inner{g_{t},\wt-\wtpp}-D_{\psi_{t}}(\wtpp|\wt)-\Delta_{t}(\wtpp)-\varphi_{t}(\wtpp)\\
  &\le
    \inner{g_{t},\wt-\wtpp}-D_{\psi_{t}}(\wtpp|\wt)-\varphi_{t}(\wtpp)\\
  &=
    \inner{g_{t},\wt-\wtpp}-D_{\psi_{t}}(\wtpp|\wt)-\eta\norm{g_{t}}^{2}\norm{\wtpp}-b_{t}\norm{\wtpp}.
\end{align*}
We proceed by showing that the regularizers $\psi_{t}(\cdot)$
satisfy the conditions of \Cref{lemma:md-stability}. we
have $\psi_{t}(\w)=\Psi_{t}(\norm{\w})=2\int_{0}^{\norm{\w}}\frac{\Log{x/\alpha_{t}+1}}{\eta}dx$
and
\begin{align*}
  \Psi_{t}'(x)=2\frac{\Log{x/\alpha_{t}+1}}{\eta}, \qquad
  \Psi_{t}''(x)=\frac{2}{\eta\brac{x+\alpha_{t}}}, \qquad
  \Psi_{t}'''(x)=\frac{-2}{\eta\brac{x+\alpha_{t}}^{2}},
\end{align*}
so $\Psi_{t}(x)\ge 0$, $\Psi_{t}'(x)\ge 0$, $\Psi_{t}''(x)\ge 0$, and $\Psi_{t}'''(x)\le 0$ for all $x>0$.
Moreover,
\begin{align*}
  \frac{-\Psi_{t}'''(x)}{\Psi_{t}''(x)^{2}}
  &=
    \frac{2}{\eta(x+\alpha_{t})^{2}}\frac{\eta^{2}(x+\alpha_{t})^{2}}{2^{2}}=\frac{\eta}{2},
\end{align*}
so assuming $\eta\le\frac{1}{G}$ and letting $\eta_{t}(\norm{\w})=\eta\norm{\w}$,
we have $\abs{\Psi_{t}'''(x)}\le\frac{\eta_{t}'(x)}{2}\Psi_{t}''(x)^{2}$ for all $x> 0$,
and $\eta_{t}(x)$ is a $1/G$ Lipschitz convex function. Hence,
using \Cref{lemma:md-stability} we have
\begin{align*}
  \delta_{t}
  &\le
    \inner{g_{t},\wt-\wtpp}-D_{\psi_{t}}(\wtpp|\wt)-\eta\norm{g_{t}}^{2}\norm{\wtpp}-b_{t}\norm{\wtpp}\\
  &=
    \inner{g_{t},\wt-\wtpp}-D_{\psi_{t}}(\wtpp|\wt)-\eta_{t}(\norm{\wtpp})\norm{g_{t}}^{2}-b_{t}\norm{\wtpp}\\
  &\le
    \frac{2\norm{g_{t}}^{2}}{\Psi_{t}''(0)}-b_{t}\norm{\wtpp}=\eta\alpha_{t}\norm{g_{t}}^{2}-b_{t}\norm{\wtpp}.
\end{align*}
Plugging this back into the full regret bound we have
\begin{align*}
R_{T}(\vec{\cmp})
  &\le
    \frac{2M\Log{M/\alpha_{T+1}+1}}{\eta}+2\sum_{t=1}^{T-1}\frac{\norm{\cmp_{t}-\cmp_{\tpp}}\Log{\norm{\wtpp}/\alpha_{\tpp}+1}}{\eta}+\sumtT\varphi_{t}(\cmp_{t})\\
  &\qquad
    +\sumtT\eta\alpha_{t}\norm{g_{t}}^{2}-b_{t}\norm{\wtpp}\\
  &=
    \frac{2M\Log{M/\alpha_{T+1}+1}}{\eta}+\sum_{t=1}^{T-1}\sbrac{\frac{2\norm{\cmp_{\tpp}-\cmp_{t}}\Log{\norm{\wtpp}/\alpha_{\tpp}+1}}{\eta}-b_{t}\norm{\wtpp}}\\
  &\qquad
    +\sumtT\brac{\eta\norm{g_{t}}^{2}+b_{t}}\norm{\cmp_{t}}
    +\eta\sumtT\alpha_{t}\norm{g_{t}}^{2}.
\end{align*}

\end{proof}

With this result in hand, we prove \Cref{prop:dynamic-fixed-eta} by choosing
values $\alpha_{t}=\frac{\epsilon G^{2}}{V_{t}\log^{2}(V_{t}/G^{2})}$ and
$b_{t}=\eta\norm{g_{t}}^{2}$. The full version of the result is given below.

\begin{manualproposition}{\ref{prop:dynamic-fixed-eta}}
  Let $\ell_{1},\ldots,\ell_{T}$ be $G$-Lipschitz convex functions and
  $g_{t}\in\partial\ell_{t}(\wt)$ for all $t$.
  Let $\epsilon > 0$, $V_{t}=4G^{2}+\norm{g}_{1:\tmm}^{2}$, and
  $\alpha_{t}=\frac{\epsilon G^{2}}{V_{t}\log^{2}\brac{V_{t}/G^{2}}}$.
  For all $t$, set
  \(
  \psi_{t}(\w)=2\int_{0}^{\norm{\w}}\frac{\Log{x/\alpha_{t}+1}}{\eta}dx
  \), and
  \(
  \varphi_{t}(\w)=2\eta\norm{g_{t}}^{2}\norm{\w}.
  \)
  Then after each round \Cref{alg:centered-md}
  updates
  \begin{align*}
        \theta_t & = \nabla \psi_t(\wt) - g_t\\
      \wtpp &= \frac{\alpha_{t+1}\theta_t}{\|\theta_t\|}\left[\exp\left [\frac{\eta}{2}\max\left(\|\theta_t\|-2\eta\|g_t\|^2,0\right)\right]-1\right]
  \end{align*}
  where we we define $C\frac{x}{\|x\|}=\zeros$ for all $C$ when $x=\zeros$.
  Moreover,
  for any $\cmp_{1},\ldots,\cmp_{T}$ in $\R^{d}$, \Cref{alg:centered-md} guarantees
  \begin{align*}
    R_{T}(\vec{\cmp})
    &\le
    2\epsilon G+\frac{4\brac{M+P_{T}}\sbrac{\Log{\frac{9MT^{2}}{4\alpha_{T+1}}+1}\maxOp 1}}{\eta}
      +2\eta\sumtT\norm{g_{t}}^{2}\norm{\cmp_{t}}.
  \end{align*}
  where $M=\max_t\norm{\cmp_{t}}$.
\end{manualproposition}
\begin{proof}
First, we will verify the update equation, and then show the regret bound.
  To compute the update, observe that from the first-order optimality
  conditions, there is some $\nabla \phi_t(w_{t+1})\in \partial \phi_t(w_{t+1})$ such that
  \begin{align*}
    g_{t}+\grad \psi_{t}(\wtpp)-\grad \psi_{t}(\wt)+\grad \phi_{t}(\wtpp)=\zeros
\end{align*}
Now, notice that we can write $\grad \phi_t(\wtpp) = \nabla \psi_{t+1}(\wtpp) - \nabla \psi_t(\wtpp)  + \nabla \varphi_t(\wtpp)$ for some $\nabla \varphi_t(\wtpp)\in \partial \varphi_t(\wtpp)$. Thus, we have:
\begin{align*}
    g_{t}+\grad\psi_{\tpp}(\wtpp)-\grad\psi_{t}(\wt) + \nabla \varphi_t(\wtpp)=\zeros
\end{align*}
Moreover, \emph{any} value for $\wtpp$ such that there is a $\varphi_t(\wtpp)\in \partial \varphi_t(\wtpp)$ satisfying the above condition is valid solution to the mirror descent update. We justify our update equation in two cases. 

First, consider the case $\max(\|\theta_t\|-2\eta \|g_t\|,0)=0$, In this case, the
update equation suggests $\wtpp=\zeros$. To justify this, notice that
$\partial \varphi_t(\zeros)$ consists of all vectors of norm at most
$2\eta \norm{g_t}^2$. Further, $\grad \psi_{\tpp}(\zeros)=\zeros$.
Thus, whenever $\max(\|\theta_t\|-2\eta \|g_t\|,0)=0$, we can set $\wtpp=\zeros$ as described by our update.

Now, let us suppose $\max(\|\theta_t\|-2\eta \|g_t\|,0)=\|\theta_t\|-2\eta\|g_t\|>0$. Note that this implies $\theta_t\ne \zeros$, and the update equation sets $\wtpp\ne \zeros$. In the case $\wtpp\ne \zeros$, $\varphi_t(\wtpp)$ is differentiable so that $\varphi_t(\wtpp) = 2\eta\norm{g_{t}}^{2}\frac{\wtpp}{\norm{\wtpp}}$. Thus, we need to establish that indeed a non-zero $\wtpp$ given by the update equation is a solution to the optimality condition:
\begin{align*}
      g_{t}+\grad\psi_{\tpp}(\wtpp)-\grad\psi_{t}(\wt)+2\eta\norm{g_{t}}^{2}\frac{\wtpp}{\norm{\wtpp}}
    =
      \zeros.
\end{align*}

  Writing
  $\psi_{t}(\w)=\Psi_{t}(\norm{\w})=\int_{0}^{\norm{\w}}\Psi_{t}'(x)dx$, we have
  $\grad\psi_{\tpp}(\wtpp)=\frac{\wtpp}{\norm{\wtpp}}\Psi_{t+1}'(\norm{\wtpp})$ (where we define $\frac{\wtpp}{\norm{\wtpp}}\cdot 0=\zeros$) and hence the optimality condition can be re-written:
  \begin{align*}
      \frac{\wtpp}{\norm{\wtpp}}\sbrac{\Psi_{t+1}'(\norm{\wtpp})+2\eta\norm{g_{t}}^{2}}
    &=
      \grad\psi_{t}(\wt)-g_{t}
    = \theta_t
  \end{align*}
 Now we need only verify that our expression $\wtpp = \frac{\alpha_{t+1}\theta_t}{\|\theta_t\|}\left[\exp\left[\frac{\eta}{2}(\|\theta_t\| - 2\eta\|g_t\|)\right]-1\right]$ satisfies this condition. Fortunately, this is easily checked by observing the stated update satisfies:
 \begin{align*}
     \Psi_{t+1}'(\norm{\wtpp}) &= \frac{2}{\eta} \log(\|\wtpp\|/\alpha_{t+1}+1) = \|\theta_t\|-2\eta\|g_t\|^2.
 \end{align*}

\vspace{2 em}

  Turning now to the regret,
we begin by replacing the comparator sequence with an auxiliary sequence
$\hat \cmp_{1},\ldots,\hat\cmp_{T}$ to be determined later. This alternative sequence will eventually be designed to have some useful stability properties while still being ``close'' to the real sequence $\cmp_1,\dots,\cmp_T$:
\begin{align*}
  R_{T}(\vec{\cmp})
  &=
    \sumtT\inner{g_{t},\wt-\cmp_{t}}
    =
    \sumtT\inner{g_{t},\wt-\hat\cmp_{t}}+\sumtT\inner{g_{t},\hat\cmp_{t}-\cmp_{t}}\\
    &\le
    R_{T}(\vec{\hat\cmp})+\sumtT\norm{g_{t}}\norm{\hat\cmp_{t}-\cmp_{t}}
\end{align*}
The first term is bounded via \Cref{prop:partial-dynamic} as
\begin{align*}
  R_{T}(\vec{\hat\cmp})
  &\le
    \frac{2\hat{M}\Log{\hat{M}/\alpha_{T+1}+1}}{\eta}
    +2\eta\sumtT\norm{g_{t}}^{2}\norm{\hat\cmp_{t}}
    +\eta\sumtT\alpha_{t}\norm{g_{t}}^{2}\\
  &\qquad
    +\sum_{t=1}^{T-1}\sbrac{\frac{2\norm{\hat\cmp_{t}-\hat\cmp_{\tpp}}\Log{\norm{\wtpp}/\alpha_{\tpp}+1}}{\eta}-\eta\norm{g_{t}}^{2}\norm{\wtpp}}\\
\end{align*}
where $\hat{M}=\max_{t\le T}\norm{\hat\cmp_{t}}$.
We focus first on
bounding the sum in the second line. To do so, we first provide the definition of $\hat\cmp_t$:

\begin{center}
Let $\cT>0$ and set
$\hat \cmp_{T}=\cmp_{T}$ and
$\hat\cmp_{t}=\begin{cases}\cmp_{t}&\text{if
  }\norm{g_{t}}\ge\cT\\\hat\cmp_{\tpp}&\text{otherwise}\end{cases}$ for $t<T$.
\end{center}

\noindent Hence, by definition we have $\norm{\hat\cmp_{t}-\hat\cmp_{\tpp}}=0$ whenever
$\norm{g_{t}}\le \cT$, so
\begin{align*}
  &\sum_{t=1}^{T-1}\sbrac{\frac{2\norm{\hat\cmp_{t}-\hat\cmp_{\tpp}}\Log{\norm{\wtpp}/\alpha_{\tpp}+1}}{\eta}-\eta\norm{g_{t}}^{2}\norm{\wtpp}}\\
  &\qquad\le
  \sum_{t:\norm{g_{t}}\ge \cT}\sbrac{\frac{2\norm{\hat\cmp_{t}-\hat\cmp_{\tpp}}\Log{\norm{\wtpp}/\alpha_{\tpp}+1}}{\eta}-\eta\cT^{2}\norm{\wtpp}}\\
  &\qquad\le
  \sum_{t:\norm{g_{t}}\ge \cT}\sbrac{\sup_{X\ge 0}\frac{2\norm{\hat\cmp_{t}-\hat\cmp_{\tpp}}\Log{X/\alpha_{\tpp}+1}}{\eta}-\eta\cT^{2}X}\\
  &\qquad\overset{(*)}{\le}
    \sum_{t:\norm{g_{t}}\ge \cT}\frac{2\norm{\hat\cmp_{t}-\hat\cmp_{\tpp}}\Log{\frac{2\norm{\hat\cmp_{t}-\hat\cmp_{\tpp}}}{\alpha_{\tpp}\eta^{2}\cT^{2}}}}{\eta}
\end{align*}
where $(*)$ observes that either the max is obtained at $X=0$,
for which
\(
  \sup_{X\ge 0}\frac{2\norm{\hat\cmp_{t}-\hat\cmp_{\tpp}}\Log{X/\alpha_{\tpp}+1}}{\eta}-\eta\cT^{2}X=0,
\)
and otherwise the max is obtained at
$X=\frac{2\norm{\hat\cmp_{\tpp}-\cmp_{t}}}{\eta^{2}\cT^{2}}-\alpha_{\tpp}> 0$,
which leads to an upperbound of
\begin{align*}
\sup_{X\ge 0}\frac{2\norm{\hat\cmp_{t}-\hat\cmp_{\tpp}}\Log{X/\alpha_{\tpp}+1}}{\eta}-\eta\cT^{2}X
\le\frac{2\norm{\hat\cmp_{t}-\hat\cmp_{\tpp}}\Log{\frac{2\norm{\hat\cmp_{\tpp}-\hat\cmp_{t}}}{\alpha_{\tpp}\eta^{2}\cT^{2}}}}{\eta}
\end{align*}
in both cases. Moreover,
for any $t$ such that $\norm{g_{t}}\ge \cT$ let $t'$ denote the
smallest index greater than $t$ for which $\norm{g_{t'}}\ge \cT$;
then by triangle inequality we have
$\norm{\hat\cmp_{t}-\hat\cmp_{\tpp}}=\norm{\cmp_{t}-\cmp_{t'}}\le\sum_{s=t}^{t'}\norm{\cmp_{s}-\cmp_{s+1}}$
and
\begin{align*}
  \sum_{t:\norm{g_{t}}\ge \cT}\frac{2\norm{\hat\cmp_{t}-\hat\cmp_{\tpp}}\Log{\frac{2\norm{\hat\cmp_{t}-\hat\cmp_{\tpp}}}{\alpha_{\tpp}\eta^{2}\cT^{2}}}}{\eta}
  &\le
    \sum_{t:\norm{g_{t}}\ge \cT}\frac{\sum_{s=t}^{t'}2\norm{\cmp_{s}-\cmp_{s+1}}\Log{\frac{4\hat M}{\alpha_{T+1}\eta^{2}\cT^{2}}}}{\eta}\\
  &=
    \frac{2P_{T}\Log{\frac{4\hat M}{\alpha_{T+1}\eta^{2}\cT^{2}}}}{\eta}.
\end{align*}
Returning to the regret against the auxiliary comparator sequence we have
\begin{align*}
  R_{T}(\vec{\hat\cmp})
  &\le
    \frac{2\hat{M}\Log{\hat{M}/\alpha_{T+1}+1}}{\eta}
    +2\eta\sumtT\norm{g_{t}}^{2}\norm{\hat\cmp_{t}}
    +\eta\sumtT\alpha_{t}\norm{g_{t}}^{2}+ \frac{2P_{T}\Log{\frac{4\hat M}{\alpha_{\tpp}\eta^{2}\cT^{2}}}}{\eta}\\
  &\overset{(a)}{\le}
    \frac{2M\Log{M/\alpha_{T+1}+1}+2P_{T}\Log{\frac{4M}{\alpha_{\tpp}\eta^{2}\cT^{2}}}}{\eta}
    +2\eta\sumtT\norm{g_{t}}^{2}\norm{\hat\cmp_{t}}
    +\eta\sumtT\alpha_{t}\norm{g_{t}}^{2}\\
  &\overset{(b)}{\le}
    \frac{2M\Log{M/\alpha_{T+1}+1}+2P_{T}\Log{\frac{4M}{\alpha_{\tpp}\eta^{2}\cT^{2}}}}{\eta}
    +\eta\sumtT\alpha_{t}\norm{g_{t}}^{2}\\
    &\qquad
      +2\eta\sumtT\norm{g_{t}}^{2}\norm{\cmp_{t}}+2\sumtT\norm{g_{t}}\norm{\hat\cmp_{t}-\cmp_{t}}\\
  &\overset{(c)}{\le}
    \frac{2M\Log{M/\alpha_{T+1}+1}+2P_{T}\Log{\frac{4M}{\alpha_{\tpp}\eta^{2}\cT^{2}}}}{\eta}
    +2\epsilon G\\
    &\qquad
    +2\eta\sumtT\norm{g_{t}}^{2}\norm{\cmp_{t}}+2\sumtT\norm{g_{t}}\norm{\hat\cmp_{t}-\cmp_{t}},
\end{align*}
where $(a)$ observes that
$\hat{M}=\max_{t\le T}\norm{\hat\cmp_{t}}\le\max_{t\le T}\norm{\cmp_{t}}=M$ and
$(b)$ recalls $\eta\le\frac{1}{G}$ and
uses
$\eta\norm{g_{t}}^{2}\norm{\hat\cmp_{t}}\le\eta\norm{g_{t}}^{2}\brac{\norm{\cmp_{t}-\hat\cmp_{t}}+\norm{\cmp_{t}}}\le\eta\norm{g_{t}}^{2}\norm{\cmp_{t}}+\norm{g_{t}}\norm{\cmp_{t}-\hat\cmp_{t}}$,
and $(c)$ chooses $\alpha_{t}=\frac{\epsilon G^{2}}{V_{t}\log^{2}(V_{t}/G^{2})}$
for $V_{t}=4G^{2}+\norm{g}_{1:\tmm}^{2}$ and applies \Cref{lemma:v-log-sqr-integral-bound}
to bound
\begin{align*}
  \eta\sumtT\alpha_{t}\norm{g_{t}}^{2}
  &=
    \eta\epsilon  G^{2}\sumtT\frac{\norm{g_{t}}^{2}}{V_{t}\log^{2}\brac{V_{t}/G^{2}}}\\
  &\le
    2\eta\epsilon G^{2}\le 2\epsilon G
\end{align*}
Returning now to the full regret bound and recalling $\hat\cmp_{t}=\cmp_{t}$
whenever $\norm{g_{t}}\ge \cT$ and $\hat\cmp_{t}=\hat\cmp_{\tpp}$ otherwise, we have
\begin{align*}
  R_{T}(\vec{\cmp})
  &\le
    R_{T}(\vec{\hat\cmp})+\sumtT\norm{g_{t}}\norm{\hat\cmp_{t}-\cmp_{t}}\\
  &\le
    2\epsilon G+\frac{2M\Log{M/\alpha_{T+1}+1}+2P_{T}\Log{\frac{4M}{\alpha_{\tpp}\eta^{2}\cT^{2}}}}{\eta}\\
    &\qquad
      +2\eta\sumtT\norm{g_{t}}^{2}\norm{\cmp_{t}}+3\sumtT\norm{g_{t}}\norm{\hat\cmp_{t}-\cmp_{t}}\\
  &\le
    2\epsilon G+\frac{2M\Log{M/\alpha_{T+1}+1}+2P_{T}\Log{\frac{4M}{\alpha_{\tpp}\eta^{2}\cT^{2}}}}{\eta}\\
    &\qquad
      +2\eta\sumtT\norm{g_{t}}^{2}\norm{\cmp_{t}}+3\cT\sum_{t:\norm{g_{t}}\le\cT}\norm{\hat\cmp_{\tpp}-\cmp_{t}}\\
  &\overset{(a)}{\le}
    2\epsilon G+\frac{2M\Log{M/\alpha_{T+1}+1}+2P_{T}\Log{\frac{4M}{\alpha_{\tpp}\eta^{2}\cT^{2}}}}{\eta}\\
    &\qquad
    +2\eta\sumtT\norm{g_{t}}^{2}\norm{\cmp_{t}}+3\cT T P_{T}.
\end{align*}
where $(a)$ uses the fact that $\hat \cmp_{t+1} = \cmp_{t'}$ for \emph{some} $t'\ge t$, so that $\|\hat\cmp_{t+1}-\cmp_t\|\le \sum_{s=1}^{t'-1}\|\cmp_{s+1}-\cmp_s\|\le P_T$.
Since this bound holds for an arbitrary $\cT>0$ we are free to choose a $\cT$ which
tightens the upperbound, such as $\cT=\frac{4}{3\eta T}$:
\begin{align*}
  R_{T}(\vec{\cmp})
  &\le
    \inf_{\cT>0} 2\epsilon G+\frac{2M\Log{M/\alpha_{T+1}+1}+2P_{T}\Log{\frac{4M}{\alpha_{\tpp}\eta^{2}\cT^{2}}}}{\eta}\\
  &\qquad
      +2\eta\sumtT\norm{g_{t}}^{2}\norm{\cmp_{t}}+\cT 3 T P_{T}\\
    &\le
    \frac{2M\Log{M/\alpha_{T+1}+1}+2P_{T}\brac{\Log{\frac{9MT^{2}}{4\alpha_{\tpp}}}+2}}{\eta}\\
  &\qquad
      +2\epsilon G+2\eta\sumtT\norm{g_{t}}^{2}\norm{\cmp_{t}}\\
    &\le
    2\epsilon G+\frac{4\brac{M+P_{T}}\Set{\Log{\frac{9MT^{2}}{4\alpha_{T+1}}+1}\maxOp 1}}{\eta}
      +2\eta\sumtT\norm{g_{t}}^{2}\norm{\cmp_{t}}.
\end{align*}

\end{proof}

\newpage
\subsection{Proof of Theorem~\ref{thm:dynamic-combined-main}}\label{app:dynamic-combined}

The full statement of the theorem is given below.

\begin{manualtheorem}{\ref{thm:dynamic-combined-main}}
  For any $\eps>0$ and $\cmp_{1},\ldots,\cmp_{T}$ in $\R^{d}$, \Cref{alg:dynamic-meta} guarantees
  \begin{align*}
    R_{T}(\vec{\cmp})
    &\le
    2\eps G + 6\sqrt{2(M+P_{T})\sbrac{\Log{\frac{9M\Lambda_{T}}{4\eps}+1}\maxOp 1}\sumtT\norm{g_{t}}^{2}\norm{\cmp_{t}}}\\
  &\qquad
    +4G(M+P_{T})\sbrac{\Log{\frac{9M\Lambda_{T}}{4\eps}+1}\maxOp 1}.
  \end{align*}
  where
  $\Lambda_{T}= T^{2}\brac{4+\frac{\norm{g}_{1:T}^{2}}{G^{2}}}\log^{2}\brac{4 + \frac{\norm{g}_{1:T}^{2}}{G^{2}}}\lceil\log_{2}(\sqrt{T})\rceil\le O\brac{T^{3}\log^{3}(T)}$ and
  $M=\max_{t}\norm{\cmp_{t}}$.
\end{manualtheorem}
\begin{proof}
Let $\cA_{\eta}$ denote an instance of the algorithm in
\Cref{prop:dynamic-fixed-eta}, $\wt^{\eta}$ denote its iterates, and let
$R_{T}^{\cA_{\eta}}(\vec{\cmp})$ denote the dynamic regret of
$\cA_{\eta}$.
From \Cref{prop:dynamic-fixed-eta}, we have that
for any $\eta\le\frac{1}{G}$,
\begin{align*}
  R_{T}^{\cA_{\eta}}(\vec{\cmp})
  &\le
    2\epsilon G+\frac{4\brac{M+P_{T}}\sbrac{\Log{\frac{9MT^{2}}{4\alpha_{T+1}}+1}\maxOp 1}}{\eta}
      +2\eta\sumtT\norm{g_{t}}^{2}\norm{\cmp_{t}},
\end{align*}
where $\alpha_{T+1}=\frac{\epsilon G^{2}}{V_{T+1}\log^{2}\brac{V_{T+1}/G^{2}}}$
and $V_{T+1}=4G^{2}+\norm{g}_{1:T}^{2}$, $M=\max_{t\le T}\norm{\cmp_{t}}$,
$P_{T}=\sum_{t=2}^{T}\norm{\cmp_{t}-\cmp_{\tmm}}$, and $\epsilon>0$.
The stepsize which minimizes the right-hand side of the inequality
is
\begin{align*}
  \eta^{*}=\Min{\sqrt{\frac{2(M+P_{T})\sbrac{\Log{\frac{9MT^{2}}{4\alpha_{T+1}}+1}\maxOp 1}}{\sumtT\norm{g_{t}}^{2}\norm{\cmp_{t}}}},\frac{1}{G}},
\end{align*}
for which we have
\begin{align*}
  R_{T}^{\cA_{\eta^{*}}}(\vec{\cmp})
  &\le
    2\epsilon G + 4\sqrt{2(M+P_{T})\sbrac{\Log{\frac{9MT^{2}}{4\alpha_{T+1}}+1}\maxOp 1}\sumtT\norm{g_{t}}^{2}\norm{\cmp_{t}}}\\
  &\qquad
    +2G(M+P_{T})\sbrac{\Log{\frac{9MT^{2}}{4\alpha_{T+1}}+1}\maxOp 1}.
\end{align*}
In what follows, we will match this bound up to constant factors using the
iterate adding approach proposed by \cite{cutkosky2019combining}.

Suppose that we have a collection of step-sizes
$\cS=\Set{\eta\in\R: 0<\eta\le\frac{1}{G}}$
and suppose that on each round we play $\wt=\sum_{\eta\in\cS}\wt^{\eta}$
where $\wt^{\eta}$ is the output of $\cA_{\eta}$. Then for any
$\widetilde{\eta}\in\cS$ we can write
\begin{align}
  R_{T}(\vec{\cmp})
  &=
    \sumtT\inner{g_{t},\wt-\cmp_{t}}
    =
    \sumtT\inner{g_{t},\sum_{\eta\in\cS}\wt^{\eta}-\cmp_{t}}\nonumber\\
  &=
    \sumtT\inner{g_{t},\wt^{\widetilde{\eta}}-\cmp_{t}}+\sum_{\eta\ne\widetilde{\eta}\in\cS}\sumtT\inner{g_{t},\wt^{\eta}-\zeros}\nonumber\\
  &=
    R_{T}^{\cA_{\widetilde{\eta}}}(\vec{\cmp})
    +
    \sum_{\eta\ne\widetilde{\eta}\in\cS}R_{T}^{\cA_{\eta}}(\zeros)\nonumber\\
  &\le
    R_{T}^{\cA_{\widetilde{\eta}}}(\vec{\cmp})
    +
    2\epsilon G (\abs{\cS}-1).\label{eq:dynamic-combined-1}
\end{align}
Notice that since this holds for any $\widetilde{\eta}\in\cS$, it holds for
the one with the lowest dynamic regret, hence
\begin{align*}
  R_{T}(\vec{\cmp})
  &\le
    2\epsilon G(\abs{\cS}-1)+\min_{\eta\in\cS}R_{T}^{\cA_{\eta}}(\vec{\cmp}).
\end{align*}
Thus, we need only ensure that there is \textit{some} $\eta\in\cS$ which is
close to the optimal $\eta^{*}$. It is easy to see that
\begin{align*}
  \eta^{*}=\Min{\sqrt{\frac{2(M+P_{T})\sbrac{\Log{\frac{9MT^{2}}{4\alpha_{T+1}}+1}\maxOp 1}}{\sumtT\norm{g_{t}}^{2}\norm{\cmp_{t}}}},\frac{1}{G}}
  \implies
  \frac{2}{G\sqrt{T}}\le \eta^{*}\le\frac{1}{G},
\end{align*}
so if we let
$\cS=\Set{\frac{2^{k}}{G\sqrt{T}}\minOp\frac{1}{G} : 1\le k\le\lceil \log_{2}\brac{\sqrt{T}}\rceil}$,
we'll have
\begin{align*}
\eta_{\min{}}=\frac{2}{G\sqrt{T}}\le\eta^{*}\le\frac{1}{G}=\eta_{\max{}},
\end{align*}
where $\eta_{\min{}}$ and $\eta_{\max{}}$ are the smallest and largest
step-sizes
in $\cS$ respectively.
Hence, there
must be an $\eta_{k}\in\cS$ such that
$\eta_{k}\le\eta^{*}\le \eta_{k+1}\le 2\eta_{k}$.
Using $\widetilde{\eta}=\eta_{k}$ in \Cref{eq:dynamic-combined-1}
yields
\begin{align*}
  R_{T}(\vec\cmp)
  &\le
    2\epsilon G(\abs{\cS}-1)+R_{T}^{\cA_{\eta_{k}}}(\vec{\cmp})\\
  &\le
    2\epsilon G\abs{\cS}+\frac{4\brac{M+P_{T}}\sbrac{\Log{\frac{9MT^{2}}{4\alpha_{T+1}}+1}\maxOp 1}}{\eta_{k}}
    +2\eta_{k}\sumtT\norm{g_{t}}^{2}\norm{\cmp_{t}}\\
  &\le
    2\epsilon G\abs{\cS}+\frac{8\brac{M+P_{T}}\sbrac{\Log{\frac{9MT^{2}}{4\alpha_{T+1}}+1}\maxOp 1}}{\eta^*}
    +2\eta^*\sumtT\norm{g_{t}}^{2}\norm{\cmp_{t}}\\
  &=
    2\epsilon G\abs{\cS} + 6\sqrt{2(M+P_{T})\sbrac{\Log{\frac{9MT^{2}}{4\alpha_{T+1}}+1}\maxOp 1}\sumtT\norm{g_{t}}^{2}\norm{\cmp_{t}}}\\
  &\qquad
    +4G(M+P_{T})\sbrac{\Log{\frac{9MT^{2}}{4\alpha_{T+1}}+1}\maxOp 1}.
\end{align*}
The result then follows by choosing $\epsilon = \frac{\eps}{\lceil\log_{2}(\sqrt{T})\rceil}\le\frac{\eps}{\abs{\cS}}$.
\end{proof}

\newpage
\section{Proofs for Section~\ref{sec:implicit} (\secImplicit)}\label{app:implicit}

\begin{manualtheorem}{\ref{thm:optimistic-implicit}}
  Let $\ell_{1},\ldots,\ell_{T}$ and $\hat\ell_{1},\ldots,\hat\ell_{T}$ be $G$-Lipschitz convex functions.
  Let $\epsilon>0$, $k\ge 3$, and for all $t$ set
  $\Vhat_{t}=16 G^{2}+\sum_{s=1}^{t-1}\norm{\grad\ell_{s}(\w_{s})-\grad\hat \ell_{s}(\w_{s})}^{2}$,
  $\widehat{\alpha}_{t}=\frac{\epsilon G}{\sqrt{\Vhat_{t}}\log^{2}(\Vhat_{t}/G^{2})}$,
  and
  \begin{align*}
    \psi_{t}(\w)&=k\int_{0}^{\norm{\w}}\min_{\eta\le\frac{1}{2G}}\sbrac{\frac{\Log{x/\widehat{\alpha}_{t}+1}}{\eta}+\eta\Vhat_{t}}dx.
  \end{align*}
  Then for all $\cmp\in\R^{d}$, \Cref{alg:implicit-optimism} guarantees
  \begin{align*}
    R_{T}(\cmp)
    &\le
      4\epsilon G+2k\norm{\cmp}\Max{\sqrt{\Vhat_{t}\Log{\norm{\cmp}/\widehat{\alpha}_{T+1}+1}},2G\Log{\norm{\cmp}/\widehat{\alpha}_{T+1}+1}}
  \end{align*}
\end{manualtheorem}
\begin{proof}
The proof follows similar steps to
\Cref{thm:adaptive-self-stabilizing}.
Let $g_{t}\in\ell_{t}(\wt)$ and let $h_{t}\in\partial\hat\ell_{t}(\wt)$ be the
subgradient of $\hat\ell_{t}(\wt)$ for which the first-order optimality condition $h_{t}+\grad\psi_{t}(\wt)-\grad\psi_{t}(\xt)=\zeros$
holds.
Then
\begin{align*}
  \sumtT\inner{g_{t},\wt-\cmp}
  &=
    \sumtT\inner{g_{t},\xtpp-\cmp}+\inner{g_{t},\wt-\xtpp}\\
  &=
    \sumtT\inner{g_{t},\xtpp-\cmp}+\inner{h_{t},\wt-\xtpp}+\inner{g_{t}-h_{t},\wt-\xtpp}.
\end{align*}
Following the same steps as \Cref{lemma:centered-md} we have
\begin{align*}
  \sumtT\inner{g_{t},\xtpp-\cmp}
  &\le
  D_{\psi_{T+1}}(\cmp|\x_{1})-D_{\psi_{T+1}}(\cmp|\x_{T+1})+\sumtT-D_{\psi_{t}}(\xtpp|\xt)-\phi_{t}(\xtpp)\\
  &\le
    \psi_{T+1}(\cmp)+\sumtT-D_{\psi_{t}}(\xtpp|\xt)-\phi_{t}(\wtpp)
  ,
\end{align*}
where the last line observes
$\argmin_{\x\in\R^{d}}\psi_{T+1}(\x)=\psi_{T+1}(\x_{1})=0$, so
$D_{\psi_{T+1}}(\cmp|\x_{1})=\psi_{T+1}(\cmp)$
and $-D_{\psi_{T+1}}(\cmp|\x_{T+1})\le 0$.
Similarly, from the first-order optimality condition for $\wt$ we have
\begin{align*}
  \sumtT\inner{h_{t},\wt-\xtpp}
  &=\sumtT\inner{\grad\psi_{t}(\wt)-\grad\psi_{t}(\xt),\wt-\xtpp}\\
  &=\sumtT D_{\psi_{t}}(\xtpp|\xt)-D_{\psi_{t}}(\xtpp|\wt)\underbrace{-D_{\psi_{t}}(\wt|\xt)}_{\le 0}\\
    &\le \sumtT D_{\psi_{t}}(\xtpp|\xt)-D_{\psi_{t}}(\xtpp|\wt)
\end{align*}
where the second line applies the three-point relation for Bregman divergences:
\begin{align*}
  \inner{\grad f(y)-\grad f(x),x-z}=D_{f}(z|y)-D_{f}(z|x)-D_{f}(x|y).
\end{align*}
Combining these two observations yields
\begin{align*}
  R_{T}(\cmp)
  &\le
    \psi_{T+1}(\cmp)
    +\sumtT\underbrace{\inner{g_{t}-h_{t},\wt-\xtpp}-D_{\psi_{t}}(\xtpp|\wt)-\phi_{t}(\xtpp)}_{=:\delta_{t}}
\end{align*}
To bound $\delta_{t}$, define
$\widehat{g}_{t}=\grad\ell_{t}(\wt)-\grad\hat\ell_{t}(\wt)$, $\widehat{G} = 2G$,
$\Vhat_{t}=4\widehat{G}^{2}+\sum_{s=1}^{t-1}\norm{\widehat{g}_{s}}^{2}$,
$\widehat{\alpha}_{t}=\frac{\epsilon G}{\sqrt{\Vhat_{t}}\log^{2}(\Vhat_{t}/G^{2})}$,
and observe that
$\psi_{t}(\w)=k\int_{0}^{\norm{\w}}\min_{\eta\le 1/\widehat{G}}\sbrac{\frac{\Log{x/\widehat{\alpha}_{t}+1}}{\eta}+\eta\Vhat_{t}}dx$
is equivalent to the regularizer from \Cref{thm:adaptive-self-stabilizing}.
Hence, borrowing the arguments of
\Cref{thm:adaptive-self-stabilizing}, we
can bound $\sumtT\delta_{t}\le 4\epsilon G$.
Returning to our regret bound, we have
\begin{align*}
  R_{T}(\cmp)
  &\le
    \psi_{T+1}(\cmp)+4\epsilon G
  \overset{(a)}{\le}
    4\epsilon G+\norm{\cmp}\Psi_{T+1}'(\norm{\cmp})\\
  &\overset{(b)}{\le}
    4\epsilon G+
   2k\norm{\cmp}\Max{\sqrt{\Vhat_{t}\Log{\norm{\cmp}/\widehat{\alpha}_{T+1}+1}},2G\Log{\norm{\cmp}/\widehat{\alpha}_{T+1}+1}} \\
\end{align*}
where $(a)$ defines
\begin{align*}
  \Psi_{T+1}'(x)&=k\min_{\eta\le 1/2G}\sbrac{\frac{\Log{x/\widehat{\alpha}_{T+1}+1}}{\eta}+\eta\Vhat_{T+1}}\\
  &=\begin{cases}2k\sqrt{\Vhat_{T+1}\Log{x/\widehat{\alpha}_{T+1}+1}}&\text{if
    }2G\sqrt{\Log{x/\widehat{\alpha}_{T+1}+1}}\le\sqrt{\Vhat_{T+1}}\\2kG\Log{x/\widehat{\alpha}_{T+1}+1}+\frac{k\Vhat_{T+1}}{2G}&\text{otherwise}\end{cases}
\end{align*}
and observes that
$\psi_{T+1}(\cmp)=\int_{0}^{\norm{\cmp}}\Psi_{t}'(x)dx\le\norm{\cmp}\Psi_{t}'(\norm{\cmp})$
since $\Psi_{t}'$ is non-decreasing in its argument,
and $(b)$ observes that
the case
$\Psi_{t}'(x)=2kG\Log{x/\widehat{\alpha}_{T+1}+1}+\frac{k\Vhat_{T+1}}{2G}$,
coincides with
$\Vhat_{T+1}/2G\le \sqrt{\Vhat_{T+1}\Log{x/\widehat{\alpha}_{T+1}+1}}\le 2G\Log{x/\widehat{\alpha}_{T+1}+1}$,
so
\begin{align*}
  \Psi_{T+1}'(x)&\le\begin{cases}2k\sqrt{\Vhat_{T+1}\Log{x/\widehat{\alpha}_{T+1}+1}}&\text{if
    }2G\sqrt{\Log{x/\widehat{\alpha}_{T+1}+1}}\le\sqrt{\Vhat_{T+1}}\\4kG\Log{x/\widehat{\alpha}_{T+1}+1}&\text{otherwise}\end{cases}\\
            &=
              2k\Max{\sqrt{\Vhat_{T+1}\Log{x/\widehat{\alpha}_{T+1}+1}}, 2G\Log{x/\widehat{\alpha}_{T+1}+1}}
\end{align*}
\end{proof}

\newpage
\section{A Lipschitz Adaptive, Scale-free Algorithm for Unbounded Domains}\label{app:scale-invariant-alg}

The full pseudocode for our
Scale-free, Lipschitz adaptive algorithm for unbounded domains is given in
\Cref{alg:scale-invariant-full}. The update equation is derived in a similar
manner to the algorithm in \Cref{sec:pf}.

The implementation can be understood as the Leashed meta-algorithm of
\citet{cutkosky2019artificial} with an instance of the algorithm specified in
\Cref{thm:scale-invariant} as the base algorithm.
The corresponding regret guarantee is immediate using \Cref{thm:scale-invariant}
along with the with the aforementioned reductions \citep[Theorem
3]{cutkosky2019artificial}.
\begin{restatable}{mycorollary}{ScaleFreeFullAlg}\label{cor:scale-free-full-alg}
  For any $\cmp\in\R^{d}$, \Cref{alg:scale-invariant-full} guarantees
  \begin{align*}
    R_{T}(\cmp)\le
    \Ohat\Bigg(\epsilon G_T&+
      \norm{\cmp}\sbrac{\sqrt{V_{T+1}\Log{\frac{\norm{\cmp}\sqrt{B_{T+1}}}{\epsilon}+1}}\maxOp G_{T}\Log{\frac{\norm{\cmp}\sqrt{B_{T+1}}}{\epsilon}+1}}\\
      &+G_T\norm{\cmp}^{3}+G_{T}\norm{\cmp}+G_T\sqrt{\sumtT\frac{\norm{g_{t}}}{G_{t}}}\Bigg),
  \end{align*}
  where $G_{T}=\max_{\tau\le T}\norm{g_{\tau}}$ and
  $B_{T+1}=4\sum_{t=1}^{T+1}\brac{4 +\sum_{s=1}^{t-1}\frac{\norm{g_{s}}^{2}}{h_{s}^{2}}}$
  and $\Ohat(\cdot)$ hides constant and $\log(\log)$ factors.
\end{restatable}

\newcommand{\Btilde}{\widetilde{B}}
\newcommand{\btilde}{\widetilde{b}}
\begin{algorithm}
  \SetAlgoLined
  \KwInitialize{$w_{1}=\zeros$,
    $h_{1}=0$, $G_{0}=0$, $\btilde_{1}=4$, $\Btilde_{1}=4\btilde_{1}$, $\widetilde{\theta}_{1}=\zeros$}\\
  \For{$t=1:T$}{
    Define $D_{t}=\sqrt{\sum_{s=1}^{\tmm}\frac{\norm{g_{s}}}{G_{s}}}$ and $W_{t}=\Set{\w\in\R^{d}:\norm{\w}\le D_{t}}$\\
    Play $\hat\wt = \proj_{W_{t}}(\wt)=\wt\Min{1,\frac{D_{t}}{\norm{\wt}}}$\\
    Receive subgradient $g_{t}$\\
    \BlankLine
    \BlankLine
    Set $\gbar_{t}=g_{t}\Min{1,\frac{h_{t}}{\norm{g_{t}}}}$,
    $G_{t}=\Max{\norm{g_{t}}, G_{\tmm}}$, and $h_{\tpp}=G_{t}$\\
    Set
    $\widetilde{\ell}_{t}(\w)=\half\inner{\gbar_{t},\w}+\half\norm{\gbar_{t}}\Max{0, \norm{\wt}-D_{t}}$
    and compute $\gtilde_{t}\in\partial\widetilde{\ell}_{t}(\wt)$\\
    \BlankLine
    \BlankLine
    Set $\widetilde{\theta}_{\tpp} = \widetilde{\theta}_{t}-\gtilde_{t}$,
     $\Vtilde_{\tpp}=4h_{\tpp}^{2}+\norm{\gtilde}_{1:t}^{^{2}}$,
    $\btilde_{\tpp}=\btilde_{t}+\frac{\norm{\gtilde_{t}}^{2}}{h_{t}^{2}}$,
    $\Btilde_{\tpp}= \Btilde_{t}+4\btilde_{t}$, and\\
    \ \ \ \ \ \  $\widetilde{\alpha}_{\tpp}=\frac{\epsilon}{\sqrt{\Btilde_{\tpp}}\log^{2}(\Btilde_{\tpp})}$\\
    Define
    \(
    f_{\tpp}(\theta)=\begin{cases}
      \frac{\norm{\theta}^{2}}{36 \Vtilde_{\tpp}}&\text{if }\norm{\theta}\le\frac{6\Vtilde_{\tpp}}{h_{\tpp}}\\
      \frac{\norm{\theta}}{3h_{\tpp}}-\frac{\Vtilde_{\tpp}}{h_{\tpp}^{2}}&\text{otherwise}
      \end{cases}
    \)\\
    Update
    \(
      \wtpp =
     \frac{\widetilde{\alpha}_{\tpp}\widetilde{\theta}_{\tpp}}{\norm{\widetilde{\theta}_{\tpp}}}\sbrac{\Exp{f_{\tpp}\brac{\widetilde{\theta}_{\tpp}}}-1}
    \)
  }
  \caption{Unbounded, Scale-Free, Lipschitz Adaptivity}
  \label{alg:scale-invariant-full}
\end{algorithm}

\section{Proofs for Section~\ref{sec:scale-free} (\secScaleFree)}\label{app:scale-free}

\subsection{Proof of Theorem~\ref{thm:scale-invariant}}
The complete theorem is stated below.
\begin{manualtheorem}{\ref{thm:scale-invariant}}\label{thm:scale-invariant-full}
  Let $\ell_{1},\ldots,\ell_{T}$ be $G$-Lipschitz convex functions and
  $g_{t}\in\partial\ell_{t}(\wt)$
  for all $t$.
  Let $h_{1}\le\ldots\le h_{T}$ be a sequence of hints such that
  $h_{t}\ge \norm{g_{t}}$, and assume that $h_{t}$ is provided at the start of
  each round $t$.
  Let $\epsilon>0$, $k\ge 3$, $V_{t}=4h_{t}^{2}+\norm{g}_{1:\tmm}^{2}$,
  $B_{t}=4\sum_{s=1}^{t}\brac{4+\sum_{s'=1}^{s-1}\frac{\norm{g_{s'}}^{2}}{h_{s'}^{2}}}$,
  $\alpha_{t}=\frac{\epsilon }{\sqrt{B_{t}}\log^{2}(B_{t})}$,
  and set
  \begin{align*}
    \psi_{t}(\w)&=
    k\int_{0}^{\norm{\w}}\min_{\eta\le\frac{1}{h_{t}}}\sbrac{\frac{\Log{x/\alpha_{t}+1}}{\eta}+\eta V_{t}}dx.
  \end{align*}
  Then
  for all $\cmp\in\R^{d}$, \Cref{alg:centered-md} guarantees
  \begin{align*}
    R_{T}(\cmp)
    \le
      4\epsilon h_{T}+2k\norm{\cmp}\max\Bigg\{&\sqrt{V_{T+1}\Log{\frac{\norm{\cmp}\sqrt{B_{T+1}}\log^{2}\brac{B_{T+1}}}{\epsilon}+1}},\\
                                  &\ h_{T}\Log{\frac{\norm{\cmp}\sqrt{B_{T+1}}\log^{2}\brac{B_{T+1}}}{\epsilon}+1}\Bigg\}
  \end{align*}
\end{manualtheorem}
\begin{proof}
  The proof follows similar steps to \Cref{thm:adaptive-self-stabilizing}.
  We have via \Cref{lemma:centered-md} that
\begin{align*}
  R_{T}(\cmp)
  &\le
    \psi_{T+1}(\cmp)+\sumtT\underbrace{\inner{g_{t},\wt-\wtpp}-D_{\psi_{t}}(\wtpp|\wt)-\Delta_{t}(\wtpp)}_{=:\delta_{t}},
\end{align*}
so the main challenge is to bound the stability terms $\sumtT\delta_{t}$, which
we focus on first.

Let $\psiq_{t}(\w)=\Log{x/\alpha_{t}+1}$ and define
\begin{align*}
  \Psi_{t}(x)=k\int_{0}^{x}\min_{\eta\le1/h_{t}}\sbrac{\frac{\psiq_{t}(x)}{\eta}+\eta V_{t}}dx,
\end{align*}
so that $\psi_{t}(\w)=\Psi_{t}(\norm{\w})$, and observe that
\begin{align*}
  \Psi_{t}'(x)
  &=
    k\min_{1/h_{t}}\sbrac{\frac{\psiq_{t}(x)}{\eta}+\eta V_{t}}\\
  &=
    \begin{cases}
      2k\sqrt{V_{t}\psiq_{t}(x)}&\text{if }h_{t}\sqrt{\psiq_{t}(x)}\le\sqrt{V_{t}}\\
      kh_{t}\psiq_{t}(x)+\frac{kV_{t}}{h_{t}}&\text{otherwise}
    \end{cases}\\
  \Psi_{t}''(x)
  &=
    \begin{cases}
      \frac{k}{x+\alpha_{t}}\sqrt{\frac{\psiq_{t}(x)}{V_{t}}} &\text{if }h_{t}\sqrt{\psiq_{t}(x)}\le\sqrt{V_{t}}\\
      \frac{k h_{t}}{x+\alpha_{t}}&\text{otherwise}
    \end{cases}\\
  \Psi_{t}'''(x)
  &=
    \begin{cases}
      \frac{-k\sqrt{V_{t}}\brac{1+2\psiq_{t}(x)}}{2(x+\alpha_{t})^{2}\psiq_{t}(x)^{3/2}}&\text{if }h_{t}\sqrt{\psiq_{t}(x)}\le\sqrt{V_{t}}\\
      \frac{-k h_{t}}{(x+\alpha_{t})^{2}}&\text{otherwise.}
    \end{cases}
\end{align*}
Hence, $\Psi_{t}(x)\ge 0$, $\Psi_{t}'(x)\ge 0$, $\Psi_{t}''(x)\ge 0$, and
$\Psi_{t}'''(x)\le 0$ for all $x > 0$.
Moreover, for any $x> \alpha_{t}(e-1)\defeq x_{0}$ we have
\begin{align*}
    -\frac{\Psi_{t}'''(x)}{\Psi_{t}''(x)^{2}}
  &=
    \begin{cases}
      \frac{k\sqrt{V_{t}}(1+2\psiq_{t}(x))}{2(x+\alpha_{t})\psiq_{t}(x)^{3/2}}\frac{(x+\alpha_{t})^{2}\psiq_{t}(x)}{k^{2}V_{t}}&\text{if
      } h_{t}\sqrt{\psiq_{t}(x)}\le\sqrt{V_{t}}\\
      \frac{kh_{t}}{(x+\alpha_{t})^{2}}\frac{(x+\alpha_{t})^{2}}{k^{2}h_{t}^{2}}&\text{otherwise}
    \end{cases}\\
  &\le
    \begin{cases}
      \frac{1}{2k\sqrt{V_{t}}}\brac{\frac{1}{\sqrt{\psiq_{t}(x)}}+2\sqrt{\psiq_{t}(x)}}&\text{if
      } h_{t}\sqrt{\psiq_{t}(x)}\le\sqrt{V_{t}}\\
      \frac{1}{kh_{t}}&\text{otherwise,}
    \end{cases}
                       \intertext{and since $x>x_{0}$, we have $\sqrt{F_{t}(x)}> 1$
                       and $\frac{1}{\sqrt{F_{t}(x)}}\le\sqrt{F_{t}(x)}$, hence}
  &\le
    \begin{cases}
      \frac{3}{2k}\sqrt{\frac{\psiq_{t}(x)}{V_{t}}}&\text{if
      } h_{t}\sqrt{\psiq_{t}(x)}\le\sqrt{V_{t}}\\
      \frac{1}{kh_{t}}&\text{otherwise}
    \end{cases}\\
  &\le
    \frac{1}{2}\Min{\sqrt{\frac{\psiq_{t}(x)}{V_{t}}},\frac{1}{h_{t}}}\\
  &= \half \eta_{t}'(x),
\end{align*}
for $k\ge 3$ and
$\eta_{t}(x)=\int_{0}^{\norm{\w}}\Min{\sqrt{\frac{\psiq_{t}(x)}{V_{t}}},\frac{1}{h_{t}}}dx$. Notice that $\eta_t$
is convex and $1/h_{t}$ Lipschitz with $h_{t}\ge\norm{g_{t}}$. Hence,
$\Psi_{t}$ satisfies the conditions of \Cref{lemma:md-stability} with
$x_{0}=\alpha_{t}(e-1)$, so
\begin{align}
  \widehat{\delta}_{t}&=\inner{g_{t},\wt-\wtpp}-D_{\psi_{t}}(\wtpp|\wt)-\eta_{t}(\norm{\wtpp})\norm{g_{t}}^{2}
                  \le\frac{2\norm{g_{t}}^{2}}{\Psi_{t}''(x_{0})}.\label{eq:scale-free-eq-1}
\end{align}
Next, we want to show that $\delta_{t}\le\widehat{\delta}_{t}$, which will follow
if we can show that $\Delta_{t}(\w)\ge\eta_{t}(\norm{\w})\norm{g_{t}}^{2}$ for any $\w$.
Observe that for any $x>0$ we have
\begin{align*}
  \Psi_{\tpp}'(x)-\Psi_{t}'(x)
  &=
    k\min_{\eta\le \frac{1}{h_{\tpp}}}\sbrac{\frac{\psiq_{\tpp}(x)}{\eta}+\eta V_{\tpp}}-k\min_{\eta\le\frac{1}{h_{t}}}\sbrac{\frac{\psiq_{t}(x)}{\eta}+\eta V_{t}}\\
  &\ge
    k\min_{\eta\le \frac{1}{h_{t}}}\sbrac{\frac{\psiq_{\tpp}(x)}{\eta}+\eta V_{\tpp}}-k\min_{\eta\le\frac{1}{h_{t}}}\sbrac{\frac{\psiq_{t}(x)}{\eta}+\eta V_{t}}.
    \intertext{Now observe that for any $\eta\le1/h_{t}$, it holds that
    $\frac{\psiq_\tpp(x)}{\eta}+\eta V_{\tpp}=\frac{\psiq_{\tpp}(x)}{\eta}+\eta V_{t}+\eta\norm{g_{t}}^{2}\ge\min_{\eta^{*}\le1/h_{t}}\sbrac{\frac{\psiq_{\tpp}(x)}{\eta^{*}}+\eta^{*} V_{t}}+\eta\norm{g_{t}}^{2}$,
    which yields}
    &\ge
    k\norm{g_{t}}^{2}\Min{\sqrt{\frac{\psiq_{\tpp}(x)}{V_{\tpp}}},\frac{1}{h_{t}}}+k\min_{\eta\le \frac{1}{h_{t}}}\sbrac{\frac{\psiq_{\tpp}(x)}{\eta}+\eta V_{t}}-k\min_{\eta\le\frac{1}{h_{t}}}\sbrac{\frac{\psiq_{t}(x)}{\eta}+\eta V_{t}}\\
  &\overset{(a)}{\ge}
    k\norm{g_{t}}^{2}\Min{\sqrt{\frac{\psiq_{t}(x)}{V_{\tpp}}},\frac{1}{h_{t}}}+k\min_{\eta\le \frac{1}{h_{t}}}\sbrac{\frac{\psiq_{t}(x)}{\eta}+\eta V_{t}}-k\min_{\eta\le\frac{1}{h_{t}}}\sbrac{\frac{\psiq_{t}(x)}{\eta}+\eta V_{t}}\\
  &=
    k\norm{g_{t}}^{2}\Min{\sqrt{\frac{\psiq_{t}(x)}{V_{\tpp}}},\frac{1}{h_{t}}}
    \overset{(b)}{\ge}\frac{k}{\sqrt{2}}\norm{g_{t}}^{2}\Min{\sqrt{\frac{\psiq_{t}(x)}{V_{t}}},\frac{1}{h_{t}}}\\
    &\ge\norm{g_{t}}^{2}\Min{\sqrt{\frac{\psiq_{t}(x)}{V_{t}}},\frac{1}{h_{t}}}=\eta_{t}'(x)\norm{g_{t}}^{2},
\end{align*}
where $(a)$ uses that $\alpha_{\tpp}\le\alpha_{t}$ so
$F_{\tpp}(x)=\Log{x/\alpha_{\tpp}+1}\ge\Log{x/\alpha_{t}+1}=F_{t}(x)$, and $(b)$ uses
$\frac{1}{V_{t}}=\frac{1}{V_{\tpp}}\frac{V_{\tpp}}{V_{t}}\le\frac{2}{V_{\tpp}}$. From this we immediately have
\begin{align*}
  \Delta_{t}(\w)=\int_{0}^{\norm{\w}}\Psi_{\tpp}'(x)-\Psi_{t}'(x)dx \ge\norm{g_{t}}^{2}\int_{0}^{\norm{\w}}\eta_{t}'(x)dx
  =\eta_{t}(\norm{\w})\norm{g_{t}}^{2},
\end{align*}
so combining this with \Cref{eq:scale-free-eq-1}, we have
\begin{align*}
  \delta_{t}
  &=
    \inner{g_{t},\wt-\wtpp}-D_{\psi_{t}}(\wtpp|\wt)-\Delta_{t}(\wtpp)\\
  &\le
    \inner{g_{t},\wt-\wtpp}-D_{\psi_{t}}(\wtpp|\wt)-\eta_{t}(\norm{\wtpp})\norm{g_{t}}^{2}\\
  &=\widehat{\delta}_{t}
  \le
    \frac{2\norm{g_{t}}^{2}}{\Psi_{t}''(x_{0})}=\frac{2\alpha_{t}e\norm{g_{t}}^{2}}{k\sqrt{V_{t}}}\le\frac{2\alpha_{t}\norm{g_{t}}^{2}}{\sqrt{V_{t}}}
\end{align*}
for $k\ge 3$.
Returning to our regret bound, we have
\begin{align}
  R_{T}(\cmp)
  &\le
    \psi_{T+1}(\cmp)+\sumtT\delta_{t}
    \le
    \psi_{T+1}(\cmp)+2\sumtT\frac{\alpha_{t}\norm{g_{t}}^{2}}{\sqrt{V_{t}}}\nonumber\\
  &\overset{(a)}{\le}
    \norm{\cmp}\Psi_{T+1}'(\norm{\cmp})+2\sumtT\frac{\alpha_{t}\norm{g_{t}}^{2}}{\sqrt{V_{t}}}\nonumber\\
  &\overset{(b)}{\le}
    2k\norm{\cmp}\Max{\sqrt{V_{T+1}\Log{\norm{\cmp}/\alpha_{T+1}+1}},h_{T+1}\Log{\norm{\cmp}/\alpha_{T+1}+1}}\nonumber\\
  &\qquad
    +2\sumtT\frac{\alpha_{t}\norm{g_{t}}^{2}}{\sqrt{V_{t}}}\label{eq:scale-free-eq-2}
\end{align}
where $(a)$ observes that
$\Psi_{t}'(x)$ is increasing in $x$, so
\begin{align*}
  \psi_{T+1}(\cmp)=\int_{0}^{\norm{\cmp}}\Psi_{T+1}'(x)dx\le\Psi_{t}'(\norm{\cmp})\int_{0}^{\norm{\cmp}}dx
  =\norm{\cmp}\Psi_{t}'(\norm{\cmp}),
\end{align*}
and $(b)$ observes that the case
$\Psi_{t}'(x)=kh_{t}\psiq_{t}(x)+\frac{kV_{t}}{h_{t}}$ coincides with
$\frac{V_{t}}{h_{t}}\le h_{t}\psiq_{t}(x)$, so
\begin{align*}
  \Psi_{T+1}'(\norm{\cmp})
  &=
    \begin{cases}
      2k\sqrt{V_{T+1}\psiq_{T+1}(\norm{\cmp})}&\text{if }h_{T+1}\sqrt{\psiq_{T+1}(\norm{\cmp})}\le\sqrt{V_{T+1}}\\
      kh_{T+1}\psiq_{T+1}(\norm{\cmp})+\frac{kV_{T+1}}{h_{T+1}}&\text{otherwise}
    \end{cases}\\
  &\le
    \begin{cases}
      2k\sqrt{V_{T+1}\psiq_{T+1}(\norm{\cmp})}&\text{if }h_{T+1}\sqrt{\psiq_{T+1}(\norm{\cmp})}\le\sqrt{V_{T+1}}\\
      2kh_{T+1}\psiq_{T+1}(\norm{\cmp})&\text{otherwise}
    \end{cases}\\
  &=
    2k\Max{\sqrt{V_{T+1}\psiq_{T+1}(\norm{\cmp})},h_{T+1}\psiq_{T+1}(\norm{\cmp})}.
\end{align*}
Note that the regret does not depend on $g_{T+1}$, so without loss of generality
we can assume $g_{T+1}=g_{T}$ and hence $h_{T+1}=h_{T}$.
Finally, \Cref{lemma:scale-invariant-alphas}
bounds
$2\sumtT\frac{\alpha_{t}\norm{g_{t}}^{2}}{\sqrt{V_{t}}}\le 4\epsilon h_{T}$, so
plugging this into \Cref{eq:scale-free-eq-2} yields the stated result.
Notice that \Cref{lemma:scale-invariant-alphas} is responsible for removing the
``range-ratio'' problem addressed by \cite{mhammedi2020lipschitz} via a doubling-like restart scheme.
\end{proof}

\begin{restatable}{mylemma}{ScaleInvariantAlphas}\label{lemma:scale-invariant-alphas}
  Let $c\ge 4$,
  $V_{t}=ch_{t}^{2}+\norm{g}_{1:\tmm}^{2}$,
  $B_{t}=c\sum_{s=1}^{t}\brac{4+\sum_{s'=1}^{s-1}\frac{\norm{g_{s'}}^{2}}{h_{s'}^{2}}}$
  and set
  $\alpha_{t}=\frac{\epsilon}{\sqrt{B_{t}}\log^{2}(B_{t})}$. Then
  \begin{align*}
    \sumtT\frac{\alpha_{t}\norm{g_{t}}^{2}}{\sqrt{V_{t}}}
    \le2\epsilon h_{T}.
  \end{align*}
\end{restatable}
\begin{proof}
  Define $\tau_{1}=1$ and
  $\tau_{t}=\max\Set{t' : t'\le t\text{ and }\sum_{s=1}^{t'-1}\frac{\norm{g_{s}}^{2}}{h_{s}^{2}}+4<\frac{h_{t'}^{2}}{h_{\tau_{t'-1}}^{2}}}$
  for $t>1$. Then, we partition $[1,T]$ into the disjoint intervals $[1,T]=\cI_{1}\cup\ldots\cup\cI_{N}$
  over which $\tau_{t}$ is fixed.
  Denote
  $\cI_{j}=[\tautilde_{j}, \tautilde_{j+1}-1]$ where $\tautilde_{1}=1$, $\tautilde_{N+1}=T+1$,
  and $\tautilde_{j}=\min\Set{t>\tautilde_{j-1}: \tau_{t}>\tau_{t-1}}$ for $j\in[2,N]$.
  Observe that by definition, $\tau_{t}=\tautilde_{j}$ for all $t\in\cI_{j}$. Further, for all $j$ and $t\in \cI_j$, we have either $t=\tautilde_{j}$ or $\tau_{t-1} = \tautilde_{j}<t$, so that:
  \begin{align*}
      \frac{h^2_t}{h_{\tautilde_j}^2}\le 4+\sum_{s=1}^{t-1} \frac{\|g_s\|^2}{h_s^2}
  \end{align*}

  Now, we show that $V_{\tpp}/h_{\tau_{\tpp}^{2}}\le B_{\tpp}$.  Notice that
  if $t$ is the last round of an interval $\cI_{k}$, then $t+1$
  would be the start of the next epoch so $h_{\tau_{t+1}}=h_{t+1}$ and
  $V_{t+1}/h_{\tau_{t+1}}^{2}=V_{t+1}/h_{t+1}^{2}\le B_{t+1}$ (since $c\ge 1$).
  Otherwise, $t+1$ occurs before the end of interval $\cI_{k}$ so
\begin{align*}
  \frac{V_{t+1}}{h_{\tau_{t+1}}^{2}}
  &=\frac{ch_{t+1}^{2}+\norm{g}_{1:t}^{2}}{h_{\tautilde_{k}}^{2}}
  \le
    c\frac{h_{t+1}^{2}}{h_{\tautilde_{k}}^{2}}+\sum_{j=1}^{k}\sum_{\substack{s\in\cI_{j}\\ s\le t}}\frac{\norm{g_{s}}^{2}}{h_{\tautilde_{j}}^{2}}\\
  &\le
    c\frac{h_{t+1}^{2}}{h_{\tautilde_{k}}^{2}}+\sum_{j=1}^{k}\sum_{\substack{s\in\cI_{j}\\ s\le t}}\frac{h_{s}^{2}}{h_{\tautilde_{j}}^{2}}
    \intertext{Now, apply the definition of $\tautilde_j$ to get:}
  &\le
    c\brac{4+\sum_{s=1}^{t}\frac{\norm{g_{s}}^{2}}{h_{s}^{2}}}+\sum_{j=1}^{k}\sum_{\substack{s\in\cI_{j}\\ s\le t}}\brac{4+\sum_{s'=1}^{s-1}\frac{\norm{g_{s'}}^{2}}{h_{s'}^{2}}}\\
  &\le
    c\brac{4+\sum_{s=1}^{t}\frac{\norm{g_{s}}^{2}}{h_{s}^{2}}}+\sum_{s=1}^{t}\brac{4+\sum_{s'=1}^{s-1}\frac{\norm{g_{s'}}^{2}}{h_{s'}^{2}}}\\
  &\le
    c\sum_{s=1}^{t+1}\brac{4+\sum_{s'=1}^{s-1}\frac{\norm{g_{s'}}^{2}}{h_{s'}^{2}}} = B_{\tpp}.
\end{align*}
Now, using this we have that
\(
  \alpha_{t}=\frac{\eps}{\sqrt{B_{t}}\log^{2}(B_{t})}\le\frac{\eps h_{\tau_{t}}}{\sqrt{V_{t}}\log^{2}\brac{V_{t}/h_{\tau_{t}^{2}}}}
\)
and thus
\begin{align*}
  \sumtT\frac{\alpha_{t}\norm{g_{t}}^{2}}{\sqrt{V_{t}}}
  &=
    \sum_{j=1}^{N}\sum_{t\in\cI_{j}}\frac{\alpha_{t}\norm{g_{t}}^{2}}{\sqrt{V_{t}}}
  =
    \epsilon\sum_{j=1}^{N}\sum_{t\in\cI_{j}}\frac{\norm{g_{t}}^{2}}{\sqrt{V_{t}}\sqrt{B_{t}}\log^{2}(B_{t})}
  \le
    \epsilon\sum_{j=1}^{N}\sum_{t\in\cI_{j}}h_{\tau_{t}}\frac{\norm{g_{t}}^{2}}{V_{t}\log^{2}\brac{V_{t}/h_{\tau_{t}}^{2}}}\\
  &=
    \epsilon\sum_{j=1}^{N}h_{\tautilde_{j}}\sum_{t\in\cI_{j}}\frac{\norm{g_{t}}^{2}}{\brac{ch_{t}^{2}+\norm{g}_{1:\tmm}^{2}}\log^{2}\brac{\frac{ch_{t}^{2}+\norm{g}_{1:\tmm}^{2}}{h_{\tau_{t}}^{2}}}}\\
  &\le
    \epsilon\sum_{j=1}^{N}h_{\tautilde_{j}}\sum_{t\in\cI_{j}}\frac{\norm{g_{t}}^{2}}{\brac{(c-1)h_{\tautilde_{j}}^{2}+\norm{g}_{1:t}^{2}}\log^{2}\brac{\frac{(c-1)h_{\tautilde_{j}}^{2}+\norm{g}_{1:t}^{2}}{h_{\tautilde_{j}}^{2}}}}\\
  &\le
    \epsilon\sum_{j=1}^{N}h_{\tautilde_{j}}\int_{(c-1)h_{\tautilde_{j}}^{2}}^{(c-1)h_{\tautilde_{j}}^{2}+\norm{g}_{1:t}^{2}}\frac{1}{x\log^{2}(x/h_{\tautilde_{j}}^{2})}dx\\
  &=
    \epsilon\sum_{j=1}^{N}h_{\tautilde_{j}}\frac{-1}{\Log{x/h_{\tautilde_{j}}^{2}}}\Bigg|_{x=(c-1)h_{\tautilde_{j}}^{2}}^{(c-1)h_{\tautilde_{j}}^{2}+\norm{g}_{1:t}^{2}}
  \le
    \frac{\epsilon}{\Log{c-1}}\sum_{j=1}^{N}h_{\tautilde_{j}}.
\end{align*}
Notice
that each interval begins when
$\frac{h_{\tautilde_{j}}^{2}}{h_{\tautilde_{j-1}}^{2}}>\sum_{s=1}^{t-1}\frac{\norm{g_{s}}^{2}}{h_{s}^{2}}+4>4$,
so
$h_{\tautilde_{j}} > 2h_{\tautilde_{j-1}}$ and hence
\begin{align*}
  \frac{\epsilon}{\Log{c-1}}\sum_{j=1}^{N}h_{\tautilde_{j}}\le\frac{\epsilon}{\Log{c-1}}\sum_{j=0}^{N-1}\frac{1}{2^{j}}h_{\tautilde_{N}}\le\frac{2\epsilon h_{T}}{\Log{c-1}}
  \le 2\epsilon h_{T},
\end{align*}
for $c>4$.
\end{proof}

\section{Proofs for Section~\ref{sec:opt-tradeoff} (\secOptTradeoff)}
\label{app:opt-tradeoff}

\subsection{Proof of Theorem~\ref{thm:pf-ii}}%
\label{app:pf-ii}
\PFII*
\begin{proof}
  We will prove the result with
  $\alpha_{t}=\epsilon G^{1-2\beta}/ V_{t}^{\half-\beta}$ for $\beta\in (0,\half]$
  and then conclude by choosing $\beta=\half-\rho$ for $\rho\in[0,\half)$.
  Following the same steps as \Cref{thm:adaptive-self-stabilizing},
  we have
  \begin{align*}
    R_{T}(\cmp)
    &\le
      \psi_{T+1}(\cmp)+\sumtT\delta_{t}\\
    &\le
      \psi_{T+1}(\cmp)+\sumtT\frac{2\alpha_{t}\norm{g_{t}}^{2}}{\sqrt{V_{t}}}\\
    \intertext{and substituting
    $\alpha_{t}=\epsilon G^{1-2\beta}/V_{t}^{\half-\beta}$,}
    &=
      \psi_{T+1}(\cmp)+\sumtT\frac{2\epsilon G^{1-2\beta}\norm{g_{t}}^{2}}{V_{t}^{1-\beta}}\\
    &\le
      \psi_{T+1}(\cmp)+2\epsilon G^{1-2\beta}\sumtT\frac{\norm{g_{t}}^{2}}{(\norm{g}_{1:t}^{2})^{1-\beta}}
  \end{align*}
  where we've used $V_{t}=4G^{2}+\norm{g}^{2}_{1:\tmm}\ge\norm{g}_{1:t}^{2}$. Moreover, by concavity of
  $x\mapsto x^{\beta}$ for $\beta<1$ we have
  $(\norm{g}_{1:t}^{2})^{\beta}-(\norm{g}_{1:\tmm}^{2})^{\beta}\ge \frac{\beta\norm{g_{t}}^{2}}{(\norm{g}_{1:t}^{2})^{1-\beta}}$,
  and hence
  $\sumtT \frac{\norm{g_{t}}^{2}}{(\norm{g}_{1:t}^{2})^{1-\beta}}\le \frac{1}{\beta}(\norm{g}_{1:T}^{2})^{\beta}$,
  giving an overall bound of
  \begin{align*}
    R_{T}(\cmp)
    &\le
      \psi_{T+1}(\cmp)+\frac{2\epsilon G^{1-2\beta}}{\beta}(\norm{g}_{1:T}^{2})^{\beta}\\
    &\le
      k\norm{\cmp}\sbrac{\sqrt{V_{T+1}\Log{\frac{\norm{\cmp}}{\alpha_{T+1}}+1} +1}\maxOp G\Log{\frac{\norm{\cmp}}{\alpha_{T+1}}+1}}+\frac{2\epsilon G^{1-2\beta}}{\beta}(V_{T+1})^{\beta}\\
    &\le
      O\brac{\frac{\epsilon}{\beta} G^{1-2\beta}V_{T+1}^{\beta}+\norm{\cmp}\sbrac{\sqrt{V_{T+1}\Log{\frac{\norm{\cmp}V_{T+1}^{\half-\beta}}{\epsilon G^{1-2\beta}} +1}}\maxOp G\Log{\frac{\norm{\cmp}V_{T+1}^{\half-\beta}}{\epsilon G^{1-2\beta}}+1}}}
  \end{align*}
  where the first inequality bounds $\psi_{T+1}(\cmp)$ using the same argument as \Cref{thm:adaptive-self-stabilizing}.
  Substituting $\beta=\half-\rho$ gives the stated result.
\end{proof}

\subsection{Proof of Theorem~\ref{thm:scale-free-ii}}%
\label{app:scale-free-ii}
\ScaleFreeII*
\begin{proof}
  First note that
  $\lim_{\rho\to0}B_{t}^{\rho}=\lim_{\rho\to 0}\sbrac{4\sum_{s=1}^{t}\sbrac{2^{\frac{1}{\rho}}+\sum_{s'=1}^{s-1}\frac{\norm{g_{s}}^{2}}{h_{s^{2}}}}}^{\rho}=2$,
  so for the case $\rho=0$ we let $B_{t}^{\rho}=2$ for all $t$. Then following the same
  argument as
  \Cref{prop:simple-opt-tradeoff} we get
  \begin{align*}
    R_{T}(\cmp)\le
      O\brac{\epsilon\sqrt{\norm{g}_{1:T}^{2}}+\norm{\cmp}\sbrac{\sqrt{\norm{g}_{1:T}^{2}\Log{\frac{\norm{\cmp}}{\epsilon}+1}}\maxOp h_{T}\Log{\frac{\norm{\cmp}}{\epsilon}+1}}}.
  \end{align*}

  Next, we consider the case $\rho>0$. Similar to \Cref{thm:pf-ii}, we prove the result with
  $B_{t}=4\sum_{s=1}^{t}\sbrac{2^{\frac{2}{1-2\beta}}+\sum_{s'=1}^{s-1}\frac{\norm{g_{s}}^{2}}{h_{s^{2}}}}$ and
  $\alpha_{t}=\epsilon/B_{t}^{\half-\beta}$ for
  $\beta\in(0,\half)$, and then substitute $\rho=\half-\beta$ to complete the result.
  Following the same arguments as \Cref{thm:scale-invariant}, we have
  \begin{align*}
    R_{T}(\cmp)
    &\le
      2k\norm{\cmp}\sbrac{\sqrt{V_{T+1}\Log{\norm{\cmp}/\alpha_{T+1}+1}}\maxOp h_{T}\Log{\norm{\cmp}/\alpha_{T+1}+1}}+2\sumtT\frac{\alpha_{t}\norm{g_{t}}^{2}}{\sqrt{V_{t}}}
  \end{align*}
  where $V_{t}=4h_{t}^{2}+\norm{g}_{1:\tmm}^{2}$ and $h_{t}\ge \norm{g_{t}}$ for all
  $t$.

  Next, we follow the same argument as
  \Cref{lemma:scale-invariant-alphas}.
  Define $\tau_{1}=1$ and for $t>1$ define
  $\tau_{t}=\max\Set{t' : t'\le t\text{ and }\sum_{s=1}^{t'-1}\frac{\norm{g_{s}}^{2}}{h_{s}^{2}}+2^{\frac{2}{1-2\beta}}<\frac{h_{t'}^{2}}{h_{\tau_{t'-1}}^{2}}}$.
  Then, we partition $[1,T]$ into the disjoint intervals $[1,T]=\cI_{1}\cup\ldots\cup\cI_{N}$
  over which $\tau_{t}$ is fixed.
  Denote
  $\cI_{j}=[\tautilde_{j}, \tautilde_{j+1}-1]$ where $\tautilde_{1}=1$, $\tautilde_{N+1}=T+1$,
  and $\tautilde_{j}=\min\Set{t>\tautilde_{j-1}: \tau_{t}>\tau_{t-1}}$ for $j\in[2,N]$.
  Using the same argument as \Cref{lemma:scale-invariant-alphas}, it holds that
  $B_{\tpp}=4\sum_{s=1}^{\tpp}\sbrac{2^{\frac{2}{1-2\beta}}+\sum_{s'=1}^{s-1}\frac{\norm{g_{s'}}^{2}}{h_{s'}^{2}}}\ge \frac{V_{\tpp}}{h_{\tau_{t+1}}^{2}}$, so plugging this in above we
  have
  \begin{align*}
    \sumtT \frac{\alpha_{t}\norm{g_{t}}^{2}}{\sqrt{V_{t}}}
    &=
      \epsilon\sum_{j=1}^{N}\sum_{t\in \cI_{j}}\frac{\norm{g_{t}}^{2}}{B_{t}^{\half-\beta}\sqrt{V_{t}}}\\
    &\le
      \epsilon\sum_{j=1}^{N} h_{\tilde \tau_{j}}^{1-2\beta}\sum_{t\in\cI_{j}}\frac{\norm{g_{t}}^{2}}{V_{t}^{1-\beta}}\\
    &\overset{(*)}{\le}
      \epsilon\sum_{j=1}^{N} h_{\tilde \tau_{j}}^{1-2\beta}\frac{1}{\beta}(\sum_{t\in\cI_{j}}\norm{g_{t}}^{2})^{\beta}\\
    &\le
      \frac{\epsilon}{\beta} (\norm{g}_{1:T}^{2})^{\beta}\sum_{j=1}^{N}h_{\tilde \tau_{j}}^{1-2\beta},
  \end{align*}
  where $(*)$ bounds $\sum_{t\in\cI_{j}}\frac{\norm{g_{t}}^{2}}{V_{t}^{1-\beta}}$
  using the same argument as \Cref{thm:pf-ii}.
  Then, since each interval begins when
  $h_{\tautilde_{j}}^{2}/h_{\tautilde_{j-1}}^{2}\ge \sum_{s=1}^{t-1}\frac{\norm{g_{s}}^{2}}{h_{s}^{2}}+2^{\frac{2}{1-2\beta}}\ge 2^{\frac{2}{1-2\beta}}$,
  we have $h_{\tautilde_{j}}^{1-2\beta}\ge 2h_{\tautilde_{j-1}}^{1-2\beta}$, so
  \begin{align*}
    \sumtT \frac{\alpha_{t}\norm{g_{t}}^{2}}{\sqrt{V_{t}}}
    &\le
      \frac{\epsilon}{\beta} (\norm{g}_{1:T}^{2})^{\beta}\sum_{j=1}^{N}h_{\tilde \tau_{j}}^{1-2\beta}\\
    &\le
      \frac{\epsilon}{\beta} (\norm{g}_{1:T}^{2})^{\beta}\sum_{j=0}^{N-1}h_{\tautilde_{N}}^{1-2\beta}\frac{1}{2^{j}}\\
    &\le
      \frac{\epsilon}{\beta} h_{T}^{1-2\beta}(\norm{g}_{1:T}^{2})^{\beta}\sum_{j=0}^{N-1}\frac{1}{2^{j}}\\
    &\le
      \frac{2\epsilon}{\beta} h_{T}^{1-2\beta}(\norm{g}_{1:T}^{2})^{\beta}
  \end{align*}
  Plugging this back in above and substituting $\beta=\half-\rho$, we have
  \begin{align*}
    R_{T}(\cmp)
    &\le
    2k\norm{\cmp}\sbrac{\sqrt{V_{T+1}\Log{\norm{\cmp}/\alpha_{T+1}+1}}\maxOp h_{T}\Log{\norm{\cmp}/\alpha_{T+1}+1}}
    +\frac{2\epsilon}{\beta} h_{T}^{1-2\beta}(\norm{g}_{1:T}^{2})^{\beta}\\
    &\le
      O\brac{
    \frac{\epsilon h_{T}^{2\rho}}{1-2\rho}V_{T+1}^{\half-\rho}
    +
    \norm{\cmp}\sbrac{\sqrt{V_{T+1}\Log{\frac{\norm{\cmp}B_{T+1}^{\rho}}{\epsilon}+1}}\maxOp h_{T}\Log{\frac{\norm{\cmp}B_{T+1}^{\rho}}{\epsilon}+1}}},
  \end{align*}
  and
  $B_{T+1}^{\rho}=\brac{4\sum_{s=1}^{T+1}\sbrac{2^{\frac{1}{\rho}}+\sum_{s'=1}^{s-1}\frac{\norm{g_{s'}}^{2}}{h_{s'}^{2}}}}^{\rho}$.
\end{proof}

\subsection{Optimistic Trade-offs in the Horizon}%
\label{app:optimistic-tradeoff}

A result analogous to \Cref{thm:pf-ii} can be
shown for our optimistic algorithm as well, and is stated here for
completeness.
Formal proof is omitted since it follows
the same argument as
\Cref{thm:pf-ii} with only
superficial modification: 
following the same steps as \Cref{thm:optimistic-implicit},
we have
\begin{align*}
  R_{T}(\cmp)
  &\le
    2k\norm{\cmp}\sbrac{\sqrt{\hat V_{T+1}\Log{\norm{\cmp}/\hat \alpha_{T+1}+1}}\maxOp 2G\Log{\norm{\cmp}/\hat\alpha_{T+1}+1}}\\
  &\qquad
    +\sumtT\frac{2\hat \alpha_{t}\norm{\grad\ell_{t}(\wt)-\grad\hat\ell_{t}(\wt)}^{2}}{\sqrt{\hat V_{t}}},
\end{align*}
where $\hat V_{t}=16G^{2}+\sum_{s=1}^{\tmm} \norm{\grad\ell_{s}(\w_{s})-\grad\hat\ell_{s}(\w_{s})}^2$.
Now follow the same arguments as \Cref{thm:pf-ii}
to prove the following result.
\begin{restatable}{mytheorem}{OptimisticImplicitII}\label{thm:optimistic-implicit-ii}
  Under the same assumptions as \Cref{thm:adaptive-self-stabilizing},
  let $\rho\in[0,\half)$ and suppose we set $\alpha_{t}=\epsilon G^{2\rho}/\hat V_{t}^{\rho}$ for all $t$.
  Then for all $\cmp\in\R^{d}$, \Cref{alg:centered-md} guarantees
  \begin{align*}
    R_{T}(\cmp)
    &\le
      O\Bigg(\frac{\epsilon G^{2\rho}}{1-2\rho}  \hat V_{T+1}^{\half-\rho}+
      \norm{\cmp}\sbrac{\sqrt{\hat V_{T+1}\Log{\frac{\norm{\cmp}\hat V_{T+1}^{\rho}}{\epsilon G^{2\rho}}+1}}\maxOp G\Log{\frac{\norm{\cmp}\hat V_{T+1}^{\rho}}{\epsilon G^{2\rho}}+1}}\Bigg),
  \end{align*}
where $\hat V_{T+1}=16G^{2}+\sumtT\norm{\grad\ell_t(\wt)-\grad\hat\ell_t(\wt)}^2$
\end{restatable}

\newpage
\section{Supporting Lemmas}\label{app:lemmas}

In this section we collect the miscellaneous supporting lemmas
used in our proofs.

\begin{restatable}{mylemma}{RadiallySymmetric}\label{lemma:radially-symmetric-bound}
  \citep[Lemma 23]{orabona2021parameterfree}
  Let $f:\R\to\R$ and $g:\R^{d}\to\R$ be defined as $g(x)=f(\norm{x})$. If
  $f$ is twice differentiable at $\norm{x}$ and $\norm{x}>0$ then
  \begin{align*}
    \min\Set{g''(\norm{x}),\frac{g'(\norm{x})}{\norm{x}}}I\preceq\grad^{2}g(x)\preceq\max\Set{g''(\norm{x}),\frac{g'(\norm{x})}{\norm{x}}}I
  \end{align*}
\end{restatable}

\begin{restatable}{mylemma}{SimpleRadiallySymmetricBound}\label{lemma:simple-radially-symmetric-bound}
  Under the same assumptions as \Cref{lemma:radially-symmetric-bound},
  further suppose that $f'(x)$ is concave and non-negative.
  If $f$ is twice-differentiable at $\norm{x}$ and $\norm{x}>0$, then
  \begin{align*}
    \grad^{2}g(x)\succeq f''(\norm{x})I
  \end{align*}
\end{restatable}

\begin{proof}
Apply \Cref{lemma:radially-symmetric-bound},
\begin{align*}
  \grad^{2} g(x)\succeq I\min\Set{f''(\norm{x}),\frac{f'(\norm{x})}{\norm{x}}},
\end{align*}
and use the fact that $f'(x)$ is concave and $f'(x)\ge 0$
to bound
\begin{align*}
  \frac{f'(\norm{\x})}{\norm{x}}
  &\ge
    \frac{f'(0)+f''(\norm{x})(\norm{x}-0)}{\norm{x}}
    \ge f''(\norm{\x}).
\end{align*}
\end{proof}

The following integral bound is standard and included for completeness.

\begin{restatable}{mylemma}{SqrtIntegralBound}\label{lemma:sqrt-integral-bound}
  Let $V_{t}\ge\sum_{s=1}^{t}\norm{g_{s}}^{2}$. Then
  \begin{align*}
    \sumtT\frac{\norm{g_{t}}^{2}}{\sqrt{V_{t}}}\le 2\sqrt{\sum_{t=1}^{T}\norm{g_{t}}^{2}}
  \end{align*}
\end{restatable}
\begin{proof}
Using the well-known integral bound
$\sumtT a_{t}f(\sum_{s=0}^{t}a_{s})\le\int_{a_{0}}^{\sum_{s=0}^{t}a_{s}}f(x)dx$
for non-increasing $f$ (See \textit{e.g.} \citet[Lemma 4.13]{orabona2019modern}), we have
\begin{align*}
  \sumtT\frac{\norm{g_{t}}^{2}}{\sqrt{V_{t}}}\le\sumtT\frac{\norm{g_{t}}^{2}}{\sqrt{\norm{g}_{1:t}^{2}}}
  \le\int_{0}^{\norm{g}_{1:T}^{2}}\frac{1}{\sqrt{x}}dx = 2\sqrt{\norm{g}_{1:T}^{2}}.
\end{align*}
\end{proof}

\begin{restatable}{mylemma}{VLogSqrIntegralBound}\label{lemma:v-log-sqr-integral-bound}
  Let $V_{t}\ge 4G^{2}+ \sum_{s=1}^{t-1}\norm{g_{s}}^{2}$ where $G\ge\norm{g_{t}}$ for
  all $t$. Then
  \begin{align*}
    \sumtT\frac{\norm{g_{t}}^{2}}{V_{t}\log^{2}(V_{t}/G^{2})}\le2
  \end{align*}
\end{restatable}

\begin{proof}
  Let $c\ge 4$ and $V_{t}=cG^{2}+\norm{g}_{1:\tmm}^{2}$.
  As in \Cref{lemma:sqrt-integral-bound}, we apply the
  integral bound
  $\sumtT a_{t}f(\sum_{i=0}^{t}a_{t})\le\int_{a_{0}}^{\sum_{s=0}^{t}a_{s}}f(x)dx$
  for non-increasing $f$ to get
  \begin{align*}
    \sumtT\frac{\norm{g_{t}}^{2}}{V_{t}\log^{2}(V_{t}/G^{2})}
    &\le
      \sumtT\frac{\norm{g_{t}}^{2}}{\brac{(c-1)G^{2}+\norm{g}_{1:t}^{2}}\log^{2}\brac{\frac{(c-1)G^{2}+\norm{g}_{1:t}^{2}}{G^{2}}}}\\
    &\le
      \int_{(c-1)G^{2}}^{(c-1)G^{2}+\norm{g}_{1:T}^{2}}\frac{1}{x\log^{2}(x/G^{2})}dx
      =
      \frac{-2}{\Log{x/G^{2}}}\Big|_{x=(c-1)G^{2}}^{(c-1)G^{2}+\norm{g}_{1:T}^{2}}\\
    &\le
      \frac{2}{\Log{c-1}}\le 2,
  \end{align*}
  where the last line uses $\Log{c-1}\ge\Log{3}\ge 1$.

\end{proof}

\newpage
\section{A Simple Reduction for Dynamic Regret in Unbounded Domains}\label{app:dynamic-redux}

Interestingly, a dynamic regret bound of
$R_{T}(\vec{\cmp})\le\Otilde\brac{\sqrt{(M^{2}+MP_{T})\norm{g}_{1:T}^{2}}}$ can be
achieved very simply using a generalization of the one-dimensional reduction of
\cite{cutkosky2018black} to dynamic regret. Note however, that this approach
fails to achieve the improved per-comparator adaptivity observed in \Cref{sec:dynamic}.
The following lemma shows that achieving the
$R_{T}(\vec{\cmp})\le\Otilde\brac{\sqrt{(M^{2}+MP_{T})\norm{g}_{1:T}^{2}}}$ bound in
an unconstrained domain is essentially no harder than
achieving it in a bounded domain, so long as one has access to
an algorithm guaranteeing parameter-free \textit{static} regret.

\newcommand{\oneD}{\text{1D}}

\begin{algorithm}
  \SetAlgoLined
  \KwInput{1D online learning algorithm $\cA_{\oneD}$,
    online learning algorithm $\cA_{S}$ with domain equal to the unit-ball
    $S\subseteq \left\{x\in\R^{d}:\norm{x}\le 1\right\}$}\\
  \For{$t=1:T$}{
    Get point $x_{t}\in S$ from $\cA_{S}$\\
    Get point $\beta_{t}\in\R$ from $\cA_{\oneD}$\\
    Play point $\wt=\beta_{t}x_{t}\in \R^{d}$, receive subgradient $g_{t}$\\
    Send $\hat g_{t}=\inner{g_{t},x_{t}}$ to $\cA_{\oneD}$ as the $t^{\text{th}}$ loss\\
    Send $g_{t}$ to $\cA_{S}$ as the $t^{\text{th}}$ loss
  }
  \caption{One-dimensional Reduction \parencite{cutkosky2018black}}
  \label{alg:dimension-free}
\end{algorithm}

\begin{restatable}{mylemma}{DimensionFree}\label{lemma:dimension-free}
  Suppose that $\cA_{S}$ guarantees dynamic regret
  $R_{T}^{\cA_{S}}(\vec{\cmp})$ for any sequence $\cmp_{1},\ldots,\cmp_{T}$ in the
  unit-ball $S=\Set{\w\in\R^{d}:\norm{\w}\le 1}$ and suppose $\cA_{\oneD}$ obtains static regret $R_{T}^{\cA_{\oneD}}(\cmp)$ for any
  $\cmp\in\R$. Then for any $\cmp_{1},\ldots,\cmp_{T}$ in $\R^{d}$,
  \Cref{alg:dimension-free} guarantees
  \begin{align*}
    R_{T}(\vec{\cmp})
    &=
      R_{T}^{\cA_{\oneD}}(M) + M R_{T}^{\cA_{S}}\brac{\frac{\vec{\cmp}}{M}}
  \end{align*}
  where $M=\max_{t\le T}\norm{\cmp_{t}}$.
\end{restatable}
\begin{proof}
the proof follows the same reasoning as in the static regret
case \parencite[Theorem 2]{cutkosky2018black}:
\begin{align*}
  R_{T}(\vec{\cmp})
  &=
    \sumtT\inner{g_{t},\wt-\cmp_{t}}
    =
    \sumtT\inner{g_{t},\beta_{t}x_{t}-\cmp_{t}}\\
  &=
    \sumtT \inner{g_{t},x_{t}}\beta_{t}+\Big[\inner{g_{t},x_{t}}M - \inner{g_{t},x_{t}}M\Big] - \inner{g_{t},\cmp_{t}}\\
  &=
    \sumtT\inner{g_{t},x_{t}}\beta_{t}-\inner{g_{t},x_{t}}M+\sumtT\inner{g_{t},x_{t}}M-\inner{g_{t},\cmp_{t}}\\
  &=
    \sumtT\hat g_{t}(\beta_{t}-M) +M \sumtT\inner{g_{t},x_{t}-\frac{\cmp_{t}}{M}}
    = R_{T}^{\cA_{\oneD}}(M)+ M R_{T}\brac{\frac{\vec{\cmp}}{M}}
\end{align*}
\end{proof}

Using this, one could let $\cA_{\oneD}$ be any parameter-free algorithm
and let $\cA_{S}$ be any algorithm which achieves the desired
dynamic regret on a bounded domain. For instance, to get the optimal
$\sqrt{P_{T}}$ dependence we can choose  $\cA_{S}$ to be the Ader algorithm of
\textcite{zhang2018adaptive}, which will guarantee
$MR_{T}^{\cA_{S}}(\frac{\vec{\cmp}}{M})\le O\brac{MG\sqrt{T\brac{ 1 + \frac{P_{T}}{M}}}}=O\brac{G\sqrt{(M^{2}+MP_{T})T}}$.

\section{Amortized Computation for Dynamic Regret}\label{app:amortized}

\begin{figure*}
  \begin{algorithm}[H]
    \SetAlgoLined
    \KwInput{Algorithm $\cA$, Disjoint intervals $I_{1},\ldots,I_{K}$ such
      that $\cup_{k=1}^{K}I_{k}\supseteq[1,T]$}\\
    Get $\w_{1}$ from $\cA$\\
    Set $k=1$\\
    \For{$t=1:T$}{
      Play $\w_t$, observe loss $g_{t}$\\
      \uIf{$t+1\notin I_{k}$}{
        Send $\gtilde_{k}=\sum_{s\in I_{k}}g_{s}$ to $\cA$\\
        Update $k\gets k+1$\\
        Get $\wtpp$ from $\cA$
      }\Else{
        Set $\wtpp=\wt$
      }
    }
    \caption{Lazy Reduction for Amortized Computation}
    \label{alg:lazy}
  \end{algorithm}
\end{figure*}

All known algorithms which achieve the optimal
$O(\sqrt{TP_{T}})$ dynamic regret
follow a similar construction,
in which several instances of a simple base algorithm $\cA$
are run simultaneously and a meta-algorithm
combines their outputs in a way that guarantees
near-optimal performance.
Assuming the base algorithm $\cA$ uses
$O(d)$ computation per round,
the full algorithm then requires $O(d\Log{T})$
computation per round. Ideally we'd like to avoid this $\log(T)$ overhead. A simple way to  combat this difficulty
is to only update the algorithm every $O(\Log{T})$
rounds, so that the \textit{amortized} computation per round
is $O(d)$ on average.
The following proposition shows that
$R_{T}(\vec{\cmp})\le O\brac{\sqrt{TP_{T}}}$ can be
maintained up to poly-logarithmic terms using only $O(d)$ per-round computation
on average by updating only at the end of itervals $I_{k}$ of length $O\brac{\Log{T}}$.

\begin{myproposition}\label{prop:lazy}
  Suppose $\cA$ is an online learning algorithm which guarantees
  \begin{align*}
    R_{T}^{\cA}(\vec{\cmp})\le\widetilde{O}\brac{\sqrt{(M^{2}+ M P_{T})\sumtT\norm{g_{t}}^{2}}},
  \end{align*}
  for all $\cmp_{1},\ldots,\cmp_{T}$ in $\R^{d}$ with $\max_{t\le T}\norm{\cmp_{t}}\le M$.
  Then for all $\cmp_{1},\ldots,\cmp_{T}$ in $\R^{d}$, \Cref{alg:lazy} guarantees
  \begin{align*}
    R_{T}(\vec{\cmp})
    \le\widetilde{O}\brac{\max_{k\le K}|I_{k}|\sqrt{(M^{2}+M P_{T})\norm{g}^{2}_{1:T}}}
  \end{align*}
\end{myproposition}

\begin{proof}
First observe that for any interval $I=[a,b]$, we have
\begin{align*}
  \sum_{t\in I}\inner{g_{t}, \wt-\cmp_{t}} &= \sum_{t\in I}\inner{g_{t},\wt-\cmp_{b}} +\sum_{t\in I}\inner{g_{t},\cmp_{b}-\cmp_{t}},
\end{align*}
and bound the second sum as
  \begin{align*}
    \sum_{t\in I}\inner{g_{t},\sum_{s=t+1}^{b}\cmp_{s}-\cmp_{s-1}}
    &=
      \sum_{t=a}^{b}\sum_{s=t+1}^{b}\inner{g_{t},\cmp_{s}-\cmp_{s-1}}\\
    &=
      \sum_{s=a+1}^{b}\inner{g_{a:s-1},\cmp_{s}-\cmp_{s-1}} \le \sqrt{\sum_{s=a+1}^{b}\norm{g_{a:s-1}}^{2}\sum_{t=a+1}^{b}\norm{\cmp_{t}-\cmp_{\tmm}}^{2}}\\
    &\le
      \sqrt{\brac{\sum_{t=a+1}^{b}\norm{g_{t}}^{2} + \sum_{t=a+1}^{b}\sum_{t'\ne t}^{b}\norm{g_{t}}\norm{g_{t'}}}S_{I}}\\
    &\le
      \sqrt{\brac{\sum_{t=a+1}^{b}\norm{g_{t}}^{2} + \max_{s\in[a,b]}\norm{g_{s}}^{2}|I|^{2}}S_{I}}\\
    &\le
      \sqrt{2\norm{g}_{a+1:b}^{2}|I|^{2}S_{I}}
      =
      \sqrt{2\norm{g}_{a+1:b}^{2}S_{I}}|I|.
  \end{align*}
  where $S_{I}=\sum_{t=a+1}^{b}\norm{\cmp_{t}-\cmp_{\tmm}}^{2}$.
  Thus, denoting $I_{1}=[1,\tau_{1}]$,
  $I_{2}=[\tau_{1}+1,\tau_{2}],\ldots, I_{K}=[\tau_{K-1}+1,\tau_{K}]$, we can bound
  \begin{align*}
    \sumtT\inner{g_{t},\wt-\cmp_{t}} &= \sum_{k=1}^{K}\sum_{t\in I_{k}}\inner{g_{t},\wt-\cmp_{t}}
                                     =\sum_{k=1}^{K}\sum_{t\in I_{k}}\inner{g_{t}, \wt-\cmp_{\tau_{k}}} + \inner{g_{t},\cmp_{\tau_{k}}-\cmp_{t}}
    \\&\le \sum_{k=1}^{K}\sum_{t\in I_{k}}\inner{g_{t}, \wt-\cmp_{\tau_{k}}} + \sum_{k=1}^{K}\sqrt{2S_{I_{k}}\norm{g}_{t\in I_{k}}^{2}}|I_{k}|\\
                                   &\le\sum_{k=1}^{K}\inner{\sum_{t\in I_{k}}g_{t}, \w_{\tau_{k}}-\cmp_{\tau_{k}}} + \sqrt{2S_{T}\norm{g}_{1:T}^{2}}\max_{k\le K}|I_{k}|
  \end{align*}
  where the last line observes that $\wt$ is fixed within each interval.
  From the regret guarantee of algorithm $\cA$ we have
  \begin{align*}
    \sum_{k=1}^{K}\inner{\sum_{t\in I_{k}}g_{t},w_{\tau_{k}}-\cmp_{\tau_{k}}}
    &=
      \sum_{k=1}^{K}\inner{\gtilde_{\tau_{k}},\w_{\tau_{k}}-\cmp_{\tau_{k}}}
      \le \Otilde\brac{\sqrt{(M^{2}+M\hat P_{K})\norm{\gtilde}_{1:K}^{2}}}\\
    &\le
      \Otilde\brac{\max_{k\le K} |I_{k}|\sqrt{2(M^{2}+M P_{T})\norm{g}_{1:T}^{2}}},
  \end{align*}
  where the first line defines
  $\hat P_{K}=\sum_{k=2}^{K}\norm{\cmp_{\tau_{k}}-\cmp_{\tau_{k-1}}}$ and
  the last line observes $\hat P_{K}\le P_{T}$. Hence,
  \begin{align*}
    \sumtT\inner{g_{t},\wt-\cmp_{t}}
    &\le
      \Otilde\brac{\max_{k\le K}|I_{k}|\brac{\sqrt{2(M^{2}+MP_{T})\norm{g}_{1:T}^{2}}+\sqrt{2S_{T}\norm{g}_{1:T}^{2}}}}.
  \end{align*}
  The stated bound follows by observing that $S_{T}\le MP_{T}\le M^{2}+MP_{T}$ and
  hiding constants.

  \end{proof}

\end{document}